\documentclass[acmsmall,screen]{acmart}
\AtBeginDocument{%
  }

\acmJournal{TIST}
\usepackage{algorithm}
\usepackage{algpseudocode}
\usepackage{subcaption}
\usepackage[flushleft]{threeparttable}
\usepackage{booktabs}
\usepackage{multirow}
\usepackage{animate}

\begin{document}

\title{Variational Mode-Driven Graph Convolutional Network for Spatiotemporal Traffic Forecasting}

\author{Osama Ahmad}
\orcid{0009-0003-2124-6114}
\affiliation{%
  \institution{Lahore University of Management Sciences}
  \city{Lahore}
  \country{Pakistan}
}
\email{osama_ahmad@lums.edu.pk}

\author{Lukas Wesemann}
\orcid{0000-0001-9142-1342}
\affiliation{%
  \institution{Maincode}
  \city{Melbourne}
  \country{Australia}
}
\email{lukas@maincode.com}

\author{Fabian Waschkowski}
\orcid{0000-0001-5427-9551}
\affiliation{%
  \institution{Maincode}
  \city{Melbourne}
  \country{Australia}
}
\email{fabian@maincode.com}

\author{Zubair Khalid}
\orcid{0000-0001-7875-4687}
\affiliation{%
  \institution{Lahore University of Management Sciences}
  \city{Lahore}
  \country{Pakistan}
}
\affiliation{%
  \institution{Maincode}
  \city{Melbourne}
  \country{Australia}
}
\email{zubair.khalid@lums.edu.pk}


\begin{abstract}
This paper focuses on spatiotemporal (ST) traffic prediction using graph neural networks (GNNs). Given that ST data comprises non-stationary and complex temporal patterns, interpreting and predicting such trends is inherently challenging. Representing ST data in decomposed modes helps infer underlying behavior and assess the impact of noise on predictive performance. We propose a framework that decomposes ST data into interpretable modes using variational mode decomposition (VMD) and processes them through a neural network for future state forecasting. Unlike existing graph-based traffic forecasters that operate directly on raw or aggregated time series, the
proposed hybrid approach, termed the Variational Mode Graph Convolutional Network (VMGCN), first decomposes non-stationary signals into interpretable variational modes by determining the optimal mode count via reconstruction-loss minimization and then learns both intramode and cross-mode spatiotemporal dependencies through a novel attention-augmented GCN. Additionally, we analyze the significance of each mode and the effect of bandwidth constraints on multi-horizon traffic flow predictions. The proposed two-stage design yields significant accuracy gains while providing frequency-level interpretability with demonstrated superior performance on the LargeST dataset for both short-term and long-term forecasting tasks. The implementation is publicly available on \href{https://github.com/OsamaAhmad369/VMGCN}{GitHub}.
\end{abstract}




\maketitle

\section{Introduction}
Traffic Management Systems (TMS) play a vital role in the growth of urbanization. The increasing number of vehicles on the road consistently pressures traffic planning and management. An Intelligent Transportation System (ITS) enhances the efficiency, reliability, and safety of transportation systems~\cite{ITS_RL}. Accurate predictions help avoid congestion and improve traffic infrastructure, mitigating issues like air pollution, fuel and time wastage, and accidents. These problems directly impact human health, daily life, social life, and psychological states, including lack of control, fatigue, time pressure, and stress. For effective online monitoring and prediction, smart cameras or sensors must be installed at crowded intersections, and path-planning user applications should guide individuals away from traffic jams. The traffic counts or flow hold statistical significance in data analytics, allowing predictions of increasing vehicle numbers over time and planning for traffic infrastructure accordingly. Socioeconomic indicators and policies also play roles in traffic management. The policies based on speed or flow can be implemented to prevent road crashes, motivating us to address the problem of developing accurate and robust traffic prediction methods. In this paper, we are addressing traffic prediction using a deep learning method. 

\subsection{Related Work}
In traffic forecasting, both parametric and non-parametric models are fundamental approaches for predicting future states. The choice of model depends on the size and complexity of the data, as well as the available computational resources. Parametric models include time series analysis, parametric regression, and Kalman filter models. Time series analysis models, such as the auto-regressive integrated moving average (ARIMA), seasonal ARIMA (SARIMA), and exponential smoothing.  The non-parametric approaches, including K-nearest neighbors, decision trees, and support vector machine (SVM) have been widely used. 

The Box-Jenkins ARIMA model, which transforms non-stationary data to stationary data using the differences in the time series, is frequently incorporated in traffic volume prediction~\cite{ARIMA}. Another approach for prediction in~\cite{kalman}, uses measured state data to forecast traffic flow, minimizing prediction errors based on the assumption of data stationarity~(Kalman filtering theory). These parametric methods are interpretable but struggle to model complex and non-stationary traffic data effectively. Non-parametric models utilize distance metrics to determine the nearest neighbor of the current state to the historical observation in feature space~\cite{Nonparameter}. These methods are rich in data requirements, less interpretable, and computationally complex. Both parametric and non-parametric methods utilize historical observed data but do not incorporate spatial information. Due to recent advancements in deep learning, significant improvements have been made in traffic forecasting applications. The state-of-the-art methods in traffic prediction consider various factors, such as spatial topology construction, spatial dependency, temporal dependency, and external factors. These advanced methods are designed to better handle the complexities of modern traffic data and provide more accurate predictions over both short (refer to horizons of up to one hour) and long term (beyond one hour)~\cite{book_transport}.

\subsubsection*{Spatial topology construction and spatial dependency} Determining a graph structure that specifies the connectivity of spatial locations is known as the adjacency matrix. Usually, the non-Euclidean geodesic information of the sensor network is incorporated in the model using graph neural networks (GNNs). If the graph topology is defined based on the distance between the nodes, this representation is known as a static graph. The semantic adjacency matrix is computed using dynamic time warping (DTW) distance by incorporating the time series of nodes~\cite{STGODE},~\cite{wang2024score}. The binary adjacency matrix represents the presence of the connection between nodes~\cite{ASTGCN}. The road network is also included in the distance-based graph structure~\cite{DCRNN_MetaLA}. The connectivity of a graph is determined using a Gaussian thresholded kernel function and utilizes the weighted adjacency matrix~\cite{zheng2020gman}. Some literature also built an adjacency matrix using point of interest (POI) similarity~\cite{geng2019spatiotemporal}, edge-wise graph~\cite{chen2020multi}, and free-flow reachable matrix~\cite{cui2019traffic}. In an adaptive graph, the structure of the graph is constructed based on the temporal evolution of the data~\cite{huang2025transformer}. For spatial graph modeling based on the dynamic features, using node embeddings, a self-adaptive adjacency matrix is determined in an end-to-end learning mechanism~\cite{GWNET},~\cite{AGCRN},~\cite{d2stgnn}.

In spatial dependencies, we dynamically derive the correlation between nodes of features (Inter-node dependencies) or multiple features of the same node (Intra-node dependencies) from the data. To extract the spatial features of traffic flows from the road network matrix, trafficGAN~\cite{TrafficGAN} implements deformable convolution~\cite{deformablecovolution}. Deformable convolution is used to model geometric variations. To capture the spatial dependency, mostly convolutional neural network (CNN)~\cite{curbGAN},~\cite{deepstn}, and GNN~\cite{DCRNN_MetaLA},~\cite{TrafficTransformer},~\cite{AGCRN} are employed in prediction applications.  Dynamic graph convolutional recurrent network~\cite{DGCRN} incorporates the weighted combination of both static and dynamic graph convolution. The self-attention mechanism~\cite{TripletAttention} is also used in the modeling of spatial dependency. In the most recent literature on traffic prediction~\cite{DSTAGNN},~\cite{ASTGCN}, the combination of attention and GNN is also encouraged to determine the spatial correlation. The meta-graph attention, which is based on graph attention (GAT) is applied to determine the spatial correlations from traffic data ~\cite{STMetaNet}. Chebyshev spectral convolution (ChebNet) is incorporated to learn spatial dependencies for spectral graph convolution~\cite{STGCN}. To overcome the smoothing issues in GNNs in deeper networks that impact the long-range dependencies, the spatial-temporal graph ordinary differential equation (STGODE) block is proposed in~\cite{STGODE}. STGODE captures the spatial and temporal dynamic correlations based on ODE. Similarly,  neural rough differential equations have been incorporated to capture spatio-temporal dynamics~\cite{choi2023graph}. A spatio-temporal transformer graph is used to capture long-term dependencies~\cite{huang2025transformer}.

\subsubsection*{Temporal dependency} Traffic prediction involves time series data used to capture sequential relationships. Recurrent neural networks (RNNs) and their optimized variants, such as long short-term memory (LSTM) and gated recurrent units (GRUs), are commonly employed. These models are capable of capturing the long-range temporal dependencies~\cite{DGCRN},~\cite{zhang2022urban}. However, combining dynamic graphs with RNNs makes the model computationally expensive due to a large number of trainable parameters. Temporal convolution filters are also used to extract the temporal information from the data. The attention mechanism~\cite{attention} in the spatiotemporal (ST) block is commonly used in predictive models~\cite{ASTGCN},~\cite{DSTAGNN}. A combination of RNNs along with the attention mechanism is also employed. GRUs handle the short-term dependencies, while the self-attention mechanism models the long-term dependencies~\cite{d2stgnn}. Three different types of attention-- temporal, spatial, and channel-- are applied to capture spatiotemporal correlations~\cite{TripletAttention}. However, this proposed mechanism increases the computational, time complexity, and number of parameters compared to CNNs. CNNs and temporal convolutional networks (TCNs) have proven more efficient in capturing complex spatiotemporal relationships. The transformer encoder-decoder architecture is implemented for the temporal pattern in~\cite{TrafficTransformer},~\cite{STMetaNet}. Typically, the encoder is implemented to learn historical traffic information~\cite{han2024bigst} and the decoder is incorporated to make predictions.  

\subsubsection*{External factors} Traffic data consists of complex dynamics that depend on various factors such as socio-economic, land uses land covers, population density, road capacity, and cultural or public events~\cite{li2022deep}.  To apply self-attention to the point of interest (POI) of each region, POI-MetaBlock is included in a deep neural network (DNN)~\cite{poi_metablock}. To effectively capture temporal correlation, the external weather data is incorporated in DNNs for traffic flow forecasting in~\cite{weather_data}. The non-traffic data, such as rainfall~\cite{rain-fall} is also included in the model. Information such as holidays, weather, and nearby POIs is served as the input to the Large Language Models (LLMs)~\cite{R2T_LLM}. These factors serve as the features of DNNs that assist the network in extracting the spatial and temporal correlation from the traffic data. 

Fourier domain analysis is also used in traffic forecasting, where the traffic data is segregated into frequency components. However, the limitation of the Fourier Transform is that the data should be stationary. Sun et al.~\cite{sun2024modwavemlp} proposed a wavelet learning based decomposition for denoising and determining the frequency components. However, this method may suffer from modes overlapping in high-dynamical systems.  So, many signal processing-based methods have been introduced for the decoupling of signals~\cite{wang2024w}.  Various methods are commonly used for signal decomposition, including the well-known methods Empirical Mode Decomposition (EMD)~\cite{emd_hilbert}, Ensemble EMD (EEMD)~\cite{EEMD}, Complete Ensemble EMD with Adaptive Noise (CEEMDAN)~\cite{CEEMDAN}, Multivariate EMD (MEMD)~\cite{MEMD}, and Variational Mode Decomposition (VMD)~\cite{VMD}.  The traffic data is decomposed into periodic, residual, and volatile components using the EMD method. A Bi-directional LSTM model is incorporated to model the volatile component~\cite{EMD_LSTM}. A hybrid model, combining Improved CCEMDAN (ICCEMDAN) and GRU, is proposed to predict the parking demand~\cite{ICEEMDAN_GRU}.~\cite{tian2025traffic} used the permutation entropy algorithm to rank intrinsic mode functions (IMFs), derived from ICEEMDAN, IMFs having a small entropy value fed to ARIMA for traffic prediction.  Short-term traffic flow is predicted using VMD and LSTM~\cite{VMD_LSTM}. The encoder-decoder architecture is incorporated for feature extraction from mode functions decomposed by VMD in short-term traffic flow prediction~\cite{VMD_autoencoder}. However, these methods do not incorporate the spatial and topological information. Cross-channel dependencies between decomposed modes are determined by using convolution~\cite{zhang2025vmd},~\cite{su2024mdcnet}. These frameworks do not incorporate the graph structure. Existing decomposition methods, such as EMD, EEMD, and CEEMDAN-based approaches, suffer from mode-mixing, while VMD overcomes these challenges by ensuring better frequency separation and adaptability and therefore, offers non-recurrent frequency modes and improved forecasting performance~\cite{zhao2023hybrid}.

\begin{figure*}[!t]
    \centering
   \begin{subfigure}{0.32\textwidth}
   \animategraphics[controls,loop,autoplay,width=\linewidth]{1}{figures/sd_horizon_3/frame_}{0}{10} 
        \caption{}
        \label{fig:animation_horizon_3}
    \end{subfigure}\hfill
    \begin{subfigure}{0.32\textwidth}
    \animategraphics[controls,loop,autoplay,width=\linewidth]{1}{figures/sd_horizon_6/frame_}{0}{10} 
        \caption{}
        \label{fig:animation_horizon_6}
    \end{subfigure}\hfill
    \begin{subfigure}{0.32\textwidth}
    \animategraphics[controls,loop,autoplay,width=\linewidth]{1}{figures/sd_horizon_12/frame_}{0}{10} 
        \caption{}
         \label{fig:animation_horizon_12}
    \end{subfigure}
    \hfill
    \begin{subfigure}{0.32\textwidth}
        \includegraphics[width=\linewidth]{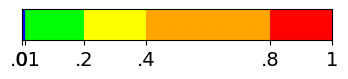}
    \end{subfigure}
    \caption{A visualization showcasing the temporal normalized loss variations on SD region for (a) Horizon 3 (b) Horizon 6 (c) Horizon 12 using our proposed model. Please use Adobe Reader for enhanced and optimal visualization. }
    \label{animation}
\end{figure*}
\subsection{Contributions}

To bridge the gap in existing approaches that either ignore the multi-frequency structure of traffic time series or require manual tuning of decomposition parameters, we introduce the Variational Mode Graph Convolutional Network (VMGCN), the first end-to-end framework, that decomposes each node signal into optimally selected variational modes and jointly learns spatiotemporal interactions across these modes using attention-enhanced graph convolutions. In contrast to attention-based spatiotemporal graph convolutional networks (ASTGCN) and Graph WaveNet (GWNET), which learn patterns directly from raw traffic flows, VMGCN leverages frequency-domain insights to disentangle non-stationary effects before applying graph learning. The main contributions of this work are summarized below:

\begin{itemize}
  \item In our hybrid framework for spatiotemporal forecasting, we propose to decompose the spatiotemporal traffic data into modes representing the granularity of the data using VMD. The spatial and temporal attention are applied to the decomposed features, followed by ChebNet graph convolution and time convolution to predict the future states over the long term. 
  Such an attention-augmented GCN processes each variational mode separately before fusing both frequency-specific and cross-mode dependencies. 
  \item We analyse the orthogonality, frequency overlapping, and residual components of decomposed modes of the graph network to reveal the significance of mode components. 
  \item We also conduct an in-depth analysis of our proposed architecture and formulate a method to determine the best-fit value of the number of modes. Additionally, we highlight the impact assessment of different hyper-parameters on the long-term prediction performance of the model.  
  \item We evaluate the predication capability of the proposed architecture on the LargeST~\cite{LargeST} dataset across three different regions using different metrics and demonstrated superior performance as compared to state-of-the-art GNNs. The collective nodes of the LargeST dataset are $6902$, sampled at $15$ minute time interval over one year (containing $241$ million spatiotemporal observations). As an illustration, Fig.~\ref{animation} demonstrates the prediction loss of our proposed model on the SD region. We also analyze the computational complexity of the proposed method. 
\end{itemize}

The rest of the paper is organized as follows. Section {\ref{sec:pre}} describes the adopted notation and formulates the problem statement. Section {\ref{sec:methodology}} presents the detailed architecture of the network used in our framework along with the mathematical formulation of the decomposition method. Section {\ref{sec:result}} presents the experimental analysis and the improvements enabled by our methods before Section {\ref{sec:conclusion}} concludes the paper.

\section{Preliminaries}
\label{sec:pre}
\subsection{Road Network} 
In graph theory, a road is modeled by considering different nodes at distinct joints, vertices, or intersections. Each node of the road is connected to another segment or node. The interconnectivity and distance among nodes are incorporated in a graph referred to as a road network, defined as $\mathcal{G}=(V,\, A,\, Y)$, where $V$ is a set of entities or nodes (vertices), $Y$ represents the features, and $A \in \mathbb{R}^{N \times N} $ is the weighted adjacency matrix. Here, $N$ represents the total number of nodes in a traffic prediction application, which could be either the number of sensors or vertices of a road network.
\subsection{Temporal Traffic Network Graph}
\begin{figure}[!t]
\centering
  \includegraphics[width=0.55\textwidth]{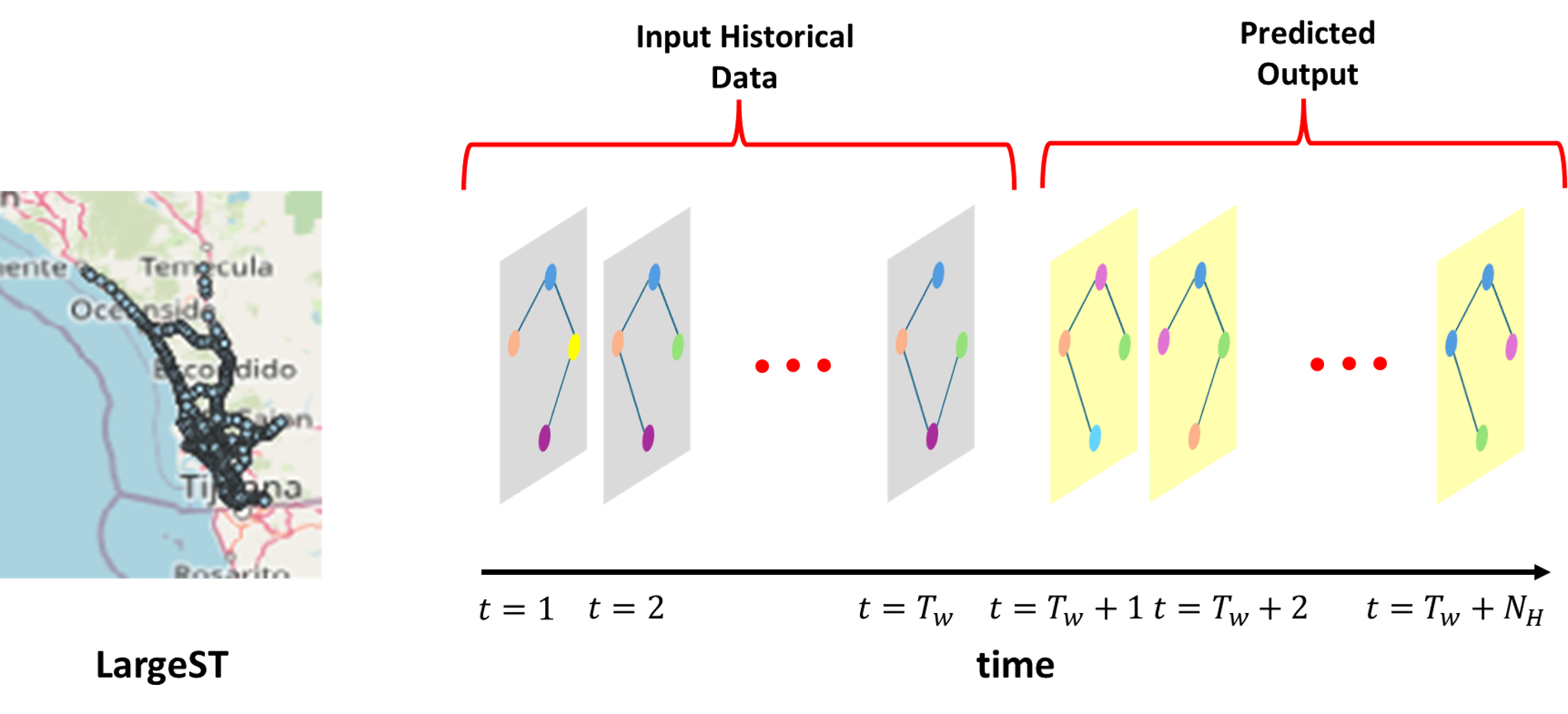}
  \caption{Left side illustrates the spatial distribution of nodes in the San Diego region, as represented in the LargeST dataset. On the right side, the TTNG structure is depicted, where the historical and current input graphs are sampled at time intervals $t=\{1,2,\dots,T_w\}$. The output graphs are then predicted for $t=\{T_w+1,T_w+2,\dots,T_w+N_H\}$.}
  \label{fig:graph_structure} 
\end{figure}

Traffic flow, the number of vehicles per unit of time in a road segment, is a key measure of traffic congestion, which also depends on the width or number of road lanes. A temporal dynamic graph is an arrangement of graph snapshots [$\mathcal{G}^{(1)},\,\mathcal{G}^{(2)},\dots,\,\mathcal{G}^{(T_w)}$]  where $\mathcal{G}^{(t)}=(V^{(t)},\, A^{(t)},\,Y^{(t)})$ includes vertices $V^{(t)}$, an adjacency matrix $ A^{(t)}$, and a feature matrix $Y^{(t)}$~\cite{GNNBook-ch15-kazemi}. Here, $T_w$ is the time window depending on the historical observation of the signal. The feature matrix $Y^{(t)}$ may include traffic information and road network characteristics. This structure,  known as the temporal traffic network graph (TTNG), explains the traffic flow among different nodes from time $[1, T_w+N_H]$. As an illustration, Fig.~\ref{fig:graph_structure} depicts the sequence of temporal graphs for a traffic network. 
\subsection{Problem Formulation}
The dynamic evolving graph can be defined as time-dependent $\mathcal{G}^{(t)}=(V^{(t)},\, A^{(t)},\,Y^{(t)})$. We assume that the vertices are constant, but the features are time-dependent and are also influenced by unforeseen conditions such as road accidents, road closures, weather, and public events, which alter the course of traffic flow. In this work, we aim to predict the future horizon traffic flow [$Y^{(t+1)},\, Y^{(t+2)},\dots, \, \boldsymbol{Y}^{(t+N_H)}$] for the next prediction horizon length $N_H$ given $\mathcal{G}^{(t)}$ for $t \in [1,\, T-N_H]$, i.e.,
\begin{equation*}
\begin{split}
    [Y^{(t-T_w+1)},\dots, Y^{(t)}] \stackrel{F}{\rightarrow}
[Y^{(t+1)},\dots,Y^{(t+N_H)}],
\end{split}
\end{equation*}
where $T$ is the length of the signal, $N_H$ is the length future horizon,  and $F$ is a deep-learning-based function that takes current and historical TTNG as input to yield the future TTNG.

\section{Methodology}
\label{sec:methodology}

\subsection{Mathematical Formulation}
The proposed prediction framework is comprised of pre-processing of the time signal using variational mode decomposition (VMD)~\cite{VMD} followed by the spatiotemporal forecasting using the deep-learning model~\cite{ASTGCN}. 
\subsubsection{Variational Mode Decomposition (VMD)}
The objective of VMD is to decouple the original signal into sub-signals $u_k$, known as modes. The analytic signal $f_A(t)$ is a complex-valued function that contains only non-negative frequency components. An oscillatory signal is a valid intrinsic mode function (IMF) if it fulfills the two main conditions; (1) the number of extrema (minima and maxima) and zero-crossings must either equal or differ at most by one;  (2) at any point, the mean value of the
envelope defined by the local maxima and the envelope defined by the local minima
is zero~(as defined in \cite{emd_hilbert}). The analytic signal is a complex signal where the real part comprises the original signal $f(t)$  and the imaginary part is the Hilbert transform $\mathcal{H}f(t)$ of $f(t)$, that is
\begin{equation}
\label{eq:1}
    f_A(t)=f(t)+j\mathcal{H}f(t).
\end{equation}
We solve the following optimization problem to decompose the analytic signal $f_A(t)$ into its variational modes~\cite{VMD}:
\begin{align}  
\begin{aligned}
   \min_{u_k, \omega_k}\quad  & \sum_k\lVert\partial_t[f_A(t)*u_k(t)] e^{-j\omega_kt}\rVert_2^2,  \\
   & \text{such that} \quad \sum_k u_k=f,
   \end{aligned}
   \label{eq:2}
\end{align}
where the `$*$' denotes the convolutional operation. The term $f_A(t)*u_k(t)$ in the objective function computes frequency-specific components using the Hilbert transform associated with each mode. The exponential term shifts the signal by the center frequency $\omega_k$ of each mode. The derivative of the analytical signal of each mode indicates the variation or changes in each mode. In this optimization problem, we are minimizing the sum of the variation of all modes. The signal is decomposed into sub-signals under the constraint that the sum of all components reproduces the original signal. 

Here, $u_k=\{u_1,u_2,\dots,u_K\}$ and $\omega_k=\{\omega_1,\omega_2,\dots,\omega_K\}$ are sets of all $K$ modes and their center frequencies, respectively. If the Dirac delta function is the original function $f=\delta(t)$, the impulse response of the Hilbert transform will be $\frac{j}{\pi t}$. The objective function in~\eqref{eq:2} can be expressed as
\begin{align}
\begin{aligned}
   \min_{u_k, \omega_k} & \left\{ \sum_k\lVert\partial_t[(\delta(t)+\frac{j}{\pi t})*u_k(t)] e^{-j\omega_kt}\rVert_2^2 \right\}, \\  
   & \text{such that} \quad \sum_k u_k=f.
\end{aligned}
\label{eq:3}
\end{align}
Now, reformulate the above constraint optimization to an unconstrained optimization problem using the Lagrangian method.   The augmented Lagrangian $\mathcal{L}$ is described as
\begin{equation}
\label{eq:4}
\begin{split}
   \mathcal{L}(u_k,\omega_k,\lambda)=\alpha \sum_k\lVert\partial_t[(\delta(t)+\frac{j}{\pi t})*u_k(t)] e^{-j\omega_kt}\rVert_2^2+\\ \lVert f(t)-\sum_ku_k(t)\rVert_2^2+
   \langle \lambda(t),f(t)- \sum_ku_k(t)\rangle,
   \end{split}
\end{equation}

\begin{equation}
\label{eq:4.2}
   u_k^{(n+1)}=\rm{arg}\min_{u_k}\mathcal{L}(\{u^{n+1}_{i<k}, u^{n}_{i\geq k}\}, \omega_i^n, \lambda^n),
\end{equation}

\begin{equation}
\label{eq:4.3}
   \omega_k^{(n+1)}=\rm{arg}\min_{\omega_k}\mathcal{L}(u^{n+1}_{i},  \{\omega^{n+1}_{i<k}, \omega^{n}_{i\geq k}\}, \lambda^n),
\end{equation}
where $\alpha$ is the bandwidth constraint in the variational mode term to control the bandwidth of the modes. The second term is a quadratic penalty for the reconstruction fidelity term and ensures the convergence properties at finite weight. The Lagrangian multipliers $\lambda(t)$ strictly enforce the constraints. The solution to the mode minimization problem is given by
\begin{equation}
\label{eq:9}
   \hat{u}_k^{(n+1)}(\omega)=\frac{\hat{f}(\omega)-\sum\limits_{i \neq k}\hat{u_i}(\omega)+\frac{\hat{\lambda}^n(\omega)}{2}}{1+2\alpha(\omega-\omega_k^n)^2},
\end{equation}
where $\hat{f}(\omega)$ is the Fourier Transform of $f$ and  $\hat{u}_k(\omega)$ denotes the modes in frequency domain. By adjusting the bandwidth constraint,  we trade off between the accuracy and smoothness of the bandwidth of the decomposed modes. The formulation of the minimization problem for center frequency~\eqref{eq:4.3} is given by
\begin{equation}
   \omega_k^{(n+1)}=\frac{\displaystyle \int_0^\infty \omega |\hat{u}_k(\omega)|^2d\omega}{\displaystyle \int_0^\infty |\hat{u}_k(\omega)|^2d\omega}.
\end{equation}

\noindent To solve the above optimization problems that require the minimization of two variables $\hat{u}_k(\omega)$ and $\omega_k$, the alternating direction method of multipliers (ADMM) decomposes this dual optimization in two sub-problems and solves each sub-problem individually as described in Algorithm \ref{alg:1}. Since the input of the signal is in the time domain, consisting of finite and discrete samples, the pre-processing step involves the use of discrete Fourier transform (DFT) to obtain a frequency domain signal. The implicit assumption is that the signal, representing a one-time period, is periodic in time. This assumption results in a discontinuity at the endpoints which can be resolved by mirroring the time signal on both ends around the central axis before applying DFT. The post-processing involves the inverse operation, where the frequency domain signal is converted back to the time domain signal using the inverse DFT. The truncation of the signal is carried out to reverse the mirroring operation to convert the signal to the original length. 
\begin{equation*}
    \phi (t)=f(t)-\sum_{k=1}^K u_k(t),
\end{equation*}
where the redemption~(residual) function $\phi (t)$ is the difference between the original signal and the summation of the modes at time $t$. 
\begin{algorithm}[!t]
\centering
\caption{Decomposition using VMD}
\label{alg:1}
\begin{algorithmic}
\State Given Signal $f(t)$
\State Initialize $\hat{u}_k^1$, $\hat{\omega}_k^1$, $\hat{\lambda}_k^1$, $K$, $n=0$ 
\\
\State $\hat{f}(\omega)$=Pre-processing($f(t)$),
\State\textbf{repeat}
 \State n=n+1
\For{k $= 1, K$}
    \State Update $\hat{u}_k^1$ for all $\omega \ge 0$: 
    \\
   \State $\hat{u}_k^{(n+1)}=\frac{\hat{f}(\omega)-\sum\limits_{i< k} \hat{u_i}^{n+1}(\omega)-\sum\limits_{i > k}\hat{u_i}^{n}(\omega)+\frac{\hat{\lambda}^n(\omega)}{2}}{1+2\alpha(\omega-\omega^n_k)^2}$
    \\
    \State Update $\omega_k:$
    \\
   \State $\omega_k^{n+1}=\frac{\sum\limits_{\omega=T}^{2T} \omega |\hat{u}_k^{n+1}(\omega)|^2}{\sum\limits_{\omega=T}^{2T} |\hat{u}_k^{n+1}(\omega)|^2}$
\\
\EndFor
\\
\State $\hat{\lambda}^{n+1}(\omega) = \hat{\lambda}^{n}(\omega) + \tau (\hat{f}(\omega)-\sum_k \hat{u}_k^{n+1}(\omega))$
\\
\State \textbf{until} convergence: $ \sum\limits_{k} \frac{\lVert \hat{u}_k^{n+1}- \hat{u}_k^n \rVert^2_2}{\lVert \hat{u}_k^n\rVert^2_2}<\epsilon$
\\
\State $u_k(t)$=Post-processing($\hat{u}_k(\omega)$)
\\
\Return $u_k(t)$ 
\end{algorithmic}
\end{algorithm}

\begin{figure*}[!t]
  \includegraphics[width=1.0\textwidth]{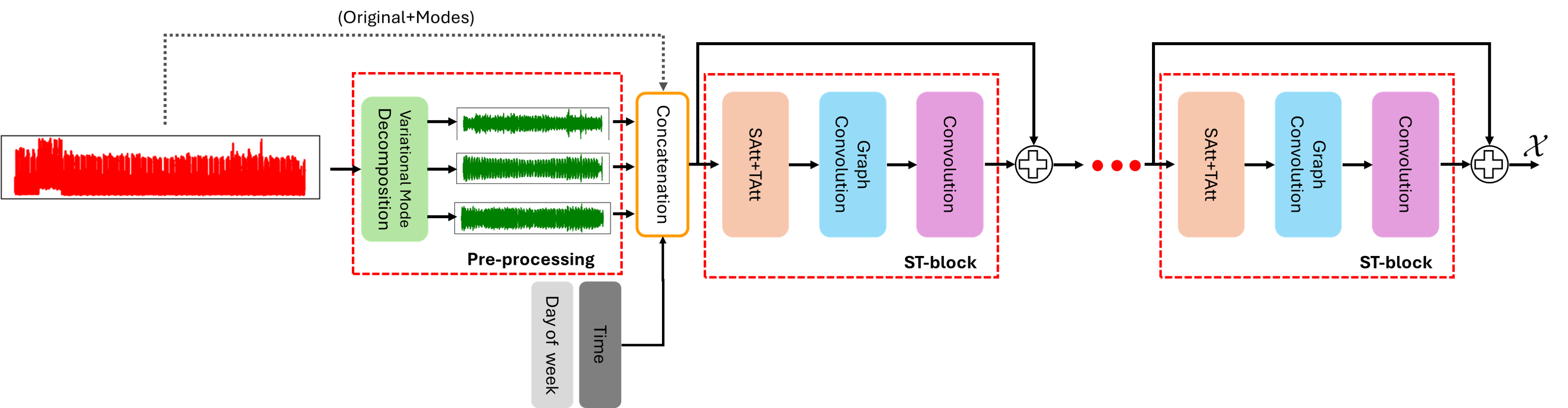} 
  \caption{Block diagram of the variational mode graph convolutional network (VMGCN). The data is decomposed into components using VMD which are then concatenated as the features along the time of the day and day of the week. The original signal can also be concatenated in architecture variants and pass these modified features graph into our attention-augmented GCN. A spatiotemporal (ST) block applies SAtt: Spatial attention, TAtt: temporal attention, graph convolution, and time convolution. The prediction outcome $\mathcal{X} \in \mathbb{R}^{N \times T_w}$ of VMGCN contains the future traffic flow from the timestamp of ($T_w+1$) to ($T_w+N_H$). }
  \label{fig:block_diagram} 
\end{figure*}
\subsubsection{Attention-Based Spatial-Temporal Graph Convolutional Networks (ASTGCN)}The input graph feature within a time window $T_w$ is defined as the $\mathbf{\mathcal{X}}=[\mathbf{X}_{t-T_w+1},\mathbf{X}_{t-T_w+2},\dots,\mathbf{X}_{t}] \in \mathbb{R}^{N \times d \times T_w}$, where $\mathbf{X}_1 \in \mathbb{R}^{d \times T_w}$ is the feature matrix for the graph at $t=1$. The number of feature channels is denoted by $d$. The spatial attention on the graph feature matrix is given by 
\begin{equation}
\mathbf{S}=\mathbf{V}_s\sigma((\mathbf{\mathcal{X}}\mathbf{W}_1\mathbf{W}_2(\mathbf{W}_3\mathbf{\mathcal{X}})^\textit{T})+\mathbf{b}_s),
\end{equation}
\begin{equation}
\mathbf{S}^{'}_{i,j}=\frac{\exp(\mathbf{S}_{i,j})}{\sum\limits_{j=1}^{N} \exp(\mathbf{S}_{i,j})},
\end{equation}
where $\mathbf{V}_s,\: \mathbf{b}_s \in \mathbb{R}^{N \times N}, \:\mathbf{W}_1\in \mathbb{R}^{T_w},\: \mathbf{W}_2 \in \mathbb{R}^{d \times T_w}, \: \mathbf{W}_3 \in \mathbb{R}^{d}$ are learnable weights characterizing the spatial attention. $\sigma$ is the sigmoid activation function. $\mathbf{S}^{'}_{i,j}$ is the normalized spatial correlation matrix between node $i$ and node $j$, which is dynamically computed. The temporal attention is described as 
\begin{equation}
\mathbf{E}=\mathbf{V}_e\sigma(((\mathbf{\mathcal{X}}^\textit{T}\mathbf{U}_1)\mathbf{U}_2(\mathbf{U}_3\mathbf{\mathcal{X}}))+\mathbf{b}_e),
\end{equation}
\begin{equation}
\mathbf{E}^{'}_{i,j}=\frac{\exp(\mathbf{E}_{i,j})}{\sum\limits_{j=1}^{T_w} \exp(\mathbf{E}_{i,j})},
\end{equation}
where $\mathbf{V}_e,\: \mathbf{b}_e \in \mathbb{R}^{T_w \times T_w}, \:\mathbf{U}_1 \in \mathbb{R}^{N},\: \mathbf{U}_2 \in \mathbb{R}^{N \times d}, \: \mathbf{U}_3 \in \mathbb{R}^{d}$ are trainable parameters to determine the temporal attention. $\mathbf{E}^{'}_{i,j}$ is the normalized temporal correlation matrix between node $i$ and node $j$. In spectral graph analysis, graph connectivity is examined using the properties associated with the adjacency matrix or Laplacian matrix. The binary adjacency matrix for the weighted graph is defined by using a Gaussian kernel thresholded function~\cite{shuman2013emerging}
\begin{equation*}
    \mathbf{A}_{i,j} = 
\begin{cases} 
1, & \text{if } \exp(-\frac{l_{i,j}^2}{\sigma^2})  \ge r, \\
0, & \text{otherwise},
\end{cases}
\end{equation*}
\\
where $l_{i,j}$ is the road network distance between node $i$ and $j$, $\sigma$ denotes the standard deviation of distances, and $r$ is the threshold. The Laplacian matrix is defined as $\mathbf{L}=\mathbf{D}-\mathbf{A}$, $\mathbf{D} \in \mathbb{R}^{N \times N}$ is the diagonal matrix consisting of the degree of each node $\mathbf{D}_{ii}=\sum_j  \mathbf{A}_{(i,j)}$, and the normalized Laplacian matrix is expressed as $\mathbf{L}=\mathbf{I}_N-\mathbf{D}^{-\frac{1}{2}}\mathbf{A}\mathbf{D}^{-\frac{1}{2}}$. The eigenvalue decomposition of the Laplacian matrix is $\mathbf{L}=\mathbf{Q}\mathbf{\Lambda}\mathbf{Q}^\textit{T}$, where eigenvalues $\mathbf{\Lambda}={\rm diag}([\lambda_1,\lambda_2,\dots,\lambda_{N-1}])$ is a diagonal matrix. In graph Fourier transform, eigenvectors $\mathbf{Q}$ form a Fourier basis consisting of complex sinusoids, and $\mathbf{Q}$ is an orthogonal matrix.  The kernel $g_\theta$ with the parameters $\theta$ is applied on the Laplacian matrix given by
\begin{equation*}
g_\theta(\mathbf{L})=g_\theta(\mathbf{Q}\mathbf{\Lambda}\mathbf{Q}^\textit{T})=\mathbf{Q}g_\theta(\mathbf{\Lambda})\mathbf{Q}^\textit{T}x.
\end{equation*}
Since the eigenvalue decomposition in spectral graphs is a computationally expensive task, the filtering operation in the graph spectral domain can be formulated using the Chebyshev graph convolution network as~\cite{ChebNet}
\begin{equation}
    g_\theta(\mathbf{L})=\sum_{m=0}^{M-1} \theta_m (T_m(\hat{\mathbf{L}})\odot \mathbf{S}^{'})\mathcal{X},
\end{equation}
where $\theta$ is a Chebyshev coefficients, $T_m(\hat{\mathbf{L}})$ is the Chebyshev polynomial of order $m$, and $\hat{\mathbf{L}}=2\mathbf{L}/\lambda_{\rm {max}}-I_n$. Here, $\lambda_{\rm {max}}$ is the maximum eigenvalue of the Laplacian matrix $\mathbf{L}$.

\subsection{Model Architecture}
The model architecture, as illustrated in Fig.~\ref{fig:block_diagram}, consists of two steps: pre-processing and learning. In the first stage of this architecture, the time series signal of each node is decomposed into $K$ modes or components. The number of modes $K$ is a hyper-parameter that must be adjusted based on the nature of the dataset. After recursively decomposing the signal, the decomposed components are concatenated along with $d^{'}$ number of additional features such as time of the day and day of the week. The new features of the graph are formed into $\mathcal{X} \in  \mathbb{R}^{N  \times(K+d^{'}) \times T_w}$.

\begin{figure}[!t]
\centering
  \includegraphics[width=0.80\textwidth]{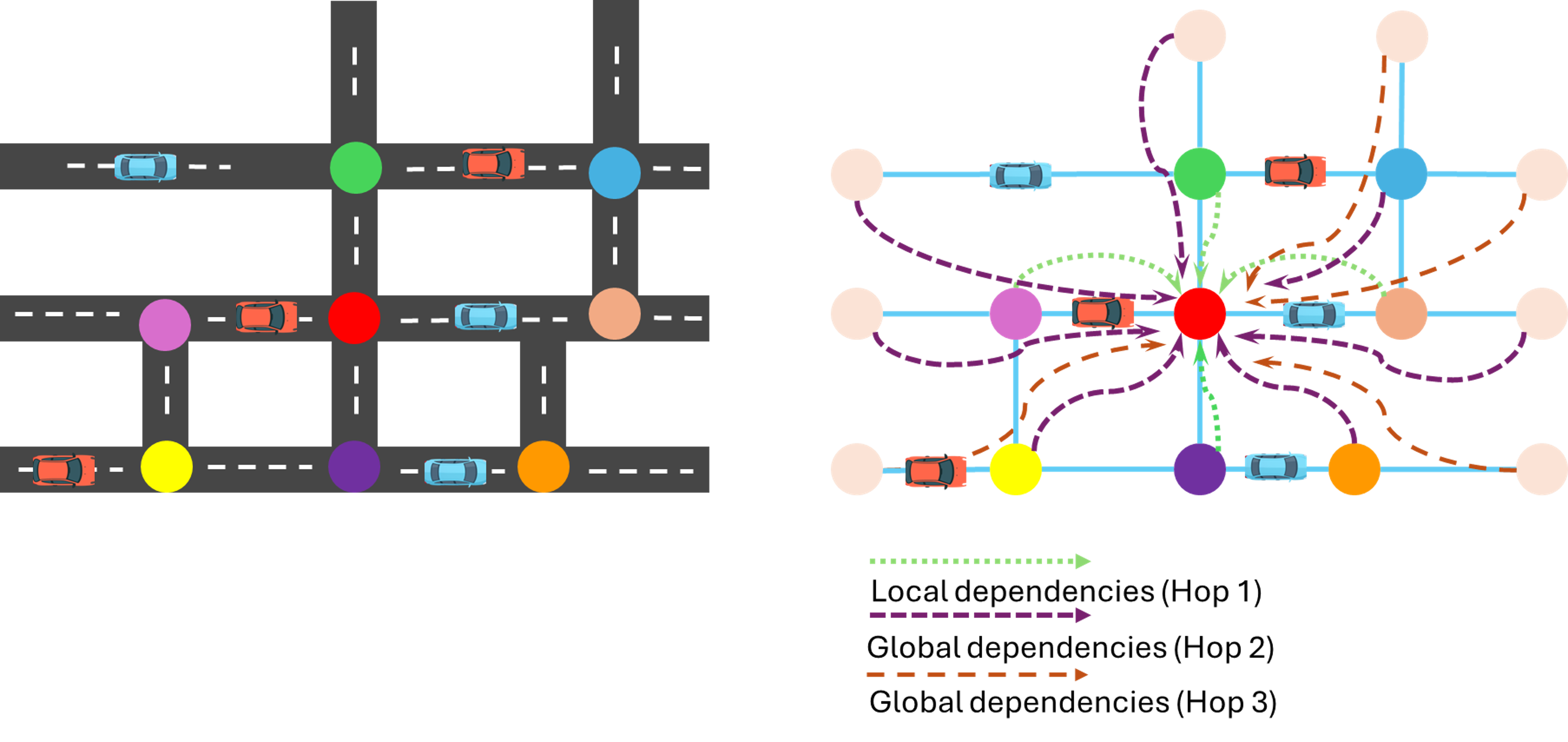} 
  \caption{Left picture depicts the mechanism to count the vehicles at each intersection. The right picture explains the local and global dependencies, where global dependencies are learned through a graph network in each hop.}
  \label{fig:road_graph} 
\end{figure}

The spatiotemporal (ST) block comprises spatial and temporal attention, graph convolution, and time convolution. Spatial and temporal attention mechanisms are applied to capture spatiotemporal correlations. Graph neural networks (GNNs) analyze the topological attributes of the graph, and Chebyshev graph convolution (ChebNet) captures the properties of the graph for the Laplacian matrix in the spectral domain. It is essential to capture global dependencies to incorporate the structural information of a road network, as illustrated in Fig.~\ref{fig:road_graph}. By increasing the order 
$M$ of the Chebyshev polynomial, the model aggregates information from nodes up to $M$ hops away. Time convolution is employed to capture the temporal relationships within the graph structure. A residual connection combines the original feature with the output of the ST block, and a 2D convolution is applied to its outcome. Multiple ST blocks are stacked together to obtain the final predicted outcome from the model. 
\section{Experiments}
\label{sec:result}
We perform our analysis for the following three cases:
\begin{itemize}
    \item VMGCN-v1: Features include original signal, modes, and time information (time of day, day of week).
    \item VMGCN-v2: Features include modes and time information.
    \item VMGCN-v3: Features include redemption function $\phi (t)$, modes, and time information. 
\end{itemize}
We compute the mean absolute percentage error (MAPE), mean absolute error (MAE), and root mean square error (RMSE) metrics to evaluate the performance of the proposed network. Before analyzing the performance of our proposed model, we first provide a brief description of the dataset and the baseline models.
\subsection{Datasets and Implementation Details}
\subsubsection*{LargeST Dataset} We use LargeST~\cite{LargeST} dataset for comparing the performance of VMGCN with the existing models. It consists of a total of $8600$ nodes that are divided into four categories based on location: Greater Los Angeles (GLA), Greater Bay Areas (GBA), San Diego (SD), and others. It contains the traffic counts with a time interval of 5 minutes from 2017-2021. It also comprises meta-features such as longitude, latitude, and number of lanes. We summarize the number of edges, nodes, the average degree of each node, and the total data points (in billions, B) in the LargeST dataset in Table I.
\subsubsection*{Implementation Details} The experiments are performed on a Linux system equipped with an Intel(R) i9 24 GB RAM and NVIDIA 3080Ti GPU. The model is trained for a one-year duration of $2019$ where the samples are aggregated with a sample size of 15 minutes. The training, validation, and test data is distributed as 60$\%$, $20\%$, $20\%$, respectively. The input horizon ($T_w$) is $12$ and the horizon for prediction ($N_H$) is also $12$. In this work, we are predicting the long-term traffic for the next $3$ hours. We consider the short-term prediction up to 1 hour (till horizon 4) and the long-term predictions beyond 1 hour (horizon greater than 4). The proposed methods are implemented using PyTorch and use vmdpy tool for decomposition~\cite{vmdpy}. The batch size for the SD region is 64, the GLA region is 8, and the GBA region is 4. Each model is trained using the Adam optimizer for 100 epochs with early stopping criteria.
\begin{table}[!t]
\centering
\caption{LargeST dataset summary.}
 \label{tab: dataset} 
\resizebox{.45\textwidth}{!}{
\begin{tabular}{|c|c|c|c|c|c|}
\hline
\textbf{Source} & \textbf{Dataset} & \textbf{Nodes} & \textbf{Edges} & \textbf{Degree}  & \textbf{Data Points} \\
\hline
 \multirow{4}{*}{LargeST} 
& \textbf{CA} & $\mathbf{8,600}$ & $\mathbf{201,363}$ & $\mathbf{23.4}$ & $\mathbf{4.52B}$ \\
& \textbf{GLA} & $\mathbf{3,834}$ & $\mathbf{98,703}$ & $\mathbf{25.7}$ & $\mathbf{2.02B}$ \\
  & \textbf{GBA} & $\mathbf{2,352}$ & $\mathbf{61,246}$ & $\mathbf{26.0}$  & $\mathbf{1.24B}$ \\
 & \textbf{SD} & $\mathbf{716}$ & $\mathbf{17,319}$ & $\mathbf{24.2}$  & $\mathbf{0.38B}$ \\
\hline
\end{tabular}
}
\end{table}

\subsection{Baseline Models}
\begin{itemize}
    \item Historical last (HL)~\cite{historical_last} is a naive method that makes predictions based on the last observation. 
     \item Long short-term memory (LSTM)~\cite{LSTM}, is a variant of RNN to capture large temporal dependencies.
     \item Diffusion convolutional recurrent neural networks (DCRNN)~\cite{DCRNN_MetaLA} captures the spatial and temporal pattern using diffusion convolution and encoding-decoding architecture, respectively. 
     \item Adaptive graph convolutional recurrent networks (AGCRN)~\cite{AGCRN} determines the node-specific spatial and temporal correlation in traffic series. 
     \item Graph WaveNet (GWNET)~\cite{GWNET} is a GNN architecture for spatial-temporal modeling.
     \item Spatial-temporal graph ode networks (STGODE)~\cite{STGODE} extracts the dynamical pattern from traffic data using ordinary differential equation (ODE).
     \item Dynamic spatial-temporal aware graph neural networks (DSTAGNN)~\cite{DSTAGNN} constructs the data-driven graph and represents the spatial relevance using multi-head attention mechanism and temporal via gated convolution.
     \item Dynamic graph convolutional recurrent networks (DGCRN)~\cite{DGCRN} employs a mechanism to generate the dynamical topology of a graph for traffic prediction.
     \item Decoupled dynamic spatial-temporal graph neural networks (D$^{2}$STGNN)~\cite{d2stgnn} incorporates the decoupling of the traffic data in the learning mechanism. 
     \item  Attention-based spatial-temporal graph convolutional networks (ASTGCN)~\cite{ASTGCN} introduces the attention mechanism and applies the graph convolution and temporal convolution in the modeling of ST data. 
     \item Spatio-Temporal Wavelets (STWave)~\cite{fang2023spatio} employed the discrete wavelet transform (DWT) to convert the traffic time series into low-frequency and multi-high-frequency components, then use their inverse DWT (IDWT) representation. Dual-channel ST encoder uses temporal attention and causal convolution to learn time dependencies, while spectral graph attention networks have been incorporated to capture spatial dynamics. And at the end, two decoders forecast the future traffic trends.
     \item BigST~\cite{han2024bigst} pre-processes the long-term historical features by using a generative pre-training method. This framework also quantifies the spatial dependencies using attention scores to construct the graph structures.
     \item Random projection mixer (RPMixer)~\cite{yeh2024rpmixer} converts the time series data into the frequency domain using Fast Fourier Transform (FFT) and passes each real and imaginary component to a linear layer. Random projection layers have been used to enhance the spatial modeling.
     \item Patch spatio-temporal graph (PatchSTG)~\cite{fang2024efficient} implements the irregular spatial patching by utilizing the leaf K-dimensional tree (KDTree) to capture the spatial knowledge dynamically.
     \item Regularized adaptive graph learning (RAGL)~\cite{wu2025regularized} achieves embedded regularization and noise suppression by introducing the stochastic shared embedding and an adaptive graph convolution (AGC) encoder in their framework to forecast the future states.
\end{itemize}
\subsection{Analysis}
\begin{figure*}[!t]
    \centering
    \begin{subfigure}[b]{0.49\textwidth} 
        \centering
        \includegraphics[width=\textwidth]{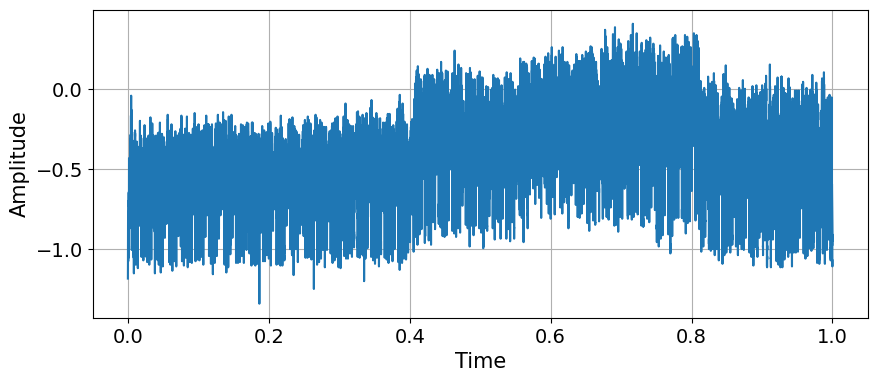}
        \caption{$f(t)$}
        \label{fig:signal}
    \end{subfigure}
    \hfill 
    \begin{subfigure}[b]{0.49\textwidth} 
        \centering
        \includegraphics[width=\textwidth]{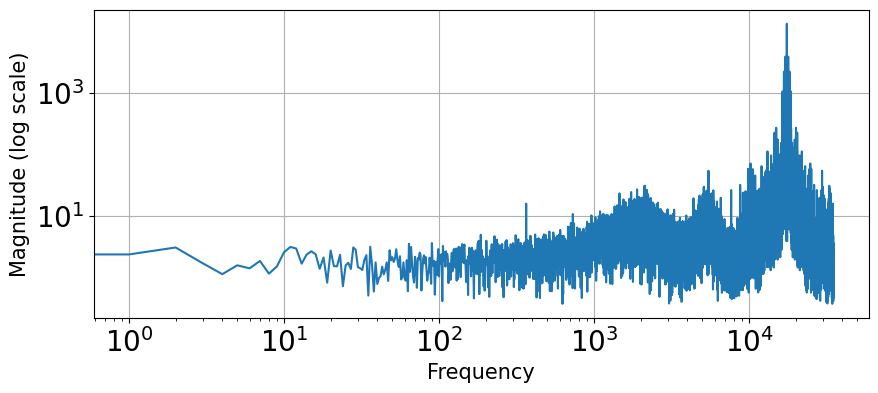}
        \caption{$ \sum\limits_{i=1}^{8} \lvert \hat{u_i}(\omega) \rvert$}
        \label{fig:frequency}
    \end{subfigure}
    
    \hfill 
    \begin{subfigure}[b]{1.0\textwidth} 
        \centering
        \includegraphics[width=\textwidth]{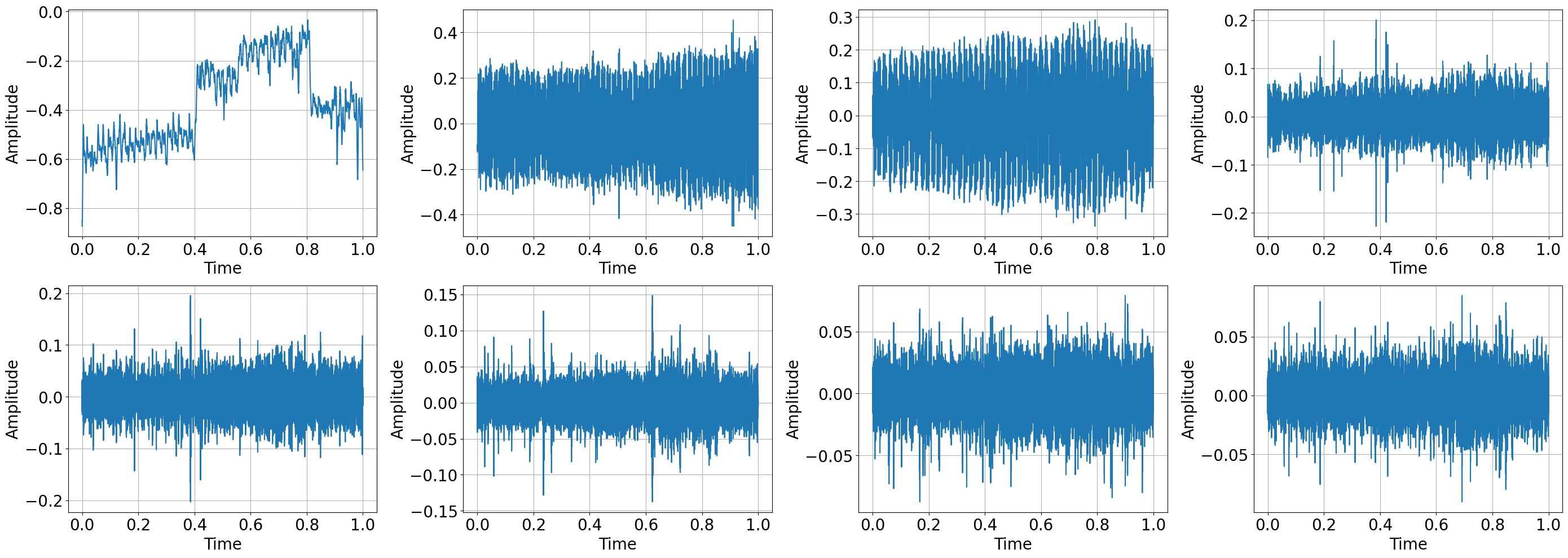}
        \caption{VMD $u_k(t)$}
        \label{fig:vmd_components}
    \end{subfigure}
    \hfill
    \begin{subfigure}[b]{0.46\textwidth}
        \centering
        \includegraphics[width=\textwidth]{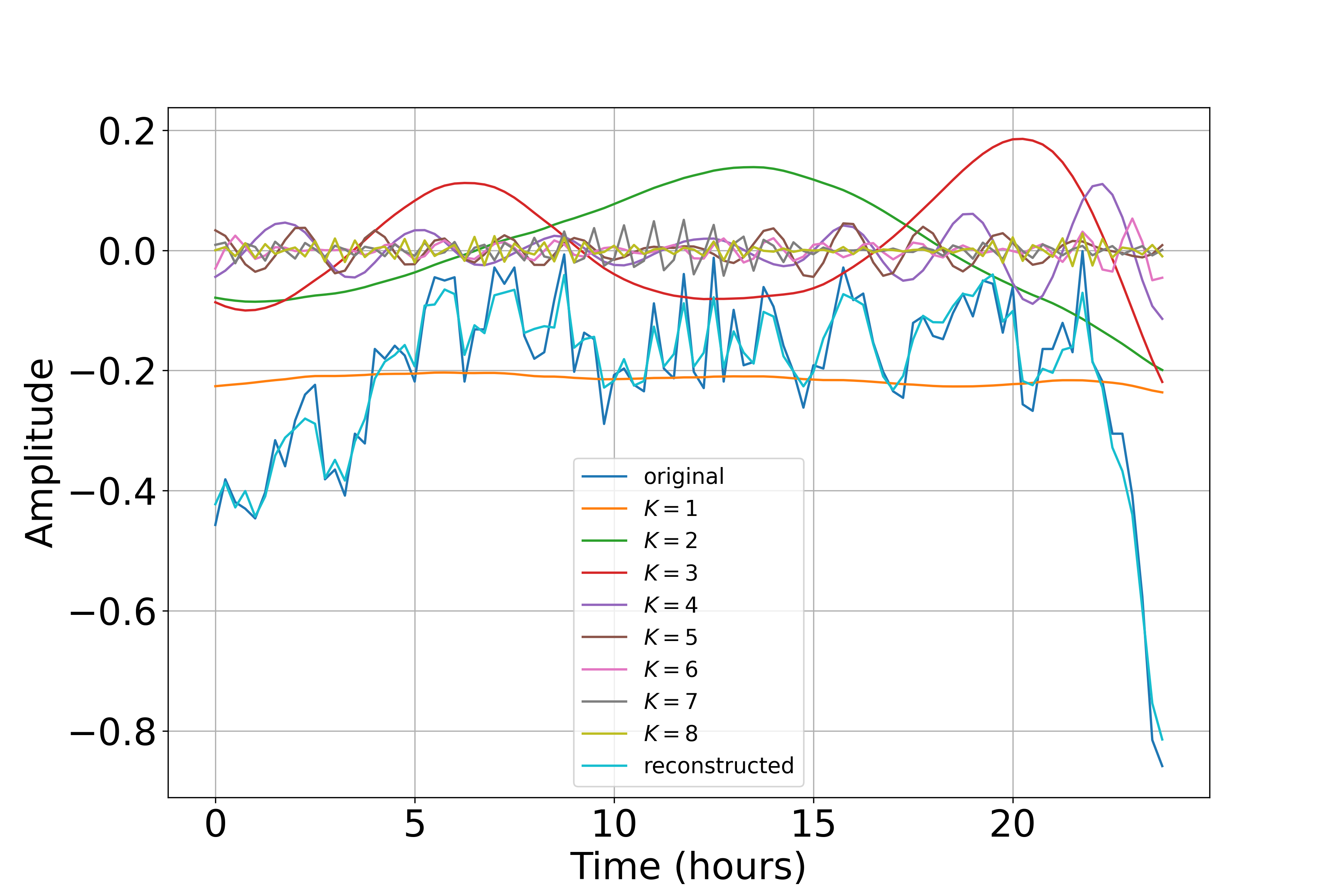}
        \caption{Reconstructed signal (weekday)}
        \label{fig:modes_reconstruction_weekdays}
    \end{subfigure}
    \begin{subfigure}[b]{0.47\textwidth} 
        \centering
        \includegraphics[width=\textwidth]{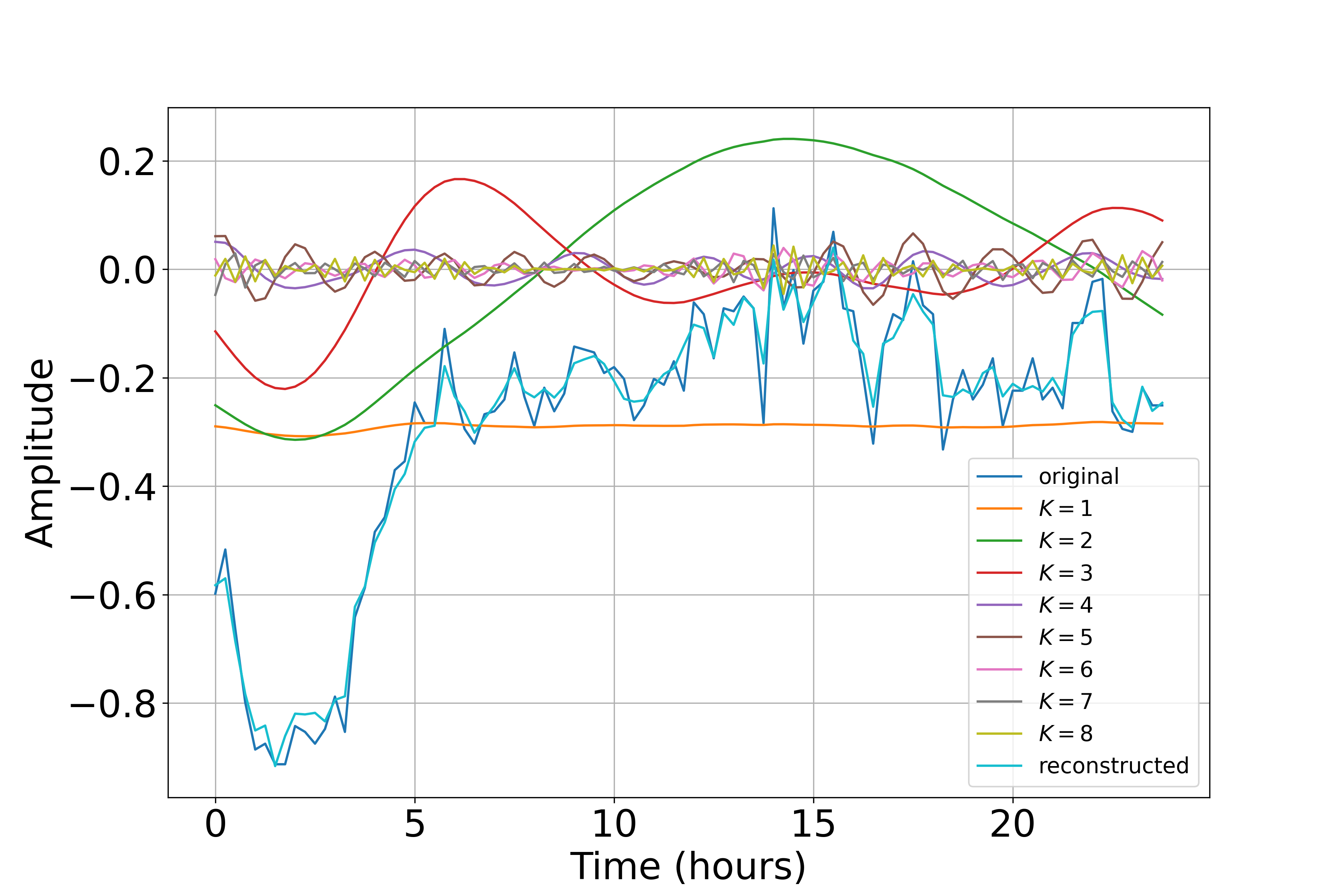}
        \caption{Reconstructed signal (weekend)}
        \label{fig:modes_reconstruction_weekends}
    \end{subfigure}
    
    \caption{Illustration of traffic signal decomposition and reconstruction. (a) The normalized traffic signal recorded by the sensor, and (b) the magnitude spectrum of the reconstructed signal from its modes, both plotted on normalized scales (0-1) with respect to time and frequency, respectively. (c) IMF1 - IMF8 plotted from lowest to highest-frequency components from top to bottom, left to right order (d) the reconstructed signal from modes for a day, weekday (16$^{\text{th}}$ June, 2019), and (e) the reconstructed signal from modes, weekend (14$^{\text{th}}$ June, 2019).}
    \label{fig:traffic_signal}
\end{figure*}

Variational Mode Decomposition (VMD) recursively decomposes traffic data into modes that can be categorized as high-frequency, intermediate-frequency, and low-frequency components. These modes capture the real-time, non-stationary, and complex nature of traffic signals, which typically consist of multiple overlapping patterns, as shown in Fig.~\ref{fig:traffic_signal}. Fig.~\ref{fig:signal} displays the counts from a 1-year (on a scale 0 to 1) traffic sensor in the SD region. The signal spectrum, illustrated in Fig.~\ref{fig:frequency}, reveals the presence of periodic, quasi-periodic, and noise components. The signal is decomposed into eight Intrinsic Mode Functions (IMFs), labeled IMF1 to IMF8 and ordered from low to high frequency, as shown in Fig.~\ref{fig:vmd_components}. These modes allow for physical interpretation in the context of real-time traffic analysis. IMF1 represents the overall trend of the original signal, capturing its low-frequency behavior, while IMF8 captures the high-frequency elements or noise. The intermediate modes (IMF2 to IMF7) capture various degrees of temporal patterns depending on the complexity of the original signal. A denoised version of the signal can be reconstructed by summing selected IMFs, typically excluding the lowest and highest frequency components, as outlined in~\cite{VMD}. Fig.~\ref{fig:modes_reconstruction_weekdays} and Fig.~\ref{fig:modes_reconstruction_weekends} provide a detailed view of the decomposed modes and the reconstructed signals over a 24-hour period on a weekday (Monday) and a weekend (Saturday), respectively. These plots show that IMF1 acts as the baseline trend, while IMF2 and IMF3 capture the envelope and overall structure of the signal. Higher-frequency IMFs account for short-term fluctuations. Notably, during early morning hours (1:00~AM to 3:00~AM), traffic volume is significantly lower on weekends compared to weekdays. This pattern has been successfully captured by IMF2 and IMF3. Each mode effectively represents aspects of traffic flow and human activity patterns, which can ultimately enhance the ability of neural networks to learn trends and make accurate traffic flow predictions. 

\begin{figure}[!t]
\centering
\begin{subfigure}[b]{0.52\textwidth}
    \includegraphics[width=\textwidth]{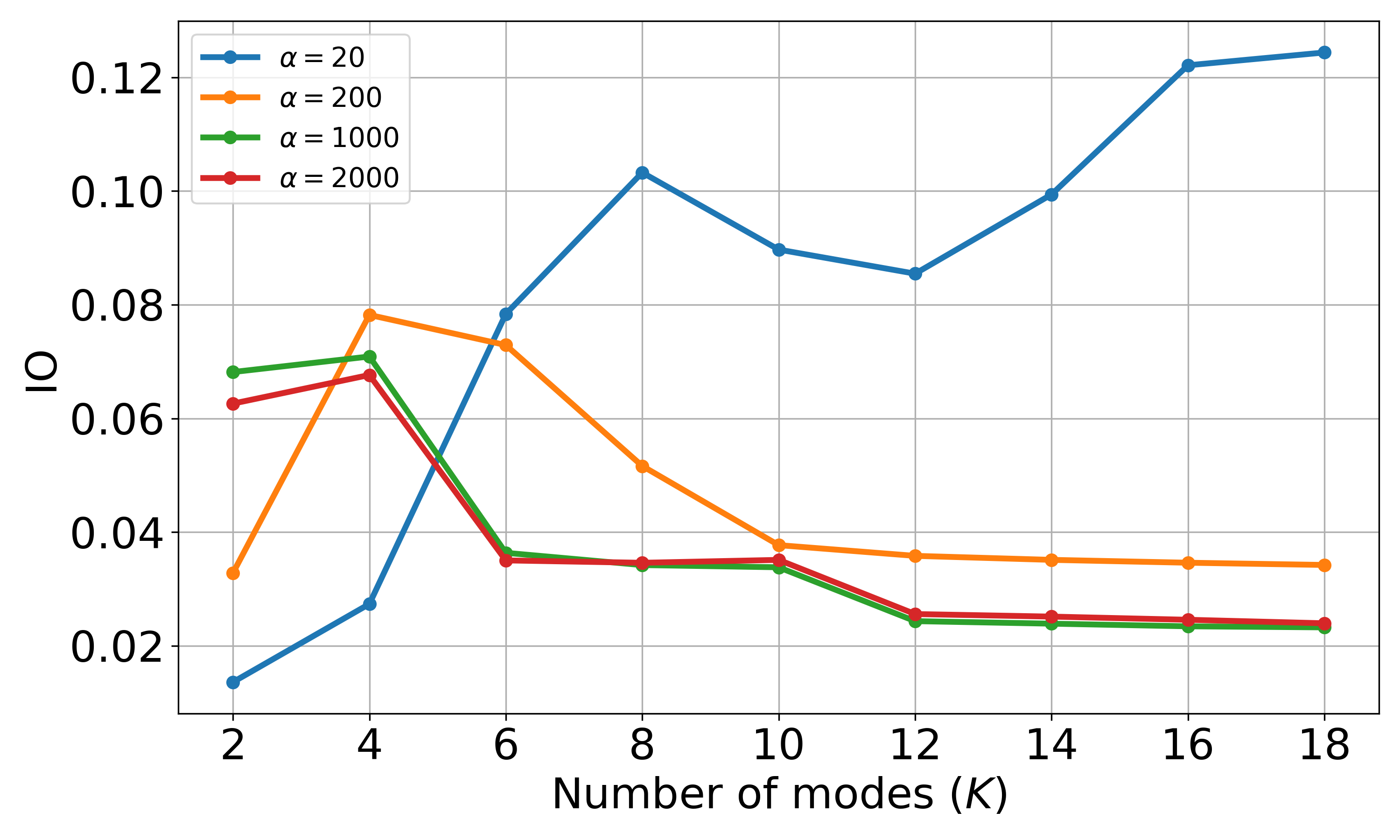} 
      \caption{}
  \label{fig:IO_alpha} 
\end{subfigure}
\hfill
 \begin{subfigure}[b]{0.48\textwidth} 
        \centering
        \includegraphics[width=\textwidth]{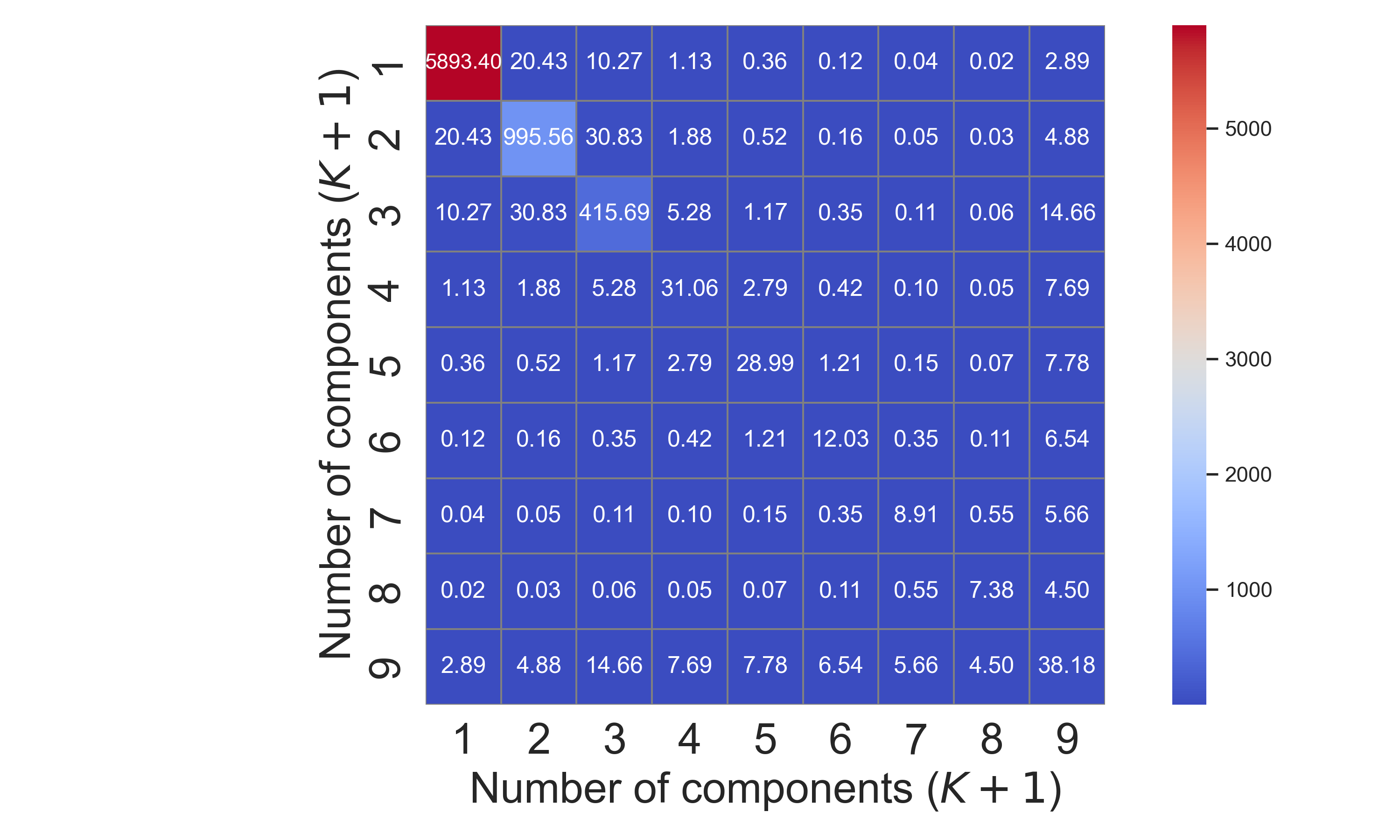}
        \caption{}
        \label{fig:IO_corr_map_unnormalized}
    \end{subfigure}
   \begin{subfigure}[b]{0.48\textwidth} 
        \centering
        \includegraphics[width=\textwidth]{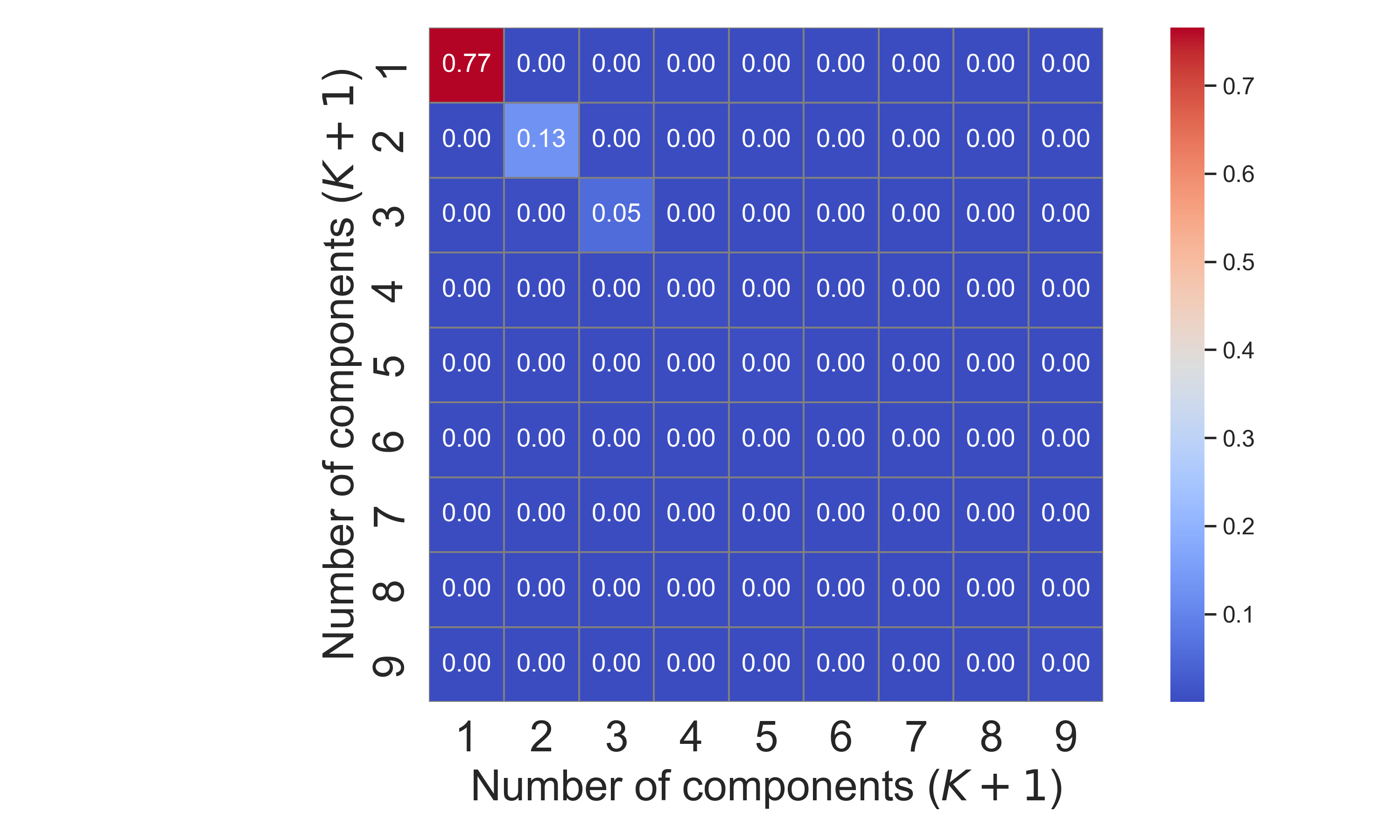}
        \caption{}
        \label{fig:IO_corr_map}
    \end{subfigure}
  \caption{Index of orthogonality on a traffic signal of the SD region. (a) shows the variation in IO with different $\alpha$ and $K$. Visualization of (b)  $\sum_{t=0}^{T}u_i(t) \, u_j(t)$, and (c) $\frac{\sum_{t=0}^{T}u_i(t) \, u_j(t)}{\sum_{t=0}^{T}f^2(t)}$ among decomposed components for $K=8$ and $\alpha=2000$.}
  \label{fig:IO} 
\end{figure}

\begin{figure}[!t]
\centering
  \includegraphics[width=0.46\textwidth]{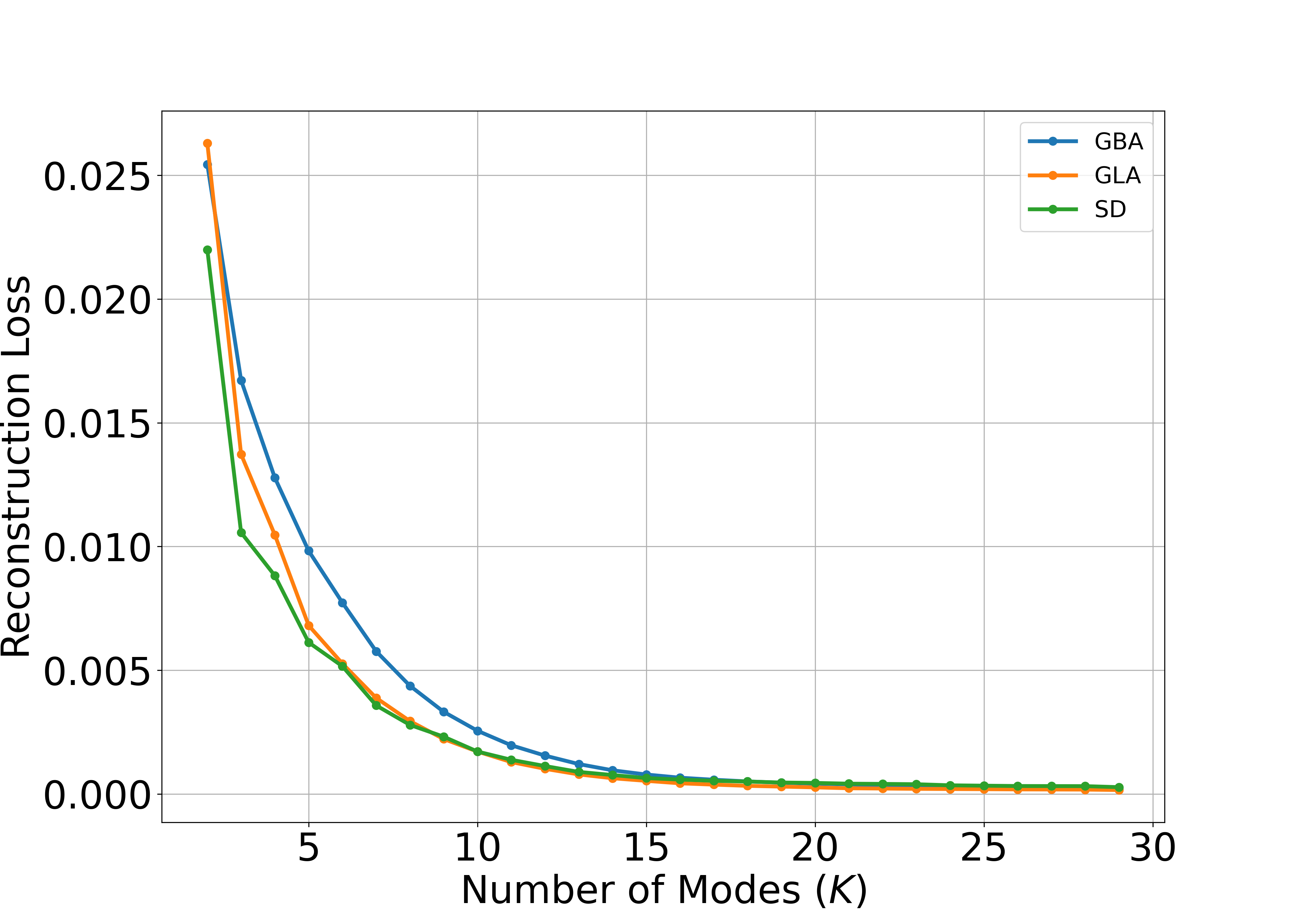} 
  \caption{The optimal value of $K$ is determined by randomly choosing 2\% of the node signals in a region of interest and decomposing these for $K\in[2,29]$. If the mean reconstruction loss is less than a threshold value for a specific value of the mode, we consider that mode as an optimal value of $K$. }
  \label{fig:mode_reconstruction_loss} 
\end{figure}

The tuning of hyperparameters remains a crucial step in solving the optimization process. In the decomposition of data, it is essential to select the appropriate values for the number of modes ($K$), bandwidth constraint ($\alpha$), tolerance for convergence ($\epsilon$), and maximum allowable iterations ($N_m$) for VMD. In this work, we consider three measures to explore the significance of VMD hyperparameters. The index of orthogonality (IO) provides a quantitative measure of the temporal independence between decomposed modes~\cite{emd_hilbert}. A lower IO indicates that the modes are non-orthogonal, better mode separation, minimal redundancy, and improved interpretability, making it an essential quality metric in signal decomposition. Ideally, this metric must be zero for off-diagonal values, but in a practical sense, it would be a non-zero, and it can be measured by 
\begin{figure}[!t]
\centering
\begin{subfigure}[b]{0.46\textwidth}
    \includegraphics[width=\textwidth]{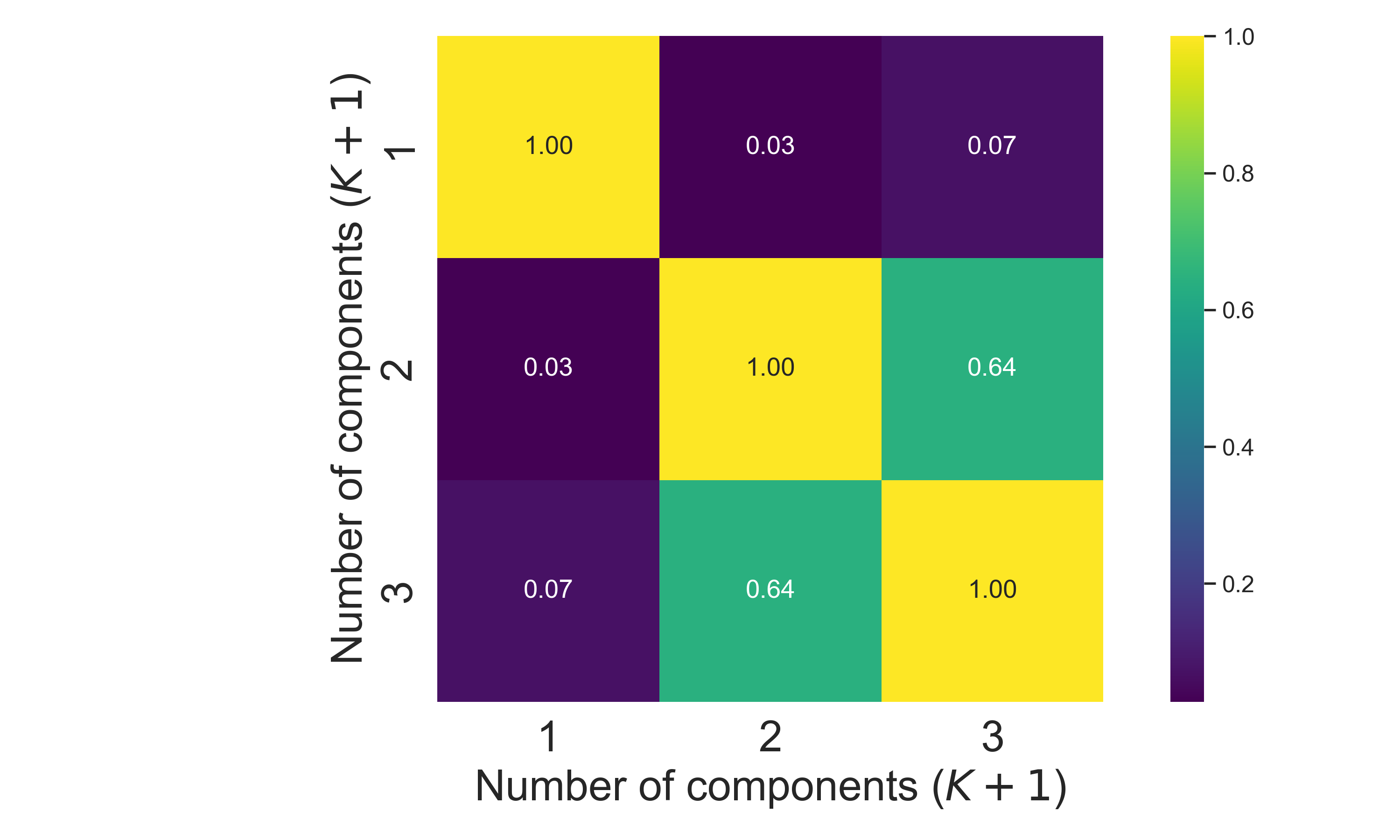} 
      \caption{}
  \label{fig:SOI_20_2} 
\end{subfigure}
   \begin{subfigure}[b]{0.48\textwidth} 
        \centering
        \includegraphics[width=\textwidth]{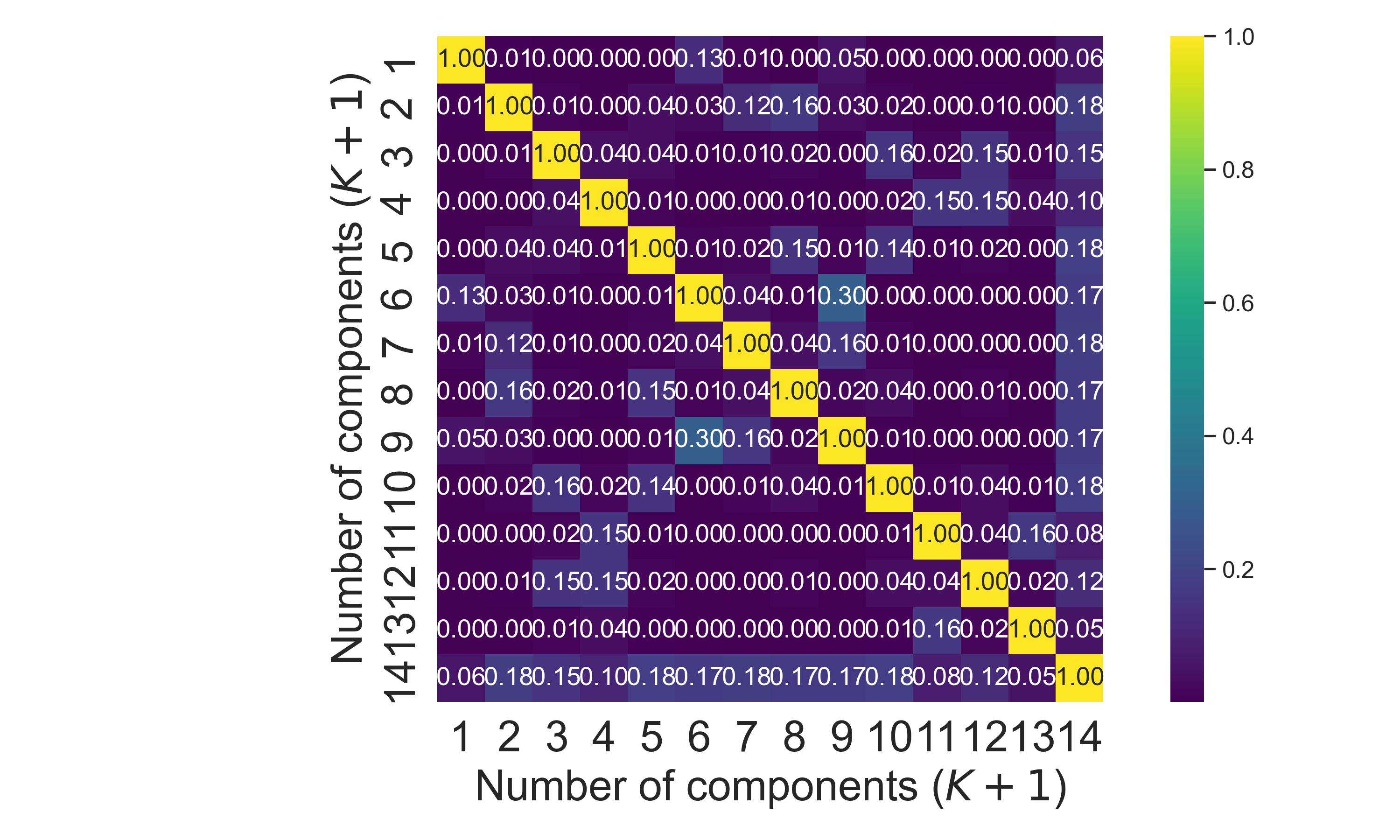}
        \caption{}
        \label{fig:SOI_20_13}
    \end{subfigure}
    \hfill
\begin{subfigure}[b]{0.48\textwidth} 
        \centering
        \includegraphics[width=\textwidth]{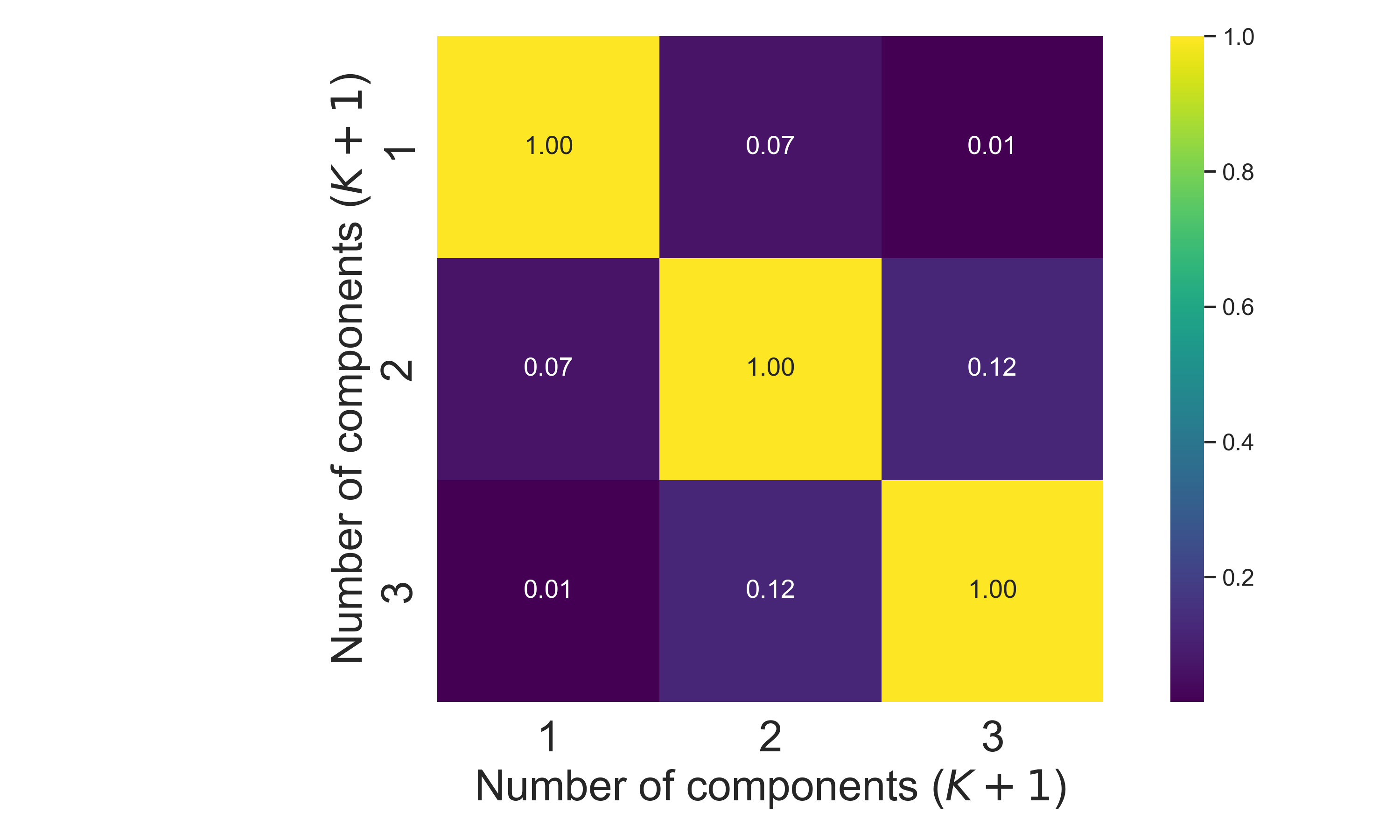}
        \caption{}
        \label{fig:SOI_1000_2}
    \end{subfigure}
\begin{subfigure}[b]{0.48\textwidth} 
        \centering
        \includegraphics[width=\textwidth]{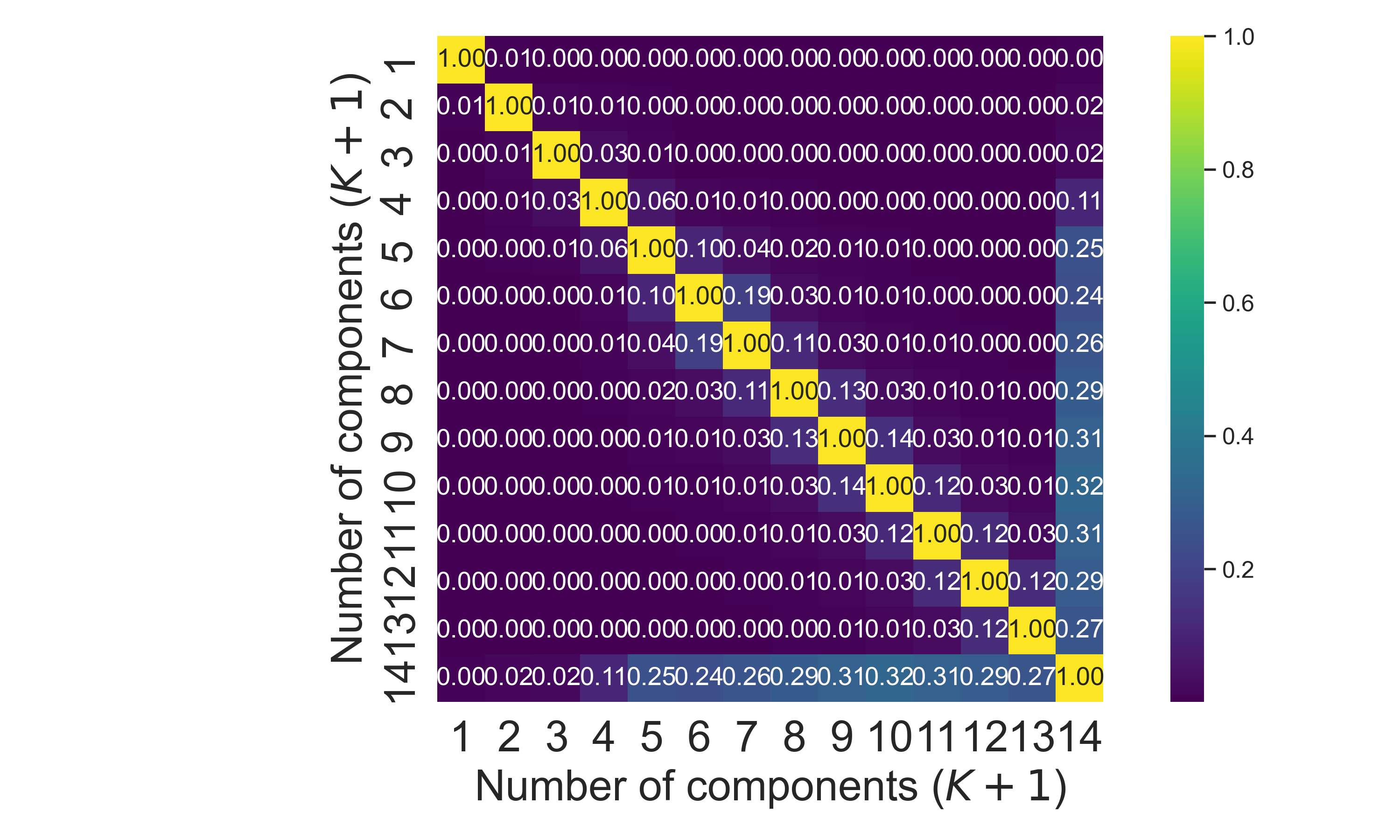}
        \caption{}
        \label{fig:SOI_1000_13}
    \end{subfigure}
\hfill
    \begin{subfigure}[b]{0.48\textwidth} 
        \centering
        \includegraphics[width=\textwidth]{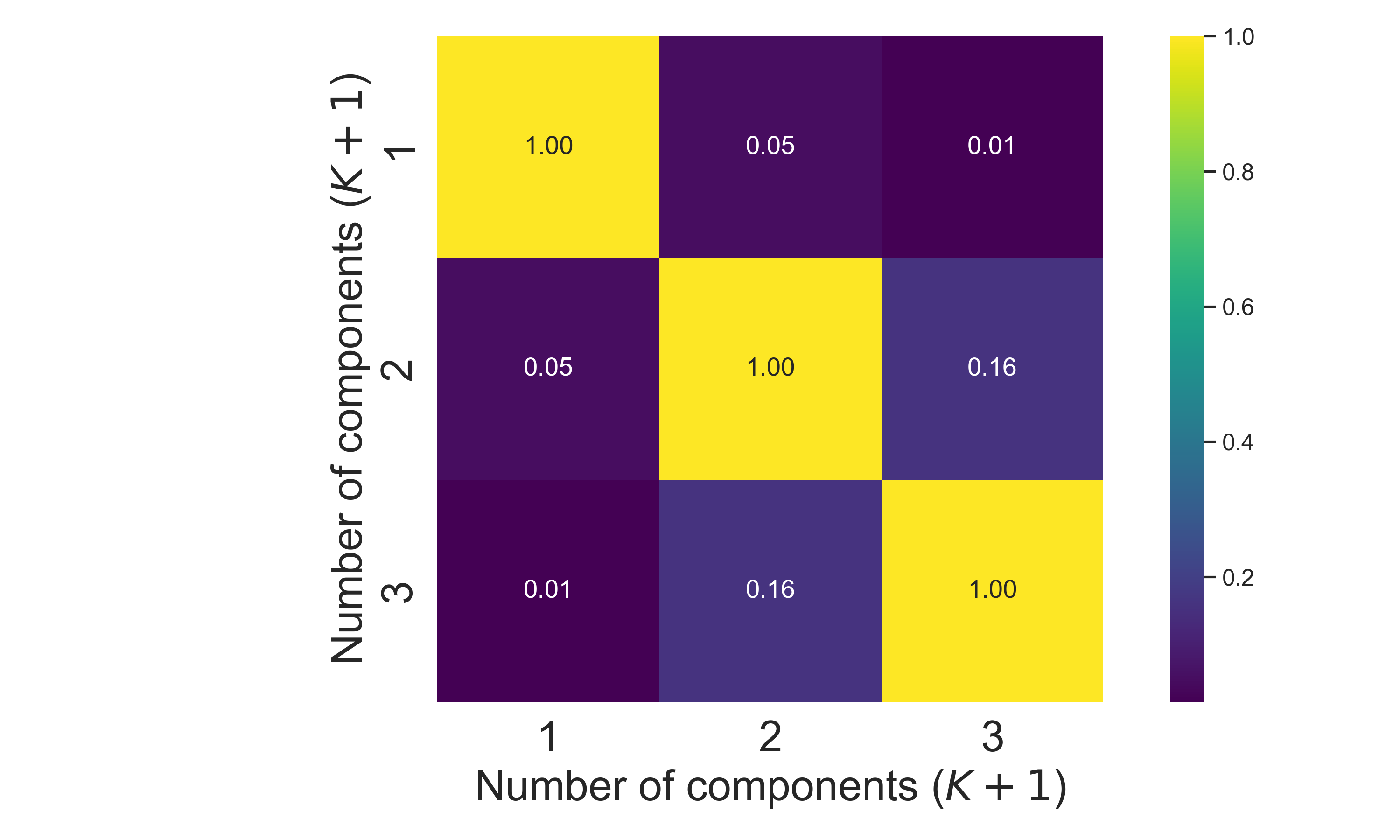}
        \caption{}
        \label{fig:SOI_2000_2}
    \end{subfigure}
    \begin{subfigure}[b]{0.48\textwidth} 
        \centering
        \includegraphics[width=\textwidth]{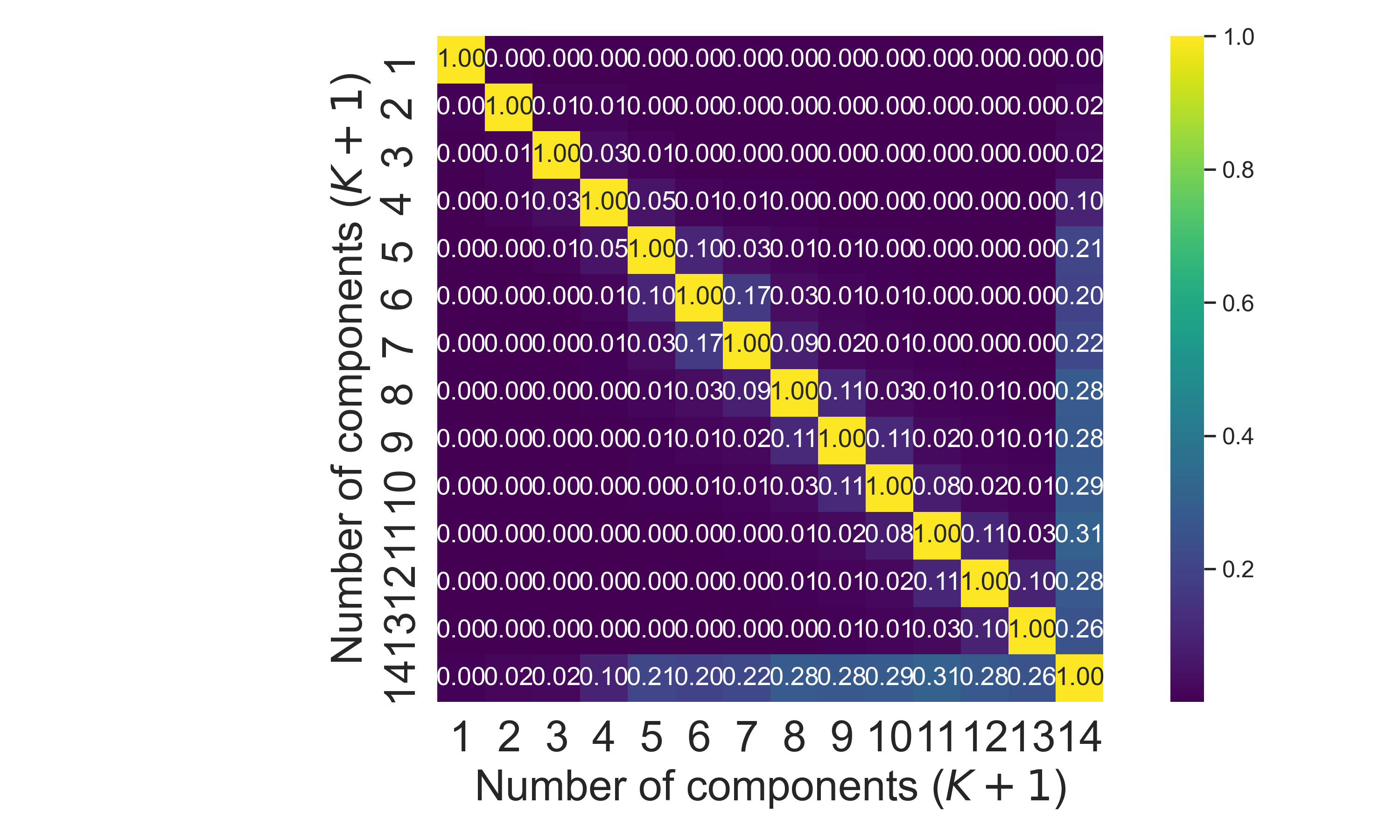}
        \caption{}
        \label{fig:SOI_2000_13}
    \end{subfigure}
  \caption{Spectral overlap index computed on the decomposed modes of a traffic signal of the SD region. (a) $K=2,\, \alpha=20$, (b) $K=13,\, \alpha=20$, (c) $K=2, \, \alpha=1000$, (d) $K=13,\, \alpha=1000$, (e) $K=2,\, \alpha=2000$, and (f) $K=13,\, \alpha=2000$.}
  \label{fig:SOI} 
\end{figure}

\begin{equation}
\text{IO} =  \left( \frac{\sum\limits_{i=1}^{K+1} \sum\limits_{j=1}^{K+1}  \sum\limits_{t=0}^{T}u_i(t) \, u_j(t)}{\sum\limits_{t=0}^{T}f^2(t)} \right),
\end{equation}
where the additional component indicates the redemption or residual function $\phi(t)$. Fig.~\ref{fig:IO_alpha} shows the trend of IO with different bandwidth constraint values ($\alpha$) and number of modes ($K$). Overall, the IO decreases with increasing value of the number of modes with higher $\alpha$ values. However, with a lower value of $\alpha=20$, the IO values keep increasing. Increasing the $\alpha$ decreases the bandwidth and vice versa. For a higher number of modes and a lower value of the bandwidth constraint, the overlaps occur in modes. Higher value of bandwidth constraint and number of modes avoids overlapping because of narrow bandwidth. The average value of IO for $20$ nodes of the SD region is $0.0661$ for $K=13$ and $\alpha=2000$. Fig.~\ref{fig:IO_corr_map_unnormalized} represents the unnormalized heatmap $\sum\limits_{t=0}^{T}u_i(t) \, u_j(t)$ and Fig.~\ref{fig:IO_corr_map} illustrates the normalized IO$_{i,j}$ for $K=8$ and $\alpha=2000$ between the modes. It shows that off-diagonal entities are zero and three components (IMF1-IMF3) share the energy, having a cumulative value of $0.95$, and the remaining components contribute to $0.05$ energy of the total signal. The spectral overlap index (SOI) measures the shared frequency content between decomposed modes. Lower SOI values reflect better spectral separation, indicating that each mode captures a distinct frequency band, essential for interpretable results and can be formulated as
\begin{equation}
\text{SOI}_{i,j} =\frac{ \sum_{\omega}\hat{u}_i(\omega) \,  \hat{u}_j(\omega)}{\sqrt{\sum_{\omega}\hat{u}_i^2(\omega) \sum_{\omega}\hat{u}_j^2(\omega)}},
\end{equation}
here $i,j \in [1 , K+1]$ includes the residual component in frequency domain $\Phi(\omega)$. By increasing $\alpha$, the modes become more spectrally distinct, as indicated in Fig.~\ref{fig:SOI}. This is further evidenced by the decrease in the SOI observed in Fig.~\ref{fig:SOI_2000_2} and Fig.~\ref{fig:SOI_2000_13}, compared to Fig.~\ref{fig:SOI_20_2} and Fig.~\ref{fig:SOI_20_13}, respectively. The difference between Fig.~\ref{fig:SOI_2000_2} and Fig.~\ref{fig:SOI_1000_2} or Fig.~\ref{fig:SOI_2000_13} and Fig.~\ref{fig:SOI_1000_13} is negligible due to higher value of $\alpha$. As we mentioned previously, increasing the number of modes decreases the deformation value; the same trend can be observed in measuring SOI. The frequency overlapping between the decomposed modes and the residual component decreases as $K$ and $\alpha$ increase. The residual is also important to determine suitable hyperparameters, as it is noticed that increasing $K$ and decreasing $\alpha$ results in low reconstruction loss. The small reconstruction loss preserves the information loss from the original signal during decomposition. To determine the optimal value of number of modes, we randomly select nodes, compute the modes for each, and subsequently calculate the average reconstruction loss of the sample data for every $K$ value. If the reconstruction loss is less than the threshold value ($\zeta$), we can consider this $K$ value as a near-best term. In our experimentation, we selected $2\%$ of the nodes from each region and applied this method to values of $K$ ranging from $2$ to $29$, as shown in Fig.~\ref{fig:mode_reconstruction_loss}. The optimal value of $K$ is identified for each region: SD ($K=13$), GBA ($K=14$), and GLA ($K=13$), for which the reconstruction loss remains below the threshold $\zeta = 1 \times 10^{-3}$. After identifying the $K$ value for each region, it is observed that any further decrease in loss beyond these values is negligible. It must be noted that our reconstruction-loss-based mode selection (Fig. ~\ref{fig:mode_reconstruction_loss}) eliminates heuristic tuning that is a common pitfall in EMD-based methods. This data-driven selection of $K$ ensures adaptability across regions. The residual component 
$\phi(t)$ and bandwidth constraint  $\alpha$ explicitly filter sensor noise. This contrasts with wavelet-based methods (e.g., STWave) where noise may alias into multiple frequency bands.

\begin{figure}[!t]
\centering
  \includegraphics[width=0.56\textwidth]{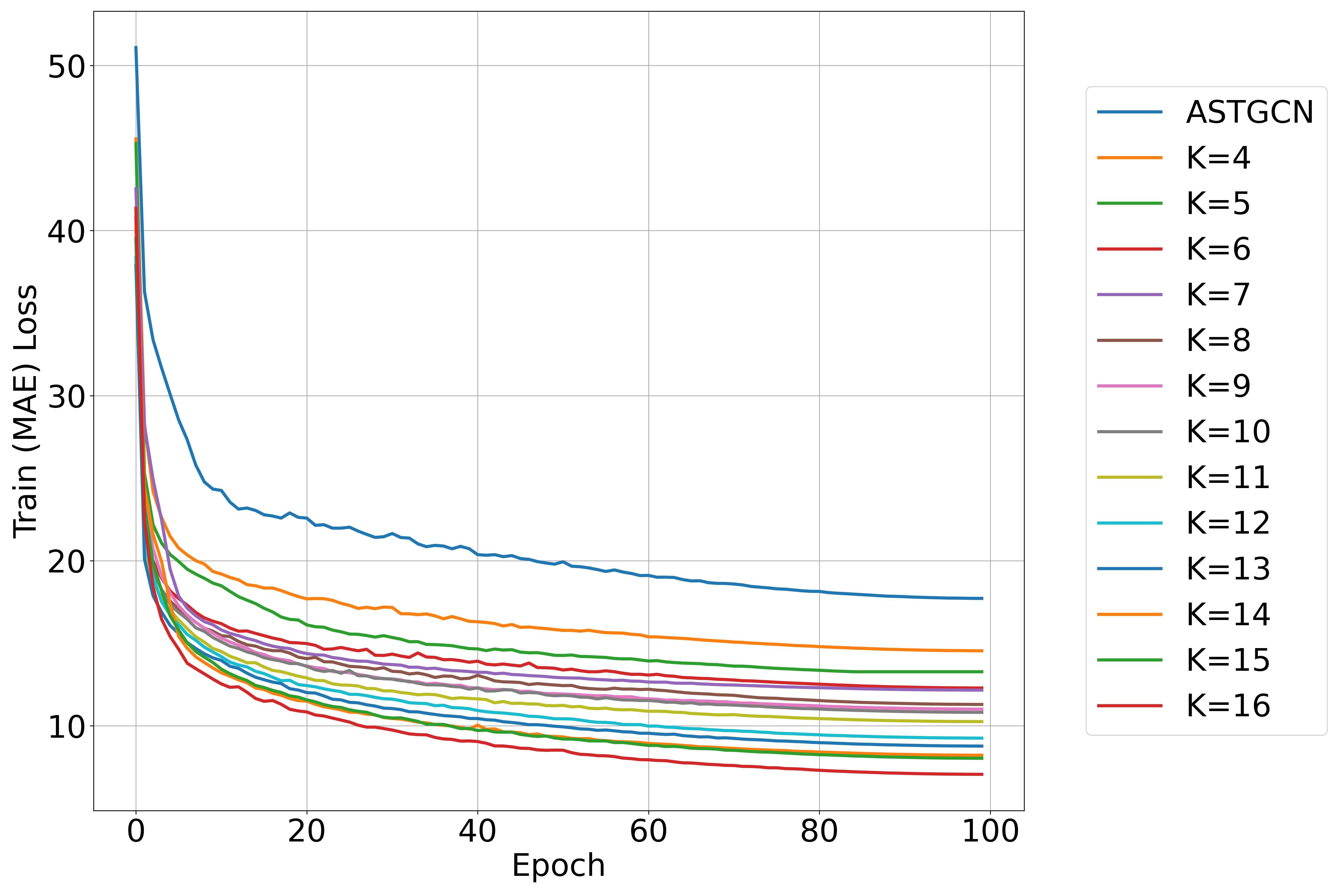} 
  \caption{Training loss with different modes on SD dataset for VMGCN-v1. It shows that the training loss decreases with the increase in the number of modes.}
  \label{fig:training_loss} 
\end{figure}

 The performance of our proposed method has been evaluated under different values of $K$ modes. Fig.~\ref{fig:training_loss} shows the MAE training loss evaluated on the SD region of the LargeST dataset, while keeping the other hyper-parameters constant. It shows that the loss gradually decreases as the number of modes increases. However, finding a suitable value of $K$ that yields the best prediction results is computationally expensive. Consequently, we have formulated a method that utilizes the reconstruction loss to determine the optimal value of $K$. 
The comparison of our experimentation with baseline models is presented in TABLE~\ref{Tab:comparison}. Our model outperforms the signal processing-based decomposition methods that utilize the Wavelet transform and Fourier transform, such as STWave and RPMixer, and representation learning based decomposition methods such as D$^2$STGNN. We evaluated our case scenarios using $\alpha=2000$, $\epsilon=1 \times 10^{-7}$, and $N_m=500$. For VMGCN-v1, incorporating the original feature with the modes generally leads to better predictions. In long-horizon prediction, VMGCN-v3 exhibits relatively lower error values compared to VMGCN-v2. The decomposition by VMD is sensitive to the length of the signal. Currently, utilizing the 1-year data makes the computed modes more effective. Typical traffic trends such as jams, congestion, and daily, or weekly patterns are well-defined in terms of frequencies. Due to the accurate representation of features, the neural networks predict future states. Increasing the number of modes enhances the performance of the model in short-term prediction, but it tends to overfit in long-term predictions. It appears that VMD effectively decomposes the signal at higher modes and removes noise, leading to more precise immediate prediction.

\begin{table*}[!th]
\centering
\caption{Comparison of performance evaluation metrics MAE, MAPE, and RMSE between different advanced methods and our models. The number of parameters (param) is described in K (kilo), $10^3$ and M (million), $10^6$, and the best performance metrics are highlighted in red bold numbers. For VMGCN-v\textcolor{red}{x}-\textcolor{blue}{$K$}, \textcolor{red}{x} describes the variant of the network and \textcolor{blue}{$K$} corresponds to the best value chosen for each region. {\huge \texttimes} parameters count is not publicly available.}
\resizebox{.9\textwidth}{!}{
\begin{threeparttable}
\begin{tabular}{c|cc|ccc|ccc|ccc|ccc}

\toprule
 Dataset & Method & Param & \multicolumn{3}{c}{Horizon 3} & \multicolumn{3}{c}{Horizon 6} & \multicolumn{3}{c}{Horizon 12} & \multicolumn{3}{c}{Average}  \\
\cmidrule(r){4-15}
& & &MAE & RMSE & MAPE& MAE & RMSE& MAPE & MAE & RMSE & MAPE& MAE & RMSE & MAPE \\
\midrule
\multirow{12}{*}{SD} 
& HL\tnote{'} & - & 33.61 & 50.97 & 20.77\% & 57.80 & 84.92 & 37.73\% & 101.74 & 140.14 & 76.84\% & 60.79 & 87.40 & 41.88\% \\
& LSTM\tnote{'} & 98K & 19.17 & 30.75 & 11.85\% & 26.11 & 41.28 & 16.53\% & 38.06 & 59.63 & 25.07\% & 26.73 & 42.14 & 17.17\% \\
& DCRNN\tnote{'} & 373K & 17.01 & 27.33 & 10.96\% & 20.80 & 33.03 & 13.72\% & 26.77 & 42.49 & 18.57\% & 20.86 & 33.13 & 13.94\% \\
& AGCRN\tnote{'} & 761K & 16.05 & 28.78 & 11.74\% & 18.37 & 32.44 & 13.37\% & 22.12 & 40.37 & 16.63\% & 18.43 & 32.97 & 13.51\% \\
& STGCN\tnote{'} & 508K & 18.23 & 30.60 & 13.75\% & 20.34 & 34.42 & 15.10\% & 23.56 & 41.70 & 17.08\% & 20.35 & 34.70 & 15.13\% \\
& GWNET\tnote{'} & 311K & 15.49 & 25.45 & 9.90\% & 18.17 & 30.16 & 11.98\% & 22.18 & 37.82 & 15.41\% & 18.12 & 30.21 & 12.08\% \\
& STGODE\tnote{'} & 729K & 16.76 & 27.26 & 10.95\% & 19.79 & 32.91 & 13.18\% & 23.60 & 41.32 & 16.60\% & 19.52 & 32.76 & 13.22\% \\
& DSTAGNN\tnote{'} & 3.9M  & 17.83 & 28.60 & 11.08\% & 21.95 & 35.37 & 14.55\% & 26.83 & 46.39 & 19.62\% & 21.52 & 35.67 & 14.52\% \\
& DGCRN\tnote{'} & 243K & 15.24 & 25.46 & 10.09\% & 17.66 & 29.65 & 11.77\% & 21.59 & 35.55 & 16.88\% & 17.38 & 28.92 & 12.43\% \\
& D$^2$STGNN\tnote{''} & 406K & 15.76 & 25.71 & 11.84\% & 18.81 & 30.68 & 14.39\% & 23.17 & 38.76 & 18.13\% & 18.71 & 30.77 & 13.99\% \\
& ASTGCN\tnote{''} & 2.15M & 19.68 & 31.53 & 12.20\% & 24.45 & 38.89 & 15.36\% & 31.52 & 49.77 & 22.15\% & 26.07 & 38.42 & 15.63\% \\
&STWave\tnote{'''} &{\Huge \texttimes}&15.80& 25.89& 10.34\%& 18.18 &30.03& 11.96\%& 21.98& 36.99&15.30\%&18.22&30.12&12.20\% \\
&BigST\tnote{'''} &{\Huge \texttimes}&16.42 &26.99 &10.86\%& 18.88 &31.60 &13.24\%& 23.00& 38.59& 15.92\%& 18.80& 31.73&12.91\%\\
&RPMixer~\cite{yeh2024rpmixer}&1.5M&15.12& 24.83 &9.97\% & 17.04  &28.24 & 10.98\%  &19.60 & 32.96 & \textbf{\textcolor{red}{13.12\%}} & 16.90 & 27.97  &11.07\%\\
&PatchSTG\tnote{'''} &2.28M&14.53& 24.34& 9.22 \%&16.86 &28.63 &11.11\%& 20.66 &36.27& 14.72\% &16.90 &29.27 &11.23\%\\
&RAGL~\cite{wu2025regularized} &{\Huge \texttimes}&13.87& 23.42& 9.01\% &16.09& 27.35& 10.63\% &19.90& 33.94& 13.35\% &16.16& 27.40& 10.62\%\\
\cmidrule(r){2-15}
& VMGCN-v1-13\tnote{''} & 2.17M & \textbf{\textcolor{red}{5.09}} & 13.08& \textbf{\textcolor{red}{4.71\%}} & \textbf{\textcolor{red}{8.93}} & 36.31 & 9.13\% & 20.10 & 100.37 & 19.23\% & 11.28& 47.61 &10.21\% \\
& VMGCN-v2-13\tnote{''} & 2.17M & 5.62 & \textbf{\textcolor{red}{9.21}} & 5.19\%& 10.27 & \textbf{\textcolor{red}{17.00}} & \textbf{\textcolor{red}{9.04\%}} & 18.59 & 31.73& 18.15\% & \textbf{\textcolor{red}{10.84}} & 18.26 & 10.10\% \\
& VMGCN-v3-13\tnote{''} & 2.17M &6.20 & 9.80& 5.51\% & 11.22 & 17.66 & 9.08\% & \textbf{\textcolor{red}{18.33}} & \textbf{\textcolor{red}{28.12}} & 15.46\% & 11.29& \textbf{\textcolor{red}{17.65}} & \textbf{\textcolor{red}{9.39\%}} \\
\hline
\hline
\multirow{13}{*}{GBA} 
& HL\tnote{'} & - & 32.57 & 48.42 & 22.78\% & 53.79 & 77.08 & 43.01\% & 92.64 & 126.22 & 92.85\% & 56.44 & 79.82 & 48.87\% \\
& LSTM\tnote{'} & 98K & 20.41 & 33.47 & 15.60\% & 27.50 & 43.64 & 23.25\% & 38.85 & 60.46 & 37.47\% & 27.88 & 44.23 & 24.31\% \\
& DCRNN\tnote{'} & 373K & 18.25 & 29.73 & 14.37\% & 22.25 & 35.04 & 19.82\% & 28.68 & 44.39 & 28.69\% & 22.35 & 35.26 & 20.15\% \\
& AGCRN\tnote{'} & 777K & 18.11 & 30.19 & 13.64\% & 20.86 & 34.42 & 16.24\% & 24.06 & 39.47 & 19.29\% & 20.55 & 33.91 & 16.06\% \\
& STGCN\tnote{'} & 1.3M & 20.62 & 33.81 & 15.84\% & 23.19 & 37.96 & 18.09\% & 26.53 & 43.88 & 21.77\% & 23.03 & 37.82 & 18.20\% \\
& GWNET\tnote{'} & 344K & 17.74 & 28.92 & 14.37\% & 20.98 & 33.50 & 17.77\% & 25.39 & 40.30 & 22.99\% & 20.78 & 33.32 & 17.76\% \\
& STGODE\tnote{'} & 788K & 18.80 & 30.53 & 15.67\% & 22.19 & 35.91 & 18.54\% & 26.27 & 43.07 & 22.71\% & 21.86 & 35.57 & 17.76\% \\
& DSTAGNN\tnote{'} & 26.9M  & 19.87 & 31.54 & 16.85\% & 23.89 & 38.11 & 19.53\% & 28.48 & 44.65 & 24.65\% & 23.39 & 37.07 & 19.58\% \\
& DGCRN\tnote{'} & 374K & 18.09 & 29.27 & 15.32\% & 21.18 & 33.78 & 18.59\% & 25.73 & 40.88 & 23.67\% & 21.10 & 33.76 & 18.58\% \\
& D$^2$STGNN\tnote{'} & 446K & 17.20 & 28.50 & 12.22\% & 20.80 & 33.53 & 15.32\% & 25.72& 40.90 & 19.90\% & 20.71 & 33.44 & 15.23\% \\
& ASTGCN\tnote{'} & 22.30M & 21.40 & 33.61 & 17.65\% & 26.70 & 40.75 & 24.02\% & 33.64 & 51.21 & 31.15\% & 26.15 & 40.25 & 23.29\% \\
&STWave\tnote{'''}&{\Huge \texttimes}& 17.95 &29.42& 13.01\% &20.99 &34.01& 15.62\% & 24.96& 40.31& 20.08\% &20.81 &33.77 &15.76\% \\
&BigST\tnote{'''}&{\Huge \texttimes}& 18.70 &30.27& 15.55\% & 22.21& 35.33& 18.54\%  &26.98 &42.73& 23.68\%  &21.95& 35.54& 18.50\% \\
&RPMixer~\cite{yeh2024rpmixer} &2.3M& 17.35& 28.69& 13.42\% &19.44& 32.04& 15.61\% &21.65 &36.20 &17.42\%& 19.06& 31.54 &15.09\%\\
&PatchSTG\tnote{'''}& 3.11M&16.81 &28.71 &12.25\% & 19.68 &33.09& 14.51\%  &23.49 &39.23 &18.93\%  &19.50& 33.16 &14.64\% \\
&RAGL~\cite{wu2025regularized} &{\Huge \texttimes}&15.71 &27.58& 10.29\%& 18.40 &31.89 &12.23\%& 22.48 &38.39 &\textbf{\textcolor{red}{15.92\%}} &18.33& 31.65 &12.18\%\\
\cmidrule(r){2-15}
 & VMGCN-v1-14'' & 22.38M &\textbf{\textcolor{red}{2.90}} & \textbf{\textcolor{red}{5.32}} & \textbf{\textcolor{red}{3.27\%}} & \textbf{\textcolor{red}{6.47}} & \textbf{\textcolor{red}{11.62}} & \textbf{\textcolor{red}{6.86\%}} & \textbf{\textcolor{red}{16.42}} & \textbf{\textcolor{red}{26.45}} & 17.55\% & \textbf{\textcolor{red}{8.04}} & \textbf{\textcolor{red}{13.55}} & \textbf{\textcolor{red}{8.57\%}} \\
  & VMGCN-v2-14'' & 22.38M &3.67 & 6.34 & 4.16\% & 7.05 & 12.25 & 7.57\% & 17.24 & 27.22 & 18.58\% & 8.75 & 14.37 & 9.41\% \\
  & VMGCN-v3-14'' & 22.38M &3.58 & 6.24 & 4.06\% & 7.06 & 12.31 & 7.66\% & 17.47 & 27.66 & 18.54\% & 8.77 & 14.45 & 9.42\% \\
\hline
\hline
\multirow{13}{*}{GLA} 
& HL\tnote{'}& - & 33.66 & 50.91 & 19.16\% & 56.88 & 83.54 & 34.85\% & 98.45 & 137.52 & 71.14\% & 56.58 & 86.19 & 38.76\% \\
& LSTM\tnote{'} & 98K & 20.09 & 32.41 & 11.82\% & 27.80 & 44.10 & 16.52\% & 39.61 & 61.57 & 25.63\% & 28.12 & 44.40 & 17.31\% \\
& DCRNN\tnote{'} & 373K & 18.33 & 29.13 & 10.78\% & 22.70 & 35.55 & 13.74\% & 29.45 & 45.88 & 18.87\% & 22.73 & 35.65 & 13.97\% \\
& AGCRN\tnote{'} & 792K & 17.57 & 30.83 & 10.86\% & 20.79 & 36.09 & 13.11\% & 25.01 & 44.82 & 16.11\% & 20.61 & 36.23 & 12.99\% \\
& STGCN\tnote{'} & 2.1M & 19.87 & 34.01 & 12.58\% & 22.54 & 38.57 & 13.94\% & 26.48 & 45.61 & 16.92\% & 22.48 & 38.55 & 14.15\% \\
& GWNET\tnote{'} & 374K & 17.30 & 27.72 & 10.69\% & 21.22 & 33.64 & 13.48\% & 27.25 & 43.03 & 18.49\% & 21.23 & 33.68 & 13.72\% \\
& STGODE\tnote{'} & 841K & 18.46 & 30.05 & 11.94\% & 22.24 & 36.68 & 14.67\% & 27.14 & 45.38 & 19.12\% & 22.02 & 36.34 & 14.93\% \\
& DSTAGNN\tnote{'} & 66.3M  & 19.35 & 30.55 & 11.33\% & 24.22 & 38.19 & 15.90\% & 230.32 & 48.37 & 23.51\% & 23.87 & 37.88 & 15.36\% \\
& DGCRN\tnote{'} & 432K & 17.63 & 8.12 & 10.50\% & 21.15 & 33.70 & 13.06\% & 26.18 & 42.16 & 17.40\% & 21.02 & 33.66 & 13.23\% \\
& D$^2$STGNN\tnote{'} & 284K & 19.31 & 30.07 & 11.82\% & 22.52 & 35.22 & 14.16\% & 27.46& 43.37 & 18.54\% & 22.35 & 35.11 & 14.37\% \\
& ASTGCN\tnote{'}  & 59.1M & 21.11 & 32.41 & 11.82\% & 27.80 & 44.67 & 17.79\% & 39.39 & 59.31 & 28.03\% & 28.12 & 44.40 & 18.62\% \\
&STWave\tnote{'''} &{\Huge \texttimes}&17.48& 28.05& 10.06\%&21.08& 33.58& 12.56\%& 25.82 &41.28& 16.51\% &20.96 &33.48& 12.70\% \\
&BigST\tnote{'''}&{\Huge \texttimes}& 18.38 &29.40& 11.68\%& 22.22& 35.53& 14.48\%& 27.98 &44.74& 19.65\% &22.08& 36.00 &14.57\% \\
&RPMixer~\cite{yeh2024rpmixer}&3.2M &16.49 &26.75& 9.75\%& 18.82 &30.56& 11.58\% &21.18& 35.10& 13.46\% &18.46 &30.13 &11.34\%\\
&PatchSTG\tnote{'''}&1.68M& 15.84 &26.34 &9.27\%& 19.06 &31.85& 11.30\% &23.32 &39.64 &14.60\% &18.96 &32.33 &11.44\% \\
&RACL~\cite{wu2025regularized}&{\Huge \texttimes} &15.06 &25.66& 8.39\%& 17.84& 30.24& 10.09\%& 21.72 &36.73& \textbf{\textcolor{red}{12.98\%}} &17.75& 30.11& 10.20\%\\
\cmidrule(r){2-15}
& VMGCN-v1-13'' & 59.2M &\textbf{\textcolor{red}{3.88}} & \textbf{\textcolor{red}{10.78}} & \textbf{\textcolor{red}{3.99\%}} & \textbf{\textcolor{red}{8.27}} & \textbf{\textcolor{red}{22.34}} & \textbf{\textcolor{red}{7.85\%}} & \textbf{\textcolor{red}{16.78}} & \textbf{\textcolor{red}{31.46}} & 14.28\%& \textbf{\textcolor{red}{9.22}} & \textbf{\textcolor{red}{20.69}} & \textbf{\textcolor{red}{8.23\%}} \\
 & VMGCN-v2-13'' & 59.2M &4.54 & 11.47 & 4.58\% & 8.75 & 21.60 & 8.54\% & 18.21 & 32.56 & 16.39\% & 9.99 & 21.10 & 9.17\% \\
  & VMGCN-v3-13'' & 59.2M &4.62 & 11.54 & 4.65\% & 8.84 & 21.51 & 8.65\% & 17.98 & 31.53 & 15.57\% & 9.98 & 20.78 & 9.09\% \\
\hline
\hline
\end{tabular}
\begin{tablenotes}
    \item['] results executed by~\cite{LargeST}.
    \item[''] results executed by us.
    \item['''] results executed by~\cite{fang2024efficient}.
  \end{tablenotes}
\end{threeparttable}
}
\label{Tab:comparison}
\end{table*}

\begin{figure}[!t]
    \centering
    \begin{subfigure}[b]{0.32\textwidth} 
        \centering
        \includegraphics[width=\textwidth]{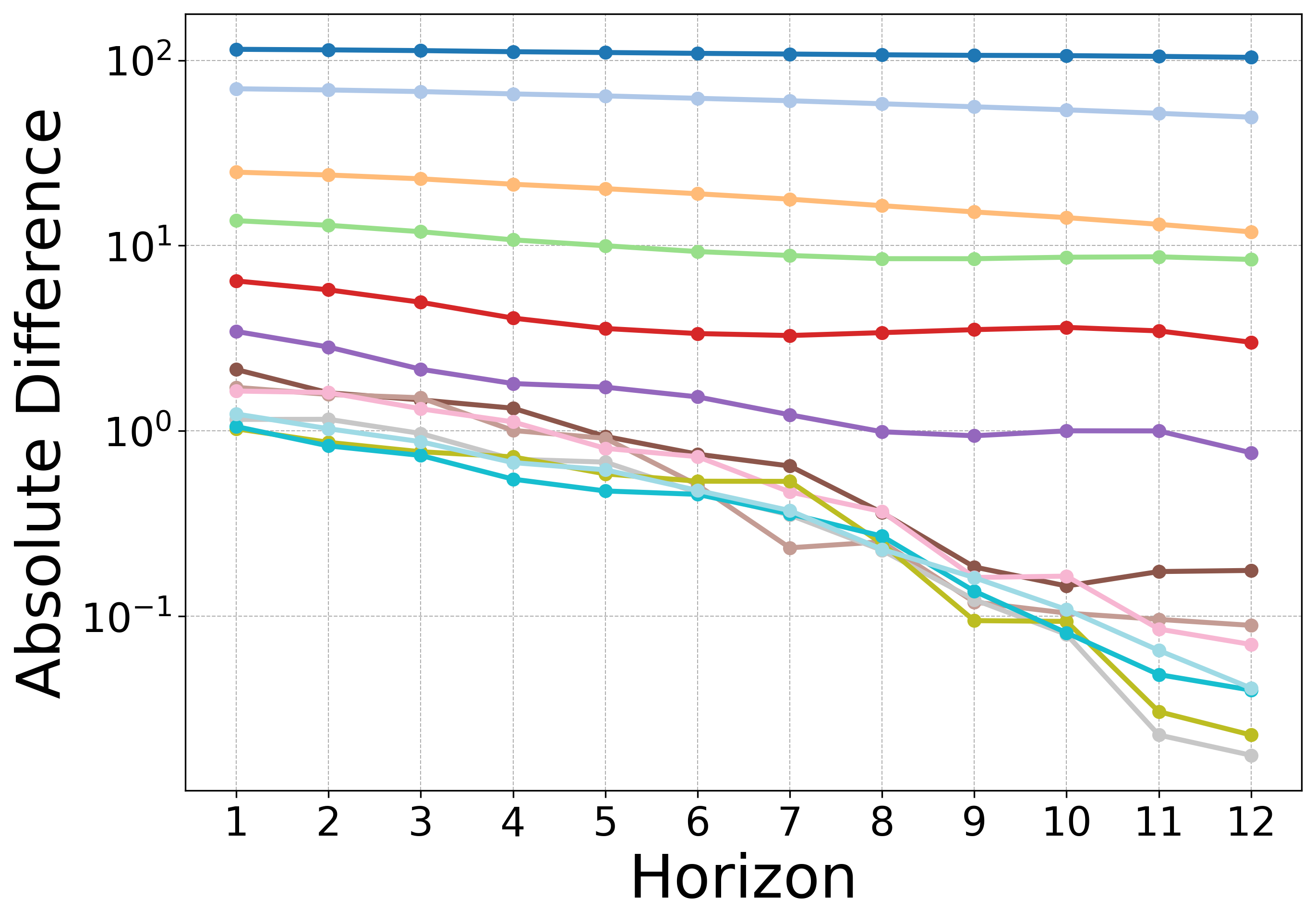}
        \caption{MAE}
        \label{fig:noise_modes_accuracy(a)}
    \end{subfigure}
    \hfill 
    \begin{subfigure}[b]{0.32\textwidth} 
        \centering
        \includegraphics[width=\textwidth]{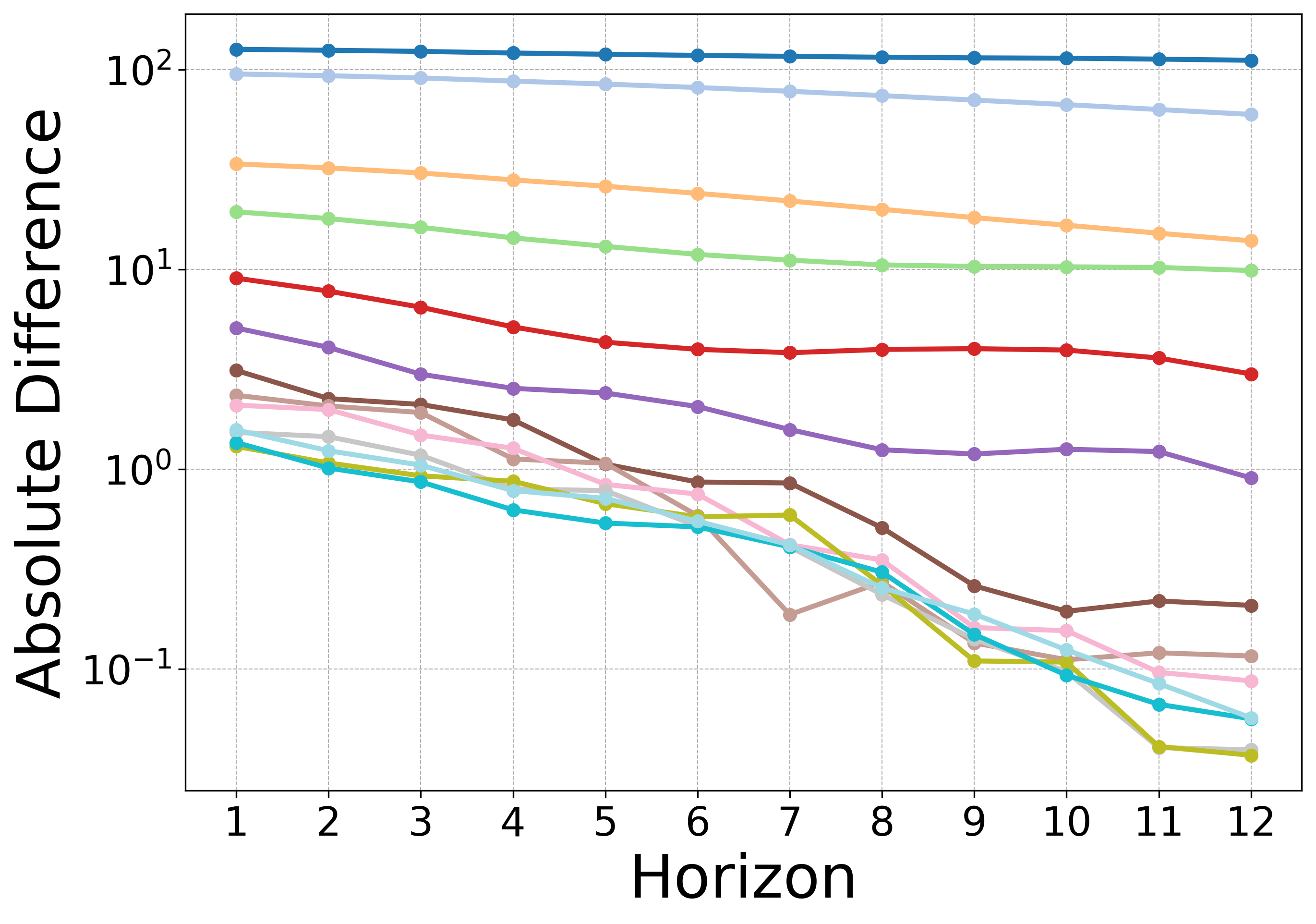}
        \caption{RMSE}
        \label{fig:noise_modes_accuracy(b)}
    \end{subfigure}
    \hfill
    \begin{subfigure}[b]{0.32\textwidth} 
        \centering
        \includegraphics[width=\textwidth]{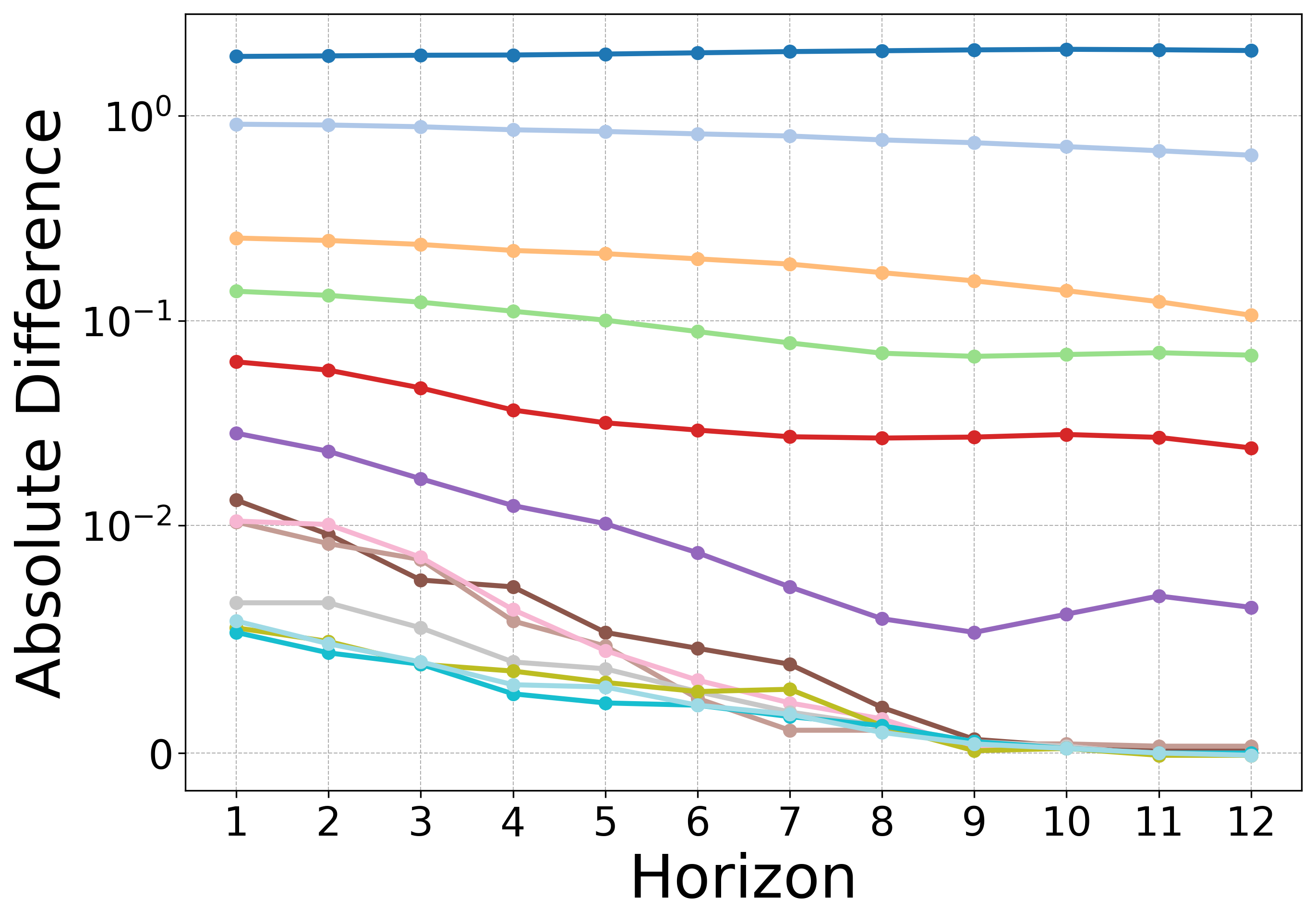}
        \caption{MAPE}
        \label{fig:noise_modes_accuracy(c)}
    
    \end{subfigure}
    \begin{subfigure}[b]{1.0\textwidth} 
        \centering
        \includegraphics[width=\textwidth]{figures/RMSE_K_legend.png}
        \label{fig:noise_modes_accuracy_legend}
    
    \end{subfigure}
    \caption{The importance of each mode in horizon prediction is studied by replacing the respective mode with a zero value and performing inference to observe the change in the performance metrics. The plot shows the difference in metrics between the original and distorted modes on the logarithmic scale. The greater the difference, the more significant the mode for that horizon. This analysis has been performed on the SD region for VMGCN-v2.}
    \label{fig:noise_modes_accuracy}
\end{figure}

\begin{figure}[!tb]
    \centering
    \begin{subfigure}[b] {0.8\textwidth}
            \includegraphics[width=1.0\textwidth,keepaspectratio=true]{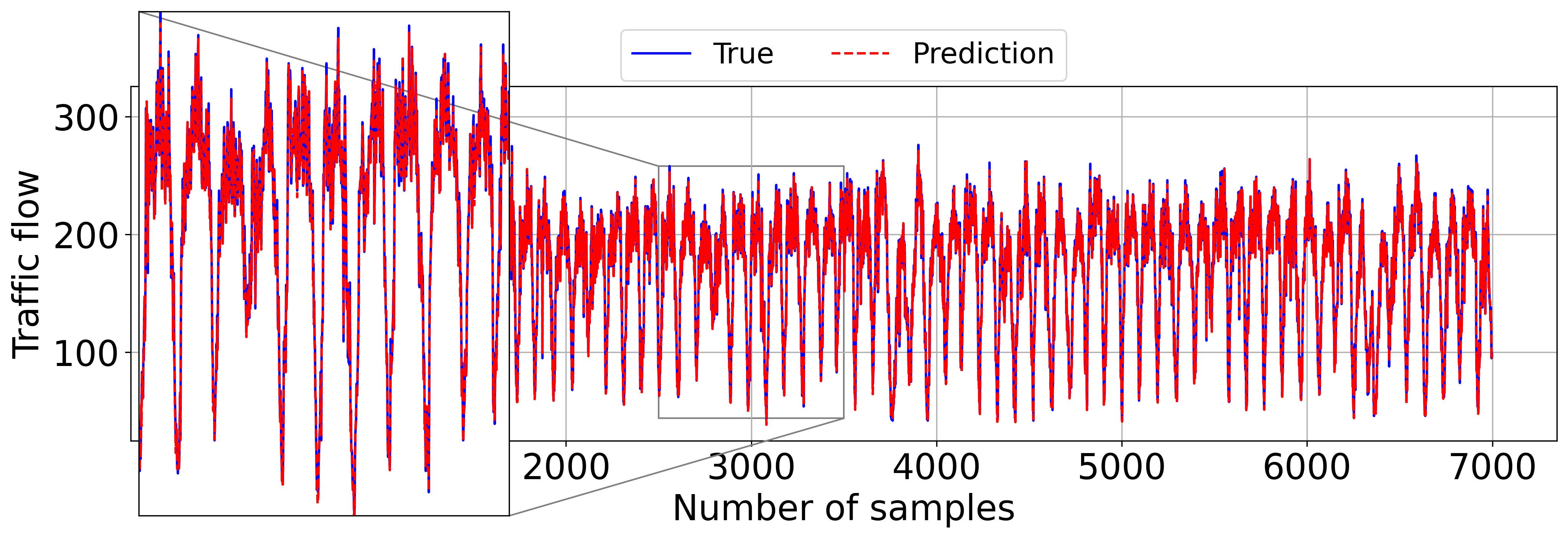}
            \caption{Sensor ID: 1114091 }
    \end{subfigure}
\begin{subfigure}[b] {0.8\textwidth}
            \includegraphics[width=1.0\textwidth,keepaspectratio=true]{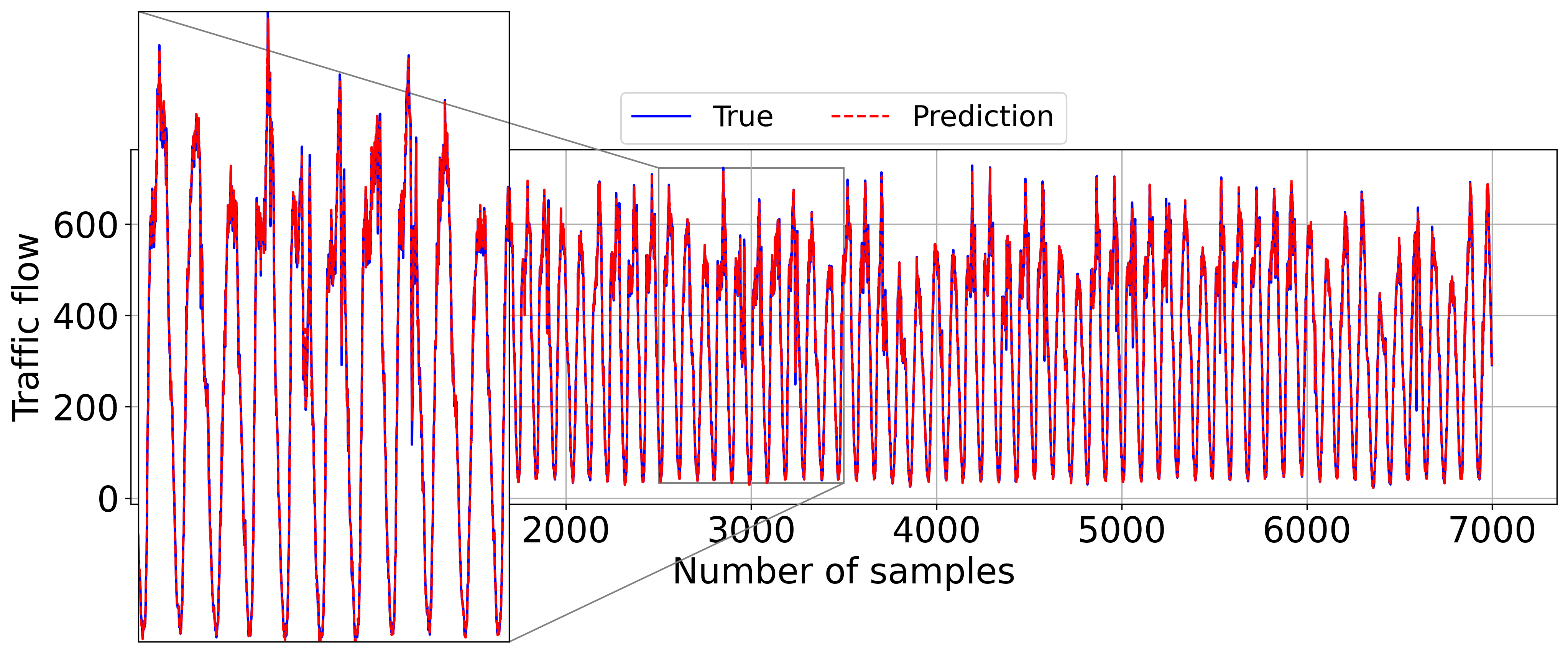}
            \caption{Sensor ID: 1108597 }
    \end{subfigure}
    \begin{subfigure}[b] {0.8\textwidth}
            \includegraphics[width=1.0\textwidth,keepaspectratio=true]{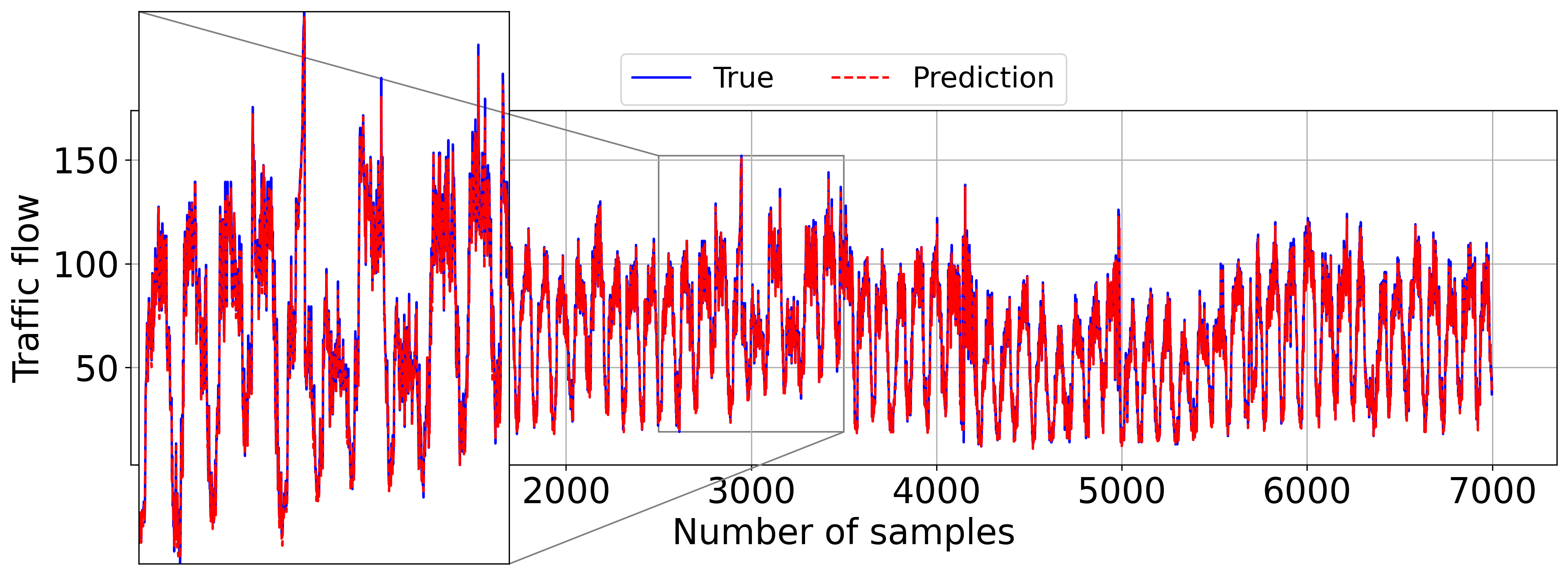}
            \caption{Sensor ID: 1123256}
    \end{subfigure}
    \caption{ This plot shows the comparison of ground truth with the predicted traffic flow for horizon 1 on three distinct locations in the SD region.}
    \label{fig:response_compare}
\end{figure}

To evaluate the effectiveness of each mode on horizon prediction, the modes are analyzed in terms of central frequencies and energy. This relationship between prediction accuracy and the modes, separated based on frequencies, is investigated in Fig.~\ref{fig:noise_modes_accuracy}. In VMGCN-v2 for the SD region, inference is performed by setting each mode to zero in the test set.  Fig.~\ref{fig:noise_modes_accuracy(a)}, Fig.~\ref{fig:noise_modes_accuracy(b)}, 
 and Fig.~\ref{fig:noise_modes_accuracy(c)} show the logarithm of non-zero $\Delta$~MAE, $\Delta$~RMSE, and $\Delta$~MAPE for all horizons, respectively. Here, $\Delta$ represents the difference between the metric with all features and the metric with zero value in $k^{\rm{th}}$ mode for a horizon. The modes from $1$ to $13$ are arranged in terms of increasing frequency components. A higher metric value signifies the importance of that mode in horizon prediction. It has been observed that lower-frequency components (IMF1-IMF4) contribute more to prediction accuracy relative to high-frequency components. As frequency increases, the noise in the mode also increases, and the pattern useful for prediction decreases. High-frequency components (IMF7-IMF13) are significant in short-term predictions; their contributions to long-term predictions decrease. Removing the high-frequency components during the training of the neural network could potentially enhance the performance capabilities of the model. We also compare the predicted traffic flow and the actual flow for Horizon $1$  plotted in Fig.~\ref{fig:response_compare} for three sensors at different locations. It has been observed that the model under-predicts some values at peak times that correspond to rush hours, and sensors 2 and 3 exhibit some periodic trends with harmonic noise. 
 \begin{table*}[!b]
\centering
\caption{Analysis of performance evaluation metrics MAE, MAPE, and RMSE for different VMD parameters~(SD region).}
\resizebox{.99\textwidth}{!}{
\begin{tabular}{c|c|c|ccc|ccc|ccc|ccc}
\toprule
 Description &Hyper-Parameters &  Reconstruction Loss & \multicolumn{3}{c}{Horizon 3} & \multicolumn{3}{c}{Horizon 6} & \multicolumn{3}{c}{Horizon 12} & \multicolumn{3}{c}{Average}  \\
\cmidrule(r){4-15}
& & &MAE & RMSE & MAPE& MAE & RMSE& MAPE & MAE & RMSE & MAPE& MAE & RMSE & MAPE \\
\midrule
VMGCN-v2&$\alpha=2000, \epsilon=10^{-7}$& $9.53\times10^{-4}$& 5.62 & 9.21 & 5.19\% & 10.27 & 17.00 & 9.04\% & 18.59 & 31.73 & 18.15\% & 10.84 & 18.26 & 10.10\%\\
VMGCN-v2&$\alpha=2000, \epsilon=10^{-6}$ & $9.53\times10^{-4}$ & 5.48 & 8.91 & 5.13\% & 10.07 & 16.00 & 8.47\% & 17.46 & 26.89 & 14.99\% & 10.42 & 16.34 & 8.92\% \\
VMGCN-v2&$\alpha=1000, \epsilon=10^{-7}$ & $3.22\times10^{-4}$  & 4.25 & 7.76 & 4.06\% & 8.65 & 14.40 & 7.79\% & 17.56 & 26.57 & 15.55\% & 9.43 & 15.10 & 8.46\% \\
VMGCN-v2&$\alpha=1000, \epsilon=10^{-6}$ & $3.22\times10^{-4}$ & 4.17 & 7.74 & 3.96\% & 8.45 & 14.30 & 7.59\% & 16.77 & 25.86 & 14.83\% & 9.41 & 14.90 & 8.16\% \\
VMGCN-v1(without IMF13)&$\alpha=2000, \epsilon=10^{-7}$ &$1.51\times10^{-3}$& 4.83 & 8.19 & 4.33\% & 9.41 & 15.11 & 7.40\% & 15.15 & 23.56 & 12.33\% & 9.34 & 14.92 &7.49\% \\
\hline
\end{tabular}
}
\label{tab:hyperparameter}
\end{table*}

\subsection{Ablative Study}
The performance of the framework has been evaluated on various combinations of decomposition hyper-parameters such as $\alpha$, and $\epsilon$, for the SD region. TABLE~\ref{tab:hyperparameter} details different performance metrics for horizons $3$, $6$, and $12$, and the average. It has been observed that for $\alpha=1000$, the reconstruction loss is relatively lower compared to $\alpha=2000$. Changing the $\epsilon$ value does not significantly affect the reconstruction loss, indicating that the convergence is sensitive to initial conditions, which may lead to local minima instead of global minimum. For VMGCN-v2, it is noticed that decreasing the bandwidth constraint and increasing the tolerance value improves the performance metrics. The overall cumulative effect on the reconstruction loss term in the region remains unaffected by changes in the $\epsilon$ term. However, the difference in the reconstructed loss term of modes per node is defined as
\begin{equation}
\frac{1}{L} \lvert \sum_i u(t)_{i,\epsilon_1} - \sum_i u(t)_{i,\epsilon_2} \rvert,
\end{equation}
where $L$ is the total length of the signal. For this comparison, we used $\epsilon_1=1\times10^{-7}$ and $\epsilon_2=1\times10^{-6}$.  With a bandwidth constraint of $\alpha=1000$, the average loss term on the SD region is measured as $1.134\times10^{-5}$, indicating a convergence difference between these sets of decomposition parameters, with different local minima being reached. The framework performs better with $\epsilon_2$ compared to $\epsilon_1$. If the bandwidth constraint is set at $\alpha=2000$, the $L_1$ loss term of the mode approaches zero. This indicates that the role of tolerance in convergence is less significant for a larger bandwidth value. From the table, we can also infer that the changing the bandwidth constraint has a dominant impact on the performance of the network. By eliminating the high-frequency (IMF13) in VMGCN-v1, the DNN is trained and the performance of the model is evaluated. It appears that the overall performance of the network is enhanced, specifically the long-term predictions. Furthermore, we observe that the high-frequency component adversely affects the long-term prediction capability of the network.

\subsection{Computational Complexity}
In TABLE~\ref{Tab:comparison}, we compare the number of parameters of our framework, which includes $K$ modes and additional features, with the state-of-the-art ASTGCN method. It shows that the increase in parameters by adding features to the traffic flow is negligible. However, the time complexity in decoupling the traffic counts takes a significant amount of time relative to training the model. The computational complexity for each element of an ST block in ASTGCN is formulated in this section. Typically,  $N \gg d $ and $N \gg T_w$ for targeted traffic prediction applications. The time complexity for normalized spatial attention $\mathbf{S}^{'}$ is $\mathcal{O}(N^3)$. The complexity for normalized temporal attention is $\mathcal{O}(N d T_w)$, for Chebyshev convolution is $\mathcal{O}(T_w M N^2 d)$, and for time convolution is $\mathcal{O}(F O_h O_w K_h K_w C)$, where 
 $   O_h= \lfloor (N-K_h)/S_h \rfloor+1,
$ and 
$    O_w= \lfloor ({T-K_w})/{S_w} \rfloor+1$
are the height and width of the output tensors of the 2D convolution, respectively. The width and height of the kernel are described by $K_w$ and $K_h$, and the strides of the convolution are described by $S_w$ and $S_h$. The number of the CNN output filter is denoted by $F$ and $C$ is the number of the Chebyshev polynomial coefficients feature channels. If we have the $B$ number of ST blocks in ASTGCN, the complexity will be $\mathcal{O}(B(N^3+NdT_w+T_wMN^2d+FO_hO_wK_hK_wC))$.

The computational complexity for VMD involves several components when processing graph signals. The initial mirroring operation of the signal of length $L$ and the subsequent FFT require $\mathcal{O}(L)$ and $\mathcal{O}(L \log L)$ operations respectively. During the decomposition, updating each of the $K$ modes over $N_{\text{iter}}$ iterations incurs a cost of $\mathcal{O}(N_{\text{iter}} K L)$, while the final signal reconstruction through inverse FFTs requires $\mathcal{O}(K L \log L)$ operations. Combining these components, the per-node complexity simplifies to $\mathcal{O}(L \log L + N_{\text{iter}} K L)$. When extended to all $N$ graph nodes (processed independently) with the iteration bound $N_{\text{iter}} \leq N_m$, the overall VMD complexity becomes $\mathcal{O}(N(L \log L + N_{\text{iter}} K L))$. This formulation highlights that the $\mathcal{O}(N L \log L)$ term dominates for typical cases where $N_{\text{iter}}$ remains moderate relative to the signal length $L$. It also shows that the update of modes $\hat{u}_k(\omega)$ and central frequencies $\omega_k$ quantify the computation cost of the decomposition of the signal. The average CPU decomposition time for the SD, GBA, and GLA regions is 6 h, 27 h, and 102 h, respectively. Since we decompose the 1-year data rather than input samples, our framework is well-suited for offline prediction of the ST data. This work focuses on evaluating the dynamic modes and their impact on the accuracy of the model. In future work, we aim to explore asynchronous VMD to decompose batches of time series concurrently to reduce wall-clock decomposition time and make VMGCN suitable for real-time forecasting. The computation of the decomposition of the signal can also be optimized and made learnable by using deep unrolling of the VMD algorithm. 
\section{Conclusion}
\label{sec:conclusion}
The proposed VMGCN-driven forecasting is the first framework to integrate variational mode decomposition with attention-based graph convolutions, achieving both state-of-the-art accuracy and frequency-level interpretability in large-scale traffic forecasting. In this work, we have successfully implemented a hybrid method for both short-term and long-term spatiotemporal (ST) forecasting. The ST data is decomposed into modes, which are then used as features for the deep-learning model to predict future states. The number of modes for real-time applications is determined using the reconstruction loss. The low-frequency modes significantly contribute to horizon prediction while high-frequency modes contain the noise from the signal. By discarding the highest frequency mode from the features, we can enhance the prediction accuracy. However, the bandwidth and tolerance values significantly influence the convergence and performance of our proposed framework. We evaluated the performance of our architecture on the traffic flow dataset, it outperforms in both short and long-term prediction compared to state-of-the-art methods. 
\bibliographystyle{ACM-Reference-Format}
\bibliography{strings}


\begin{thebibliography}{62}


\ifx \showCODEN    \undefined \def \showCODEN     #1{\unskip}     \fi
\ifx \showISBNx    \undefined \def \showISBNx     #1{\unskip}     \fi
\ifx \showISBNxiii \undefined \def \showISBNxiii  #1{\unskip}     \fi
\ifx \showISSN     \undefined \def \showISSN      #1{\unskip}     \fi
\ifx \showLCCN     \undefined \def \showLCCN      #1{\unskip}     \fi
\ifx \shownote     \undefined \def \shownote      #1{#1}          \fi
\ifx \showarticletitle \undefined \def \showarticletitle #1{#1}   \fi
\ifx \showURL      \undefined \def \showURL       {\relax}        \fi
\providecommand\bibfield[2]{#2}
\providecommand\bibinfo[2]{#2}
\providecommand\natexlab[1]{#1}
\providecommand\showeprint[2][]{arXiv:#2}

\bibitem[Bai et~al\mbox{.}(2020)]%
        {AGCRN}
\bibfield{author}{\bibinfo{person}{Lei Bai}, \bibinfo{person}{Lina Yao}, \bibinfo{person}{Can Li}, \bibinfo{person}{Xianzhi Wang}, {and} \bibinfo{person}{Can Wang}.} \bibinfo{year}{2020}\natexlab{}.
\newblock \showarticletitle{Adaptive graph convolutional recurrent network for traffic forecasting}.
\newblock \bibinfo{journal}{\emph{Advances in Neural Information Processing Systems}}  \bibinfo{volume}{33} (\bibinfo{year}{2020}), \bibinfo{pages}{17804--17815}.
\newblock


\bibitem[Cai et~al\mbox{.}(2020)]%
        {TrafficTransformer}
\bibfield{author}{\bibinfo{person}{Ling Cai}, \bibinfo{person}{Krzysztof Janowicz}, \bibinfo{person}{Gengchen Mai}, \bibinfo{person}{Bo Yan}, {and} \bibinfo{person}{Rui Zhu}.} \bibinfo{year}{2020}\natexlab{}.
\newblock \showarticletitle{Traffic transformer: {C}apturing the continuity and periodicity of time series for traffic forecasting}.
\newblock \bibinfo{journal}{\emph{Transactions in GIS}} \bibinfo{volume}{24}, \bibinfo{number}{3} (\bibinfo{year}{2020}), \bibinfo{pages}{736--755}.
\newblock


\bibitem[Carvalho et~al\mbox{.}(2020)]%
        {vmdpy}
\bibfield{author}{\bibinfo{person}{Vin{\'\i}cius~R Carvalho}, \bibinfo{person}{M{\'a}rcio~FD Moraes}, \bibinfo{person}{Ant{\^o}nio~P Braga}, {and} \bibinfo{person}{Eduardo~MAM Mendes}.} \bibinfo{year}{2020}\natexlab{}.
\newblock \showarticletitle{Evaluating five different adaptive decomposition methods for {EEG} signal seizure detection and classification}.
\newblock \bibinfo{journal}{\emph{Biomedical Signal Processing and Control}}  \bibinfo{volume}{62} (\bibinfo{year}{2020}), \bibinfo{pages}{102073--102085}.
\newblock


\bibitem[Chen et~al\mbox{.}(2020)]%
        {chen2020multi}
\bibfield{author}{\bibinfo{person}{Weiqi Chen}, \bibinfo{person}{Ling Chen}, \bibinfo{person}{Yu Xie}, \bibinfo{person}{Wei Cao}, \bibinfo{person}{Yusong Gao}, {and} \bibinfo{person}{Xiaojie Feng}.} \bibinfo{year}{2020}\natexlab{}.
\newblock \showarticletitle{Multi-range attentive bicomponent graph convolutional network for traffic forecasting}. In \bibinfo{booktitle}{\emph{Proceedings of the AAAI conference on artificial intelligence}}, Vol.~\bibinfo{volume}{34}. \bibinfo{pages}{3529--3536}.
\newblock


\bibitem[Choi and Park(2023)]%
        {choi2023graph}
\bibfield{author}{\bibinfo{person}{Jeongwhan Choi} {and} \bibinfo{person}{Noseong Park}.} \bibinfo{year}{2023}\natexlab{}.
\newblock \showarticletitle{Graph neural rough differential equations for traffic forecasting}.
\newblock \bibinfo{journal}{\emph{ACM Transactions on Intelligent Systems and Technology}} \bibinfo{volume}{14}, \bibinfo{number}{4} (\bibinfo{year}{2023}), \bibinfo{pages}{1--27}.
\newblock


\bibitem[Cui et~al\mbox{.}(2019)]%
        {cui2019traffic}
\bibfield{author}{\bibinfo{person}{Zhiyong Cui}, \bibinfo{person}{Kristian Henrickson}, \bibinfo{person}{Ruimin Ke}, {and} \bibinfo{person}{Yinhai Wang}.} \bibinfo{year}{2019}\natexlab{}.
\newblock \showarticletitle{Traffic graph convolutional recurrent neural network: {A} deep learning framework for network-scale traffic learning and forecasting}.
\newblock \bibinfo{journal}{\emph{IEEE Transactions on Intelligent Transportation Systems}} \bibinfo{volume}{21}, \bibinfo{number}{11} (\bibinfo{year}{2019}), \bibinfo{pages}{4883--4894}.
\newblock


\bibitem[Dai et~al\mbox{.}(2017)]%
        {deformablecovolution}
\bibfield{author}{\bibinfo{person}{Jifeng Dai}, \bibinfo{person}{Haozhi Qi}, \bibinfo{person}{Yuwen Xiong}, \bibinfo{person}{Yi Li}, \bibinfo{person}{Guodong Zhang}, \bibinfo{person}{Han Hu}, {and} \bibinfo{person}{Yichen Wei}.} \bibinfo{year}{2017}\natexlab{}.
\newblock \showarticletitle{Deformable convolutional networks}. In \bibinfo{booktitle}{\emph{Proceedings of the IEEE International Conference on Computer Vision}}. \bibinfo{pages}{764--773}.
\newblock


\bibitem[Defferrard et~al\mbox{.}(2016)]%
        {ChebNet}
\bibfield{author}{\bibinfo{person}{Micha{\"e}l Defferrard}, \bibinfo{person}{Xavier Bresson}, {and} \bibinfo{person}{Pierre Vandergheynst}.} \bibinfo{year}{2016}\natexlab{}.
\newblock \showarticletitle{Convolutional neural networks on graphs with fast localized spectral filtering}.
\newblock \bibinfo{journal}{\emph{Advances in Neural Information Processing Systems}}  \bibinfo{volume}{29} (\bibinfo{year}{2016}).
\newblock


\bibitem[Dragomiretskiy and Zosso(2013)]%
        {VMD}
\bibfield{author}{\bibinfo{person}{Konstantin Dragomiretskiy} {and} \bibinfo{person}{Dominique Zosso}.} \bibinfo{year}{2013}\natexlab{}.
\newblock \showarticletitle{Variational mode decomposition}.
\newblock \bibinfo{journal}{\emph{IEEE Transactions on Signal Processing}} \bibinfo{volume}{62}, \bibinfo{number}{3} (\bibinfo{year}{2013}), \bibinfo{pages}{531--544}.
\newblock


\bibitem[Fang et~al\mbox{.}(2025)]%
        {fang2024efficient}
\bibfield{author}{\bibinfo{person}{Yuchen Fang}, \bibinfo{person}{Yuxuan Liang}, \bibinfo{person}{Bo Hui}, \bibinfo{person}{Zezhi Shao}, \bibinfo{person}{Liwei Deng}, \bibinfo{person}{Xu Liu}, \bibinfo{person}{Xinke Jiang}, {and} \bibinfo{person}{Kai Zheng}.} \bibinfo{year}{2025}\natexlab{}.
\newblock \showarticletitle{{Efficient large-scale traffic forecasting with Transformers: A spatial data management perspective}}.
\newblock \bibinfo{journal}{\emph{SIGKDD}} (\bibinfo{year}{2025}).
\newblock


\bibitem[Fang et~al\mbox{.}(2023)]%
        {fang2023spatio}
\bibfield{author}{\bibinfo{person}{Yuchen Fang}, \bibinfo{person}{Yanjun Qin}, \bibinfo{person}{Haiyong Luo}, \bibinfo{person}{Fang Zhao}, \bibinfo{person}{Bingbing Xu}, \bibinfo{person}{Liang Zeng}, {and} \bibinfo{person}{Chenxing Wang}.} \bibinfo{year}{2023}\natexlab{}.
\newblock \showarticletitle{When spatio-temporal meet {W}avelets: {D}isentangled traffic forecasting via efficient spectral graph attention networks}. In \bibinfo{booktitle}{\emph{39th International Conference on Data Engineering (ICDE)}}. IEEE, \bibinfo{pages}{517--529}.
\newblock


\bibitem[Fang et~al\mbox{.}(2021)]%
        {STGODE}
\bibfield{author}{\bibinfo{person}{Zheng Fang}, \bibinfo{person}{Qingqing Long}, \bibinfo{person}{Guojie Song}, {and} \bibinfo{person}{Kunqing Xie}.} \bibinfo{year}{2021}\natexlab{}.
\newblock \showarticletitle{Spatial-temporal graph {ODE} networks for traffic flow forecasting}. In \bibinfo{booktitle}{\emph{Proceedings of the 27th ACM SIGKDD Conference on Knowledge Discovery \& Data Mining}}. \bibinfo{pages}{364--373}.
\newblock


\bibitem[Geng et~al\mbox{.}(2019)]%
        {geng2019spatiotemporal}
\bibfield{author}{\bibinfo{person}{Xu Geng}, \bibinfo{person}{Yaguang Li}, \bibinfo{person}{Leye Wang}, \bibinfo{person}{Lingyu Zhang}, \bibinfo{person}{Qiang Yang}, \bibinfo{person}{Jieping Ye}, {and} \bibinfo{person}{Yan Liu}.} \bibinfo{year}{2019}\natexlab{}.
\newblock \showarticletitle{Spatiotemporal multi-graph convolution network for ride-hailing demand forecasting}. In \bibinfo{booktitle}{\emph{Proceedings of the AAAI conference on artificial intelligence}}, Vol.~\bibinfo{volume}{33}. \bibinfo{pages}{3656--3663}.
\newblock


\bibitem[Guo et~al\mbox{.}(2023)]%
        {VMD_autoencoder}
\bibfield{author}{\bibinfo{person}{Kaixin Guo}, \bibinfo{person}{Xin Yu}, \bibinfo{person}{Gaoxiang Liu}, {and} \bibinfo{person}{Shaohu Tang}.} \bibinfo{year}{2023}\natexlab{}.
\newblock \showarticletitle{A long-term traffic flow prediction model based on variational mode decomposition and auto-correlation mechanism}.
\newblock \bibinfo{journal}{\emph{Applied Sciences}} \bibinfo{volume}{13}, \bibinfo{number}{12} (\bibinfo{year}{2023}), \bibinfo{pages}{7139--7153}.
\newblock


\bibitem[Guo et~al\mbox{.}(2019)]%
        {ASTGCN}
\bibfield{author}{\bibinfo{person}{Shengnan Guo}, \bibinfo{person}{Youfang Lin}, \bibinfo{person}{Ning Feng}, \bibinfo{person}{Chao Song}, {and} \bibinfo{person}{Huaiyu Wan}.} \bibinfo{year}{2019}\natexlab{}.
\newblock \showarticletitle{Attention based spatial-temporal graph convolutional networks for traffic flow forecasting}. In \bibinfo{booktitle}{\emph{Proceedings of the AAAI conference on artificial intelligence}}, Vol.~\bibinfo{volume}{33}. \bibinfo{pages}{922--929}.
\newblock


\bibitem[Guo et~al\mbox{.}(2024)]%
        {R2T_LLM}
\bibfield{author}{\bibinfo{person}{Xusen Guo}, \bibinfo{person}{Qiming Zhang}, \bibinfo{person}{Junyue Jiang}, \bibinfo{person}{Mingxing Peng}, \bibinfo{person}{Hao~Frank Yang}, {and} \bibinfo{person}{Meixin Zhu}.} \bibinfo{year}{2024}\natexlab{}.
\newblock \showarticletitle{{Towards Responsible and Reliable Traffic Flow Prediction with Large Language Models}}.
\newblock \bibinfo{journal}{\emph{Available at SSRN 4805901}} (\bibinfo{year}{2024}).
\newblock


\bibitem[Hamed et~al\mbox{.}(1995)]%
        {ARIMA}
\bibfield{author}{\bibinfo{person}{Mohammad~M Hamed}, \bibinfo{person}{Hashem~R Al-Masaeid}, {and} \bibinfo{person}{Zahi M~Bani Said}.} \bibinfo{year}{1995}\natexlab{}.
\newblock \showarticletitle{Short-term prediction of traffic volume in urban arterials}.
\newblock \bibinfo{journal}{\emph{Journal of Transportation Engineering}} \bibinfo{volume}{121}, \bibinfo{number}{3} (\bibinfo{year}{1995}), \bibinfo{pages}{249--254}.
\newblock


\bibitem[Han et~al\mbox{.}(2024)]%
        {han2024bigst}
\bibfield{author}{\bibinfo{person}{Jindong Han}, \bibinfo{person}{Weijia Zhang}, \bibinfo{person}{Hao Liu}, \bibinfo{person}{Tao Tao}, \bibinfo{person}{Naiqiang Tan}, {and} \bibinfo{person}{Hui Xiong}.} \bibinfo{year}{2024}\natexlab{}.
\newblock \showarticletitle{{BigST}: {L}inear complexity spatio-temporal graph neural network for traffic forecasting on large-scale road networks}. In \bibinfo{booktitle}{\emph{Proceedings of the VLDB Endowment}}, Vol.~\bibinfo{volume}{17}. \bibinfo{publisher}{VLDB Endowment}, \bibinfo{pages}{1081--1090}.
\newblock


\bibitem[Haydari and Y{\i}lmaz(2020)]%
        {ITS_RL}
\bibfield{author}{\bibinfo{person}{Ammar Haydari} {and} \bibinfo{person}{Yasin Y{\i}lmaz}.} \bibinfo{year}{2020}\natexlab{}.
\newblock \showarticletitle{Deep reinforcement learning for intelligent transportation systems: {A} survey}.
\newblock \bibinfo{journal}{\emph{IEEE Transactions on Intelligent Transportation Systems}} \bibinfo{volume}{23}, \bibinfo{number}{1} (\bibinfo{year}{2020}), \bibinfo{pages}{11--32}.
\newblock


\bibitem[Hochreiter and Schmidhuber(1997)]%
        {LSTM}
\bibfield{author}{\bibinfo{person}{Sepp Hochreiter} {and} \bibinfo{person}{J{\"u}rgen Schmidhuber}.} \bibinfo{year}{1997}\natexlab{}.
\newblock \showarticletitle{Long short-term memory}.
\newblock \bibinfo{journal}{\emph{Neural computation}} \bibinfo{volume}{9}, \bibinfo{number}{8} (\bibinfo{year}{1997}), \bibinfo{pages}{1735--1780}.
\newblock


\bibitem[Hou et~al\mbox{.}(2021)]%
        {weather_data}
\bibfield{author}{\bibinfo{person}{Yue Hou}, \bibinfo{person}{Zhiyuan Deng}, {and} \bibinfo{person}{Hanke Cui}.} \bibinfo{year}{2021}\natexlab{}.
\newblock \showarticletitle{{Short-Term Traffic Flow Prediction with Weather Conditions: Based on Deep Learning Algorithms and Data Fusion}}.
\newblock \bibinfo{journal}{\emph{Complexity}} \bibinfo{volume}{2021}, \bibinfo{number}{1} (\bibinfo{year}{2021}), \bibinfo{pages}{6662959--6662973}.
\newblock


\bibitem[Huang et~al\mbox{.}(2025)]%
        {huang2025transformer}
\bibfield{author}{\bibinfo{person}{Enfu Huang}, \bibinfo{person}{Zhanshan Zhao}, \bibinfo{person}{Jiao Yin}, \bibinfo{person}{Jinli Cao}, {and} \bibinfo{person}{Hua Wang}.} \bibinfo{year}{2025}\natexlab{}.
\newblock \showarticletitle{{Transformer-Enhanced Adaptive Graph Convolutional Network for Traffic Flow Prediction}}.
\newblock \bibinfo{journal}{\emph{ACM Transactions on Intelligent Systems and Technology}} (\bibinfo{year}{2025}).
\newblock


\bibitem[Huang et~al\mbox{.}(2022)]%
        {EMD_LSTM}
\bibfield{author}{\bibinfo{person}{Haichao Huang}, \bibinfo{person}{Jingya Chen}, \bibinfo{person}{Rui Sun}, {and} \bibinfo{person}{Shuang Wang}.} \bibinfo{year}{2022}\natexlab{}.
\newblock \showarticletitle{Short-term traffic prediction based on time series decomposition}.
\newblock \bibinfo{journal}{\emph{Physica A: Statistical Mechanics and its Applications}}  \bibinfo{volume}{585} (\bibinfo{year}{2022}), \bibinfo{pages}{126441--126456}.
\newblock


\bibitem[Huang et~al\mbox{.}(1998)]%
        {emd_hilbert}
\bibfield{author}{\bibinfo{person}{Norden~E Huang}, \bibinfo{person}{Zheng Shen}, \bibinfo{person}{Steven~R Long}, \bibinfo{person}{Manli~C Wu}, \bibinfo{person}{Hsing~H Shih}, \bibinfo{person}{Quanan Zheng}, \bibinfo{person}{Nai-Chyuan Yen}, \bibinfo{person}{Chi~Chao Tung}, {and} \bibinfo{person}{Henry~H Liu}.} \bibinfo{year}{1998}\natexlab{}.
\newblock \showarticletitle{The empirical mode decomposition and the {Hilbert} spectrum for nonlinear and non-stationary time series analysis}.
\newblock \bibinfo{journal}{\emph{Proceedings of the Royal Society of London. Series A: Mathematical, Physical and Engineering Sciences}} \bibinfo{volume}{454}, \bibinfo{number}{1971} (\bibinfo{year}{1998}), \bibinfo{pages}{903--995}.
\newblock


\bibitem[Jia et~al\mbox{.}(2017)]%
        {rain-fall}
\bibfield{author}{\bibinfo{person}{Yuhan Jia}, \bibinfo{person}{Jianping Wu}, \bibinfo{person}{Moshe Ben-Akiva}, \bibinfo{person}{Ravi Seshadri}, {and} \bibinfo{person}{Yiman Du}.} \bibinfo{year}{2017}\natexlab{}.
\newblock \showarticletitle{Rainfall-integrated traffic speed prediction using deep learning method}.
\newblock \bibinfo{journal}{\emph{IET Intelligent Transport Systems}} \bibinfo{volume}{11}, \bibinfo{number}{9} (\bibinfo{year}{2017}), \bibinfo{pages}{531--536}.
\newblock


\bibitem[Kazemi(2022)]%
        {GNNBook-ch15-kazemi}
\bibfield{author}{\bibinfo{person}{M.~Seyed Kazemi}.} \bibinfo{year}{2022}\natexlab{}.
\newblock \showarticletitle{Dynamic Graph Neural Networks}.
\newblock In \bibinfo{booktitle}{\emph{Graph Neural Networks: Foundations, Frontiers, and Applications}}, \bibfield{editor}{\bibinfo{person}{Lingfei Wu}, \bibinfo{person}{Peng Cui}, \bibinfo{person}{Jian Pei}, {and} \bibinfo{person}{Liang Zhao}} (Eds.). \bibinfo{publisher}{Springer Singapore}, \bibinfo{address}{Singapore}, \bibinfo{pages}{323--349}.
\newblock


\bibitem[Lan et~al\mbox{.}(2022)]%
        {DSTAGNN}
\bibfield{author}{\bibinfo{person}{Shiyong Lan}, \bibinfo{person}{Yitong Ma}, \bibinfo{person}{Weikang Huang}, \bibinfo{person}{Wenwu Wang}, \bibinfo{person}{Hongyu Yang}, {and} \bibinfo{person}{Pyang Li}.} \bibinfo{year}{2022}\natexlab{}.
\newblock \showarticletitle{D{STAGNN}: {Dynamic} spatial-temporal aware graph neural network for traffic flow forecasting}. In \bibinfo{booktitle}{\emph{International Conference on Machine Learning}}. PMLR, \bibinfo{pages}{11906--11917}.
\newblock


\bibitem[Li et~al\mbox{.}(2023)]%
        {DGCRN}
\bibfield{author}{\bibinfo{person}{Fuxian Li}, \bibinfo{person}{Jie Feng}, \bibinfo{person}{Huan Yan}, \bibinfo{person}{Guangyin Jin}, \bibinfo{person}{Fan Yang}, \bibinfo{person}{Funing Sun}, \bibinfo{person}{Depeng Jin}, {and} \bibinfo{person}{Yong Li}.} \bibinfo{year}{2023}\natexlab{}.
\newblock \showarticletitle{Dynamic graph convolutional recurrent network for traffic prediction: {Benchmark} and solution}.
\newblock \bibinfo{journal}{\emph{ACM Transactions on Knowledge Discovery from Data}} \bibinfo{volume}{17}, \bibinfo{number}{1} (\bibinfo{year}{2023}), \bibinfo{pages}{1--21}.
\newblock


\bibitem[Li and Zhong(2023)]%
        {ICEEMDAN_GRU}
\bibfield{author}{\bibinfo{person}{Guangxin Li} {and} \bibinfo{person}{Xiang Zhong}.} \bibinfo{year}{2023}\natexlab{}.
\newblock \showarticletitle{Parking demand forecasting based on improved complete ensemble empirical mode decomposition and GRU model}.
\newblock \bibinfo{journal}{\emph{Engineering Applications of Artificial Intelligence}}  \bibinfo{volume}{119} (\bibinfo{year}{2023}), \bibinfo{pages}{105717--105728}.
\newblock


\bibitem[Li et~al\mbox{.}(2022)]%
        {li2022deep}
\bibfield{author}{\bibinfo{person}{He Li}, \bibinfo{person}{Xuejiao Li}, \bibinfo{person}{Liangcai Su}, \bibinfo{person}{Duo Jin}, \bibinfo{person}{Jianbin Huang}, {and} \bibinfo{person}{Deshuang Huang}.} \bibinfo{year}{2022}\natexlab{}.
\newblock \showarticletitle{Deep spatio-temporal adaptive {3D} convolutional neural networks for traffic flow prediction}.
\newblock \bibinfo{journal}{\emph{ACM Transactions on Intelligent Systems and Technology (TIST)}} \bibinfo{volume}{13}, \bibinfo{number}{2} (\bibinfo{year}{2022}), \bibinfo{pages}{1--21}.
\newblock


\bibitem[Li et~al\mbox{.}(2018)]%
        {DCRNN_MetaLA}
\bibfield{author}{\bibinfo{person}{Yaguang Li}, \bibinfo{person}{Rose Yu}, \bibinfo{person}{Cyrus Shahabi}, {and} \bibinfo{person}{Yan Liu}.} \bibinfo{year}{2018}\natexlab{}.
\newblock \showarticletitle{Diffusion convolutional recurrent neural network: {D}ata-driven traffic forecasting}.
\newblock \bibinfo{journal}{\emph{International Conference on Learning Representations (ICLR)}} (\bibinfo{year}{2018}).
\newblock


\bibitem[Liang et~al\mbox{.}(2021)]%
        {historical_last}
\bibfield{author}{\bibinfo{person}{Yuxuan Liang}, \bibinfo{person}{Kun Ouyang}, \bibinfo{person}{Yiwei Wang}, \bibinfo{person}{Ye Liu}, \bibinfo{person}{Junbo Zhang}, \bibinfo{person}{Yu Zheng}, {and} \bibinfo{person}{David~S Rosenblum}.} \bibinfo{year}{2021}\natexlab{}.
\newblock \showarticletitle{Revisiting convolutional neural networks for citywide crowd flow analytics}. In \bibinfo{booktitle}{\emph{Machine Learning and Knowledge Discovery in Databases: European Conference, ECML PKDD 2020, Ghent, Belgium, September 14--18, 2020, Proceedings, Part I}}. Springer, \bibinfo{pages}{578--594}.
\newblock


\bibitem[Lin et~al\mbox{.}(2019)]%
        {deepstn}
\bibfield{author}{\bibinfo{person}{Ziqian Lin}, \bibinfo{person}{Jie Feng}, \bibinfo{person}{Ziyang Lu}, \bibinfo{person}{Yong Li}, {and} \bibinfo{person}{Depeng Jin}.} \bibinfo{year}{2019}\natexlab{}.
\newblock \showarticletitle{Deep{STN}+: {C}ontext-aware spatial-temporal neural network for crowd flow prediction in metropolis}. In \bibinfo{booktitle}{\emph{Proceedings of the AAAI conference on artificial intelligence}}, Vol.~\bibinfo{volume}{33}. \bibinfo{pages}{1020--1027}.
\newblock


\bibitem[Liu et~al\mbox{.}(2024)]%
        {LargeST}
\bibfield{author}{\bibinfo{person}{Xu Liu}, \bibinfo{person}{Yutong Xia}, \bibinfo{person}{Yuxuan Liang}, \bibinfo{person}{Junfeng Hu}, \bibinfo{person}{Yiwei Wang}, \bibinfo{person}{Lei Bai}, \bibinfo{person}{Chao Huang}, \bibinfo{person}{Zhenguang Liu}, \bibinfo{person}{Bryan Hooi}, {and} \bibinfo{person}{Roger Zimmermann}.} \bibinfo{year}{2024}\natexlab{}.
\newblock \showarticletitle{Largest: {A} benchmark dataset for large-scale traffic forecasting}.
\newblock \bibinfo{journal}{\emph{Advances in Neural Information Processing Systems}}  \bibinfo{volume}{36} (\bibinfo{year}{2024}).
\newblock


\bibitem[Lu(2023)]%
        {VMD_LSTM}
\bibfield{author}{\bibinfo{person}{Jingyi Lu}.} \bibinfo{year}{2023}\natexlab{}.
\newblock \showarticletitle{An efficient and intelligent traffic flow prediction method based on {LSTM} and variational modal decomposition}.
\newblock \bibinfo{journal}{\emph{Measurement: Sensors}}  \bibinfo{volume}{28} (\bibinfo{year}{2023}), \bibinfo{pages}{100843--100850}.
\newblock


\bibitem[Nie et~al\mbox{.}(2024)]%
        {TripletAttention}
\bibfield{author}{\bibinfo{person}{Xuesong Nie}, \bibinfo{person}{Xi Chen}, \bibinfo{person}{Haoyuan Jin}, \bibinfo{person}{Zhihang Zhu}, \bibinfo{person}{Yunfeng Yan}, {and} \bibinfo{person}{Donglian Qi}.} \bibinfo{year}{2024}\natexlab{}.
\newblock \showarticletitle{Triplet attention transformer for spatiotemporal predictive learning}. In \bibinfo{booktitle}{\emph{Proceedings of the IEEE/CVF Winter Conference on Applications of Computer Vision}}. \bibinfo{pages}{7036--7045}.
\newblock


\bibitem[Okutani and Stephanedes(1984)]%
        {kalman}
\bibfield{author}{\bibinfo{person}{Iwao Okutani} {and} \bibinfo{person}{Yorgos~J Stephanedes}.} \bibinfo{year}{1984}\natexlab{}.
\newblock \showarticletitle{Dynamic prediction of traffic volume through {Kalman} filtering theory}.
\newblock \bibinfo{journal}{\emph{Transportation Research Part B: Methodological}} \bibinfo{volume}{18}, \bibinfo{number}{1} (\bibinfo{year}{1984}), \bibinfo{pages}{1--11}.
\newblock


\bibitem[Pan et~al\mbox{.}(2019)]%
        {STMetaNet}
\bibfield{author}{\bibinfo{person}{Zheyi Pan}, \bibinfo{person}{Yuxuan Liang}, \bibinfo{person}{Weifeng Wang}, \bibinfo{person}{Yong Yu}, \bibinfo{person}{Yu Zheng}, {and} \bibinfo{person}{Junbo Zhang}.} \bibinfo{year}{2019}\natexlab{}.
\newblock \showarticletitle{Urban traffic prediction from spatio-temporal data using deep meta learning}. In \bibinfo{booktitle}{\emph{Proceedings of the 25th ACM SIGKDD International Conference on Knowledge Discovery \& Data Mining}}. \bibinfo{pages}{1720--1730}.
\newblock


\bibitem[Rehman and Mandic(2010)]%
        {MEMD}
\bibfield{author}{\bibinfo{person}{Naveed Rehman} {and} \bibinfo{person}{Danilo~P Mandic}.} \bibinfo{year}{2010}\natexlab{}.
\newblock \showarticletitle{Multivariate empirical mode decomposition}.
\newblock \bibinfo{journal}{\emph{Proceedings of the Royal Society A: Mathematical, Physical and Engineering Sciences}} \bibinfo{volume}{466}, \bibinfo{number}{2117} (\bibinfo{year}{2010}), \bibinfo{pages}{1291--1302}.
\newblock


\bibitem[Shao et~al\mbox{.}(2022)]%
        {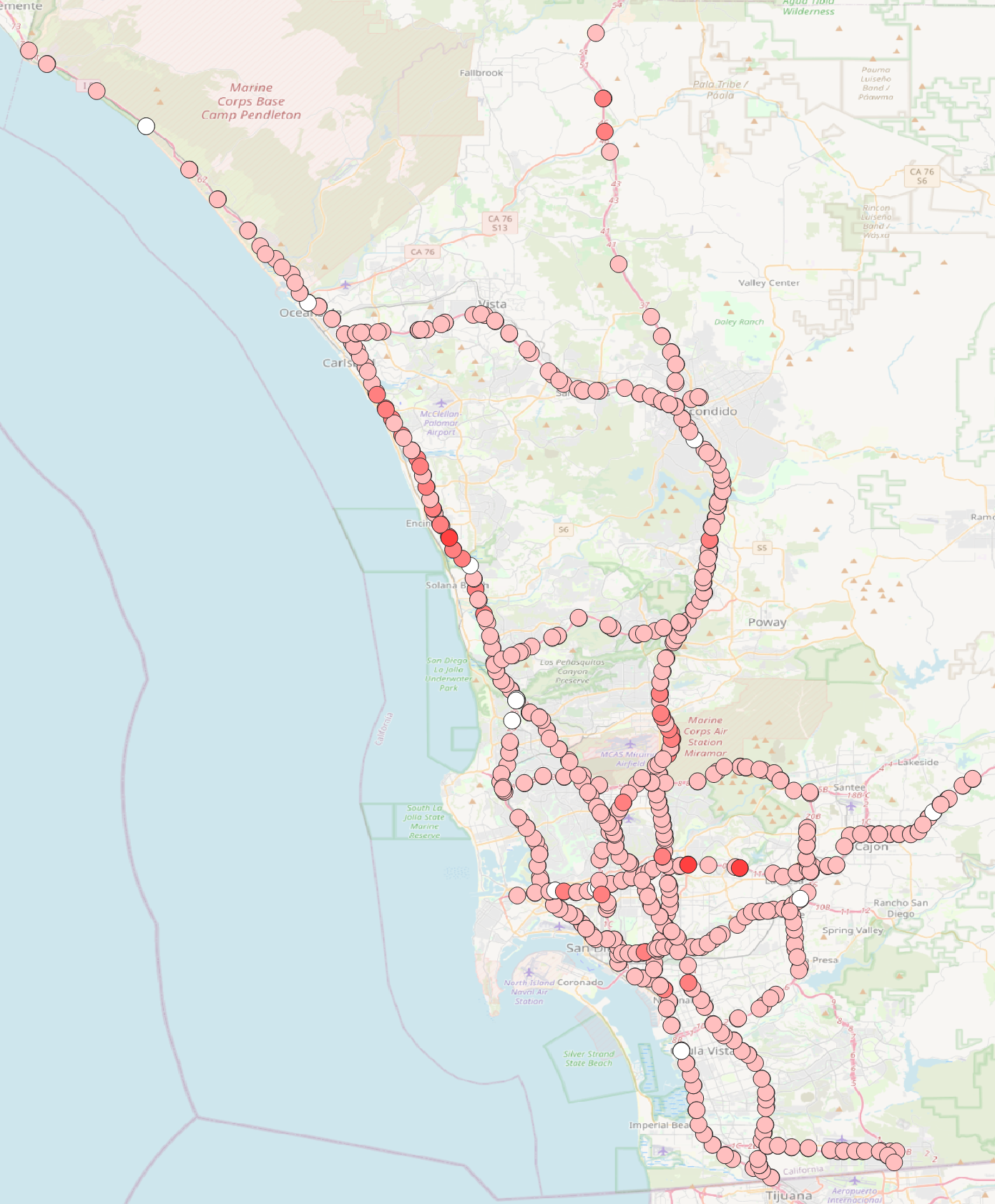}
\bibfield{author}{\bibinfo{person}{Zezhi Shao}, \bibinfo{person}{Zhao Zhang}, \bibinfo{person}{Wei Wei}, \bibinfo{person}{Fei Wang}, \bibinfo{person}{Yongjun Xu}, \bibinfo{person}{Xin Cao}, {and} \bibinfo{person}{Christian~S Jensen}.} \bibinfo{year}{2022}\natexlab{}.
\newblock \showarticletitle{Decoupled dynamic spatial-temporal graph neural network for traffic forecasting}.
\newblock \bibinfo{journal}{\emph{Proceedings of the VLDB Endowment}} (\bibinfo{year}{2022}), \bibinfo{pages}{2733--2746}.
\newblock


\bibitem[Shuman et~al\mbox{.}(2013)]%
        {shuman2013emerging}
\bibfield{author}{\bibinfo{person}{David~I Shuman}, \bibinfo{person}{Sunil~K Narang}, \bibinfo{person}{Pascal Frossard}, \bibinfo{person}{Antonio Ortega}, {and} \bibinfo{person}{Pierre Vandergheynst}.} \bibinfo{year}{2013}\natexlab{}.
\newblock \showarticletitle{The emerging field of signal processing on graphs: {E}xtending high-dimensional data analysis to networks and other irregular domains}.
\newblock \bibinfo{journal}{\emph{IEEE Signal Processing Magazine}} \bibinfo{volume}{30}, \bibinfo{number}{3} (\bibinfo{year}{2013}), \bibinfo{pages}{83--98}.
\newblock


\bibitem[Smith et~al\mbox{.}(2002)]%
        {Nonparameter}
\bibfield{author}{\bibinfo{person}{Brian~L Smith}, \bibinfo{person}{Billy~M Williams}, {and} \bibinfo{person}{R~Keith Oswald}.} \bibinfo{year}{2002}\natexlab{}.
\newblock \showarticletitle{Comparison of parametric and nonparametric models for traffic flow forecasting}.
\newblock \bibinfo{journal}{\emph{Transportation Research Part C: Emerging Technologies}} \bibinfo{volume}{10}, \bibinfo{number}{4} (\bibinfo{year}{2002}), \bibinfo{pages}{303--321}.
\newblock


\bibitem[Su et~al\mbox{.}(2024)]%
        {su2024mdcnet}
\bibfield{author}{\bibinfo{person}{Jing Su}, \bibinfo{person}{Dirui Xie}, \bibinfo{person}{Yuanzhi Duan}, \bibinfo{person}{Yue Zhou}, \bibinfo{person}{Xiaofang Hu}, {and} \bibinfo{person}{Shukai Duan}.} \bibinfo{year}{2024}\natexlab{}.
\newblock \showarticletitle{{MDCNet: Long-term time series forecasting with mode decomposition and 2D convolution}}.
\newblock \bibinfo{journal}{\emph{Knowledge-Based Systems}}  \bibinfo{volume}{299} (\bibinfo{year}{2024}), \bibinfo{pages}{111986--111997}.
\newblock


\bibitem[Sun et~al\mbox{.}(2024)]%
        {sun2024modwavemlp}
\bibfield{author}{\bibinfo{person}{Ke Sun}, \bibinfo{person}{Pei Liu}, \bibinfo{person}{Pengfei Li}, {and} \bibinfo{person}{Zhifang Liao}.} \bibinfo{year}{2024}\natexlab{}.
\newblock \showarticletitle{{ModWaveMLP: MLP-based mode decomposition and Wavelet denoising model to defeat complex structures in traffic forecasting}}. In \bibinfo{booktitle}{\emph{Proceedings of the AAAI Conference on artificial intelligence}}, Vol.~\bibinfo{volume}{38}. \bibinfo{pages}{9035--9043}.
\newblock


\bibitem[Tian et~al\mbox{.}(2025)]%
        {tian2025traffic}
\bibfield{author}{\bibinfo{person}{Xiujuan Tian}, \bibinfo{person}{Shuaihu Wu}, \bibinfo{person}{Xue Xing}, \bibinfo{person}{Huanying Liu}, \bibinfo{person}{Heyao Gao}, {and} \bibinfo{person}{Chun Chen}.} \bibinfo{year}{2025}\natexlab{}.
\newblock \showarticletitle{Traffic flow prediction based on improved deep extreme learning machine}.
\newblock \bibinfo{journal}{\emph{Scientific Reports}} \bibinfo{volume}{15}, \bibinfo{number}{1} (\bibinfo{year}{2025}), \bibinfo{pages}{7421--7448}.
\newblock


\bibitem[Torres et~al\mbox{.}(2011)]%
        {CEEMDAN}
\bibfield{author}{\bibinfo{person}{Mar{\'\i}a~E Torres}, \bibinfo{person}{Marcelo~A Colominas}, \bibinfo{person}{Gaston Schlotthauer}, {and} \bibinfo{person}{Patrick Flandrin}.} \bibinfo{year}{2011}\natexlab{}.
\newblock \showarticletitle{A complete ensemble empirical mode decomposition with adaptive noise}. In \bibinfo{booktitle}{\emph{International Conference on Acoustics, Speech and Signal Processing (ICASSP)}}. IEEE, \bibinfo{pages}{4144--4147}.
\newblock


\bibitem[Van~Lint and Van~Hinsbergen(2012)]%
        {book_transport}
\bibfield{author}{\bibinfo{person}{JWC Van~Lint} {and} \bibinfo{person}{CPIJ Van~Hinsbergen}.} \bibinfo{year}{2012}\natexlab{}.
\newblock \showarticletitle{Short-term traffic and travel time prediction models}.
\newblock \bibinfo{journal}{\emph{Artificial Intelligence Applications to Critical Transportation Issues}} \bibinfo{volume}{22}, \bibinfo{number}{1} (\bibinfo{year}{2012}), \bibinfo{pages}{22--41}.
\newblock


\bibitem[Vaswani et~al\mbox{.}(2017)]%
        {attention}
\bibfield{author}{\bibinfo{person}{Ashish Vaswani}, \bibinfo{person}{Noam Shazeer}, \bibinfo{person}{Niki Parmar}, \bibinfo{person}{Jakob Uszkoreit}, \bibinfo{person}{Llion Jones}, \bibinfo{person}{Aidan~N Gomez}, \bibinfo{person}{{\L}ukasz Kaiser}, {and} \bibinfo{person}{Illia Polosukhin}.} \bibinfo{year}{2017}\natexlab{}.
\newblock \showarticletitle{Attention is all you need}.
\newblock \bibinfo{journal}{\emph{Advances in Neural Information Processing Systems}}  \bibinfo{volume}{30} (\bibinfo{year}{2017}).
\newblock


\bibitem[Wang et~al\mbox{.}(2024b)]%
        {wang2024w}
\bibfield{author}{\bibinfo{person}{Hai-Kun Wang}, \bibinfo{person}{Xuewei Zhang}, \bibinfo{person}{Haicheng Long}, \bibinfo{person}{Shunyu Yao}, {and} \bibinfo{person}{Pengjin Zhu}.} \bibinfo{year}{2024}\natexlab{b}.
\newblock \showarticletitle{{W-FENet: Wavelet-based Fourier-enhanced network model decomposition for multivariate long-term time-series forecasting}}.
\newblock \bibinfo{journal}{\emph{Neural Processing Letters}} \bibinfo{volume}{56}, \bibinfo{number}{2} (\bibinfo{year}{2024}), \bibinfo{pages}{43--66}.
\newblock


\bibitem[Wang et~al\mbox{.}(2023)]%
        {poi_metablock}
\bibfield{author}{\bibinfo{person}{Kuo Wang}, \bibinfo{person}{LingBo Liu}, \bibinfo{person}{Yang Liu}, \bibinfo{person}{GuanBin Li}, \bibinfo{person}{Fan Zhou}, {and} \bibinfo{person}{Liang Lin}.} \bibinfo{year}{2023}\natexlab{}.
\newblock \showarticletitle{Urban regional function guided traffic flow prediction}.
\newblock \bibinfo{journal}{\emph{Information Sciences}}  \bibinfo{volume}{634} (\bibinfo{year}{2023}), \bibinfo{pages}{308--320}.
\newblock


\bibitem[Wang et~al\mbox{.}(2024a)]%
        {wang2024score}
\bibfield{author}{\bibinfo{person}{Pengyu Wang}, \bibinfo{person}{Xuechen Luo}, \bibinfo{person}{Wenxin Tai}, \bibinfo{person}{Kunpeng Zhang}, \bibinfo{person}{Goce Trajcevsky}, {and} \bibinfo{person}{Fan Zhou}.} \bibinfo{year}{2024}\natexlab{a}.
\newblock \showarticletitle{Score-based Graph Learning for Urban Flow Prediction}.
\newblock \bibinfo{journal}{\emph{ACM Transactions on Intelligent Systems and Technology}} \bibinfo{volume}{15}, \bibinfo{number}{3} (\bibinfo{year}{2024}), \bibinfo{pages}{1--25}.
\newblock


\bibitem[Wu et~al\mbox{.}(2025)]%
        {wu2025regularized}
\bibfield{author}{\bibinfo{person}{Kaiqi Wu}, \bibinfo{person}{Weiyang Kong}, \bibinfo{person}{Sen Zhang}, \bibinfo{person}{Yubao Liu}, {and} \bibinfo{person}{Zitong Chen}.} \bibinfo{year}{2025}\natexlab{}.
\newblock \showarticletitle{{Regularized Adaptive Graph Learning for Large-Scale Traffic Forecasting}}.
\newblock \bibinfo{journal}{\emph{arXiv preprint arXiv:2506.07179}} (\bibinfo{year}{2025}).
\newblock


\bibitem[Wu and Huang(2009)]%
        {EEMD}
\bibfield{author}{\bibinfo{person}{Zhaohua Wu} {and} \bibinfo{person}{Norden~E Huang}.} \bibinfo{year}{2009}\natexlab{}.
\newblock \showarticletitle{Ensemble empirical mode decomposition: {A} noise-assisted data analysis method}.
\newblock \bibinfo{journal}{\emph{Advances in Adaptive Data Analysis}} \bibinfo{volume}{1}, \bibinfo{number}{01} (\bibinfo{year}{2009}), \bibinfo{pages}{1--41}.
\newblock


\bibitem[Wu et~al\mbox{.}(2019)]%
        {GWNET}
\bibfield{author}{\bibinfo{person}{Zonghan Wu}, \bibinfo{person}{Shirui Pan}, \bibinfo{person}{Guodong Long}, \bibinfo{person}{Jing Jiang}, {and} \bibinfo{person}{Chengqi Zhang}.} \bibinfo{year}{2019}\natexlab{}.
\newblock \showarticletitle{Graph {Wavenet} for deep spatial-temporal graph modeling}.
\newblock \bibinfo{journal}{\emph{in IJCAI}} (\bibinfo{year}{2019}).
\newblock


\bibitem[Yeh et~al\mbox{.}(2024)]%
        {yeh2024rpmixer}
\bibfield{author}{\bibinfo{person}{Chin-Chia~Michael Yeh}, \bibinfo{person}{Yujie Fan}, \bibinfo{person}{Xin Dai}, \bibinfo{person}{Uday~Singh Saini}, \bibinfo{person}{Vivian Lai}, \bibinfo{person}{Prince~Osei Aboagye}, \bibinfo{person}{Junpeng Wang}, \bibinfo{person}{Huiyuan Chen}, \bibinfo{person}{Yan Zheng}, \bibinfo{person}{Zhongfang Zhuang}, {et~al\mbox{.}}} \bibinfo{year}{2024}\natexlab{}.
\newblock \showarticletitle{Rpmixer: {S}haking up time series forecasting with random projections for large spatial-temporal data}. In \bibinfo{booktitle}{\emph{Proceedings of the 30th ACM SIGKDD Conference on Knowledge Discovery and Data Mining}} (Barcelona, Spain). \bibinfo{pages}{3919--3930}.
\newblock


\bibitem[Yu et~al\mbox{.}(2018)]%
        {STGCN}
\bibfield{author}{\bibinfo{person}{Bing Yu}, \bibinfo{person}{Haoteng Yin}, {and} \bibinfo{person}{Zhanxing Zhu}.} \bibinfo{year}{2018}\natexlab{}.
\newblock \showarticletitle{Spatio-temporal graph convolutional networks: {A} deep learning framework for traffic forecasting}.
\newblock \bibinfo{journal}{\emph{in Proc. IJCAI}} (\bibinfo{year}{2018}), \bibinfo{pages}{3634–--3640}.
\newblock


\bibitem[Zhang et~al\mbox{.}(2020)]%
        {curbGAN}
\bibfield{author}{\bibinfo{person}{Yingxue Zhang}, \bibinfo{person}{Yanhua Li}, \bibinfo{person}{Xun Zhou}, \bibinfo{person}{Xiangnan Kong}, {and} \bibinfo{person}{Jun Luo}.} \bibinfo{year}{2020}\natexlab{}.
\newblock \showarticletitle{Curb-{GAN}: {C}onditional urban traffic estimation through spatio-temporal generative adversarial networks}. In \bibinfo{booktitle}{\emph{Proceedings of the 26th ACM SIGKDD International Conference on Knowledge Discovery \& Data Mining}}. \bibinfo{pages}{842--852}.
\newblock


\bibitem[Zhang et~al\mbox{.}(2022)]%
        {zhang2022urban}
\bibfield{author}{\bibinfo{person}{Yingxue Zhang}, \bibinfo{person}{Yanhua Li}, \bibinfo{person}{Xun Zhou}, \bibinfo{person}{Jun Luo}, {and} \bibinfo{person}{Zhi-Li Zhang}.} \bibinfo{year}{2022}\natexlab{}.
\newblock \showarticletitle{Urban traffic dynamics prediction—a continuous spatial-temporal meta-learning approach}.
\newblock \bibinfo{journal}{\emph{ACM Transactions on Intelligent Systems and Technology (TIST)}} \bibinfo{volume}{13}, \bibinfo{number}{2} (\bibinfo{year}{2022}), \bibinfo{pages}{1--19}.
\newblock


\bibitem[Zhang et~al\mbox{.}(2019)]%
        {TrafficGAN}
\bibfield{author}{\bibinfo{person}{Yuxuan Zhang}, \bibinfo{person}{Senzhang Wang}, \bibinfo{person}{Bing Chen}, \bibinfo{person}{Jiannong Cao}, {and} \bibinfo{person}{Zhiqiu Huang}.} \bibinfo{year}{2019}\natexlab{}.
\newblock \showarticletitle{Traffic{GAN}: Network-scale deep traffic prediction with generative adversarial nets}.
\newblock \bibinfo{journal}{\emph{IEEE Transactions on Intelligent Transportation Systems}} \bibinfo{volume}{22}, \bibinfo{number}{1} (\bibinfo{year}{2019}), \bibinfo{pages}{219--230}.
\newblock


\bibitem[Zhang et~al\mbox{.}(2025)]%
        {zhang2025vmd}
\bibfield{author}{\bibinfo{person}{Yuhong Zhang}, \bibinfo{person}{Kezhen Zhong}, \bibinfo{person}{Xiaopeng Xie}, \bibinfo{person}{Yuzhe Huang}, \bibinfo{person}{Shuai Han}, \bibinfo{person}{Guozhen Liu}, {and} \bibinfo{person}{Ziyan Chen}.} \bibinfo{year}{2025}\natexlab{}.
\newblock \showarticletitle{{VMD-ConvTSMixer: Spatiotemporal channel mixing model for non-stationary time series forecasting}}.
\newblock \bibinfo{journal}{\emph{Expert Systems with Applications}} (\bibinfo{year}{2025}), \bibinfo{pages}{126535--126556}.
\newblock


\bibitem[Zhao et~al\mbox{.}(2023)]%
        {zhao2023hybrid}
\bibfield{author}{\bibinfo{person}{Zeni Zhao}, \bibinfo{person}{Sining Yun}, \bibinfo{person}{Lingyun Jia}, \bibinfo{person}{Jiaxin Guo}, \bibinfo{person}{Yao Meng}, \bibinfo{person}{Ning He}, \bibinfo{person}{Xuejuan Li}, \bibinfo{person}{Jiarong Shi}, {and} \bibinfo{person}{Liu Yang}.} \bibinfo{year}{2023}\natexlab{}.
\newblock \showarticletitle{Hybrid {VMD-CNN-GRU-based} model for short-term forecasting of wind power considering spatio-temporal features}.
\newblock \bibinfo{journal}{\emph{Engineering Applications of Artificial Intelligence}}  \bibinfo{volume}{121} (\bibinfo{year}{2023}), \bibinfo{pages}{105982--105996}.
\newblock


\bibitem[Zheng et~al\mbox{.}(2020)]%
        {zheng2020gman}
\bibfield{author}{\bibinfo{person}{Chuanpan Zheng}, \bibinfo{person}{Xiaoliang Fan}, \bibinfo{person}{Cheng Wang}, {and} \bibinfo{person}{Jianzhong Qi}.} \bibinfo{year}{2020}\natexlab{}.
\newblock \showarticletitle{{GMAN: A} graph multi-attention network for traffic prediction}. In \bibinfo{booktitle}{\emph{Proceedings of the AAAI conference on artificial intelligence}}, Vol.~\bibinfo{volume}{34}. \bibinfo{pages}{1234--1241}.
\newblock


\end{thebibliography}

\end{document}